\documentclass[11pt]{article}
\usepackage{graphicx,color}
\usepackage{subfigure}
\usepackage{amsmath, amsfonts, amssymb, amstext}
\usepackage{srctex}
\usepackage{comment}

\newtheorem{remark}{Remark}%[section]
\newtheorem{prop}{Proposition}
\newtheorem{mydef}{Definition}

\newcommand\independent{\protect\mathpalette{\protect\independenT}{\perp}}
\def\independenT#1#2{\mathrel{\rlap{$#1#2$}\mkern2mu{#1#2}}}

\usepackage{setspace}
\title{Learning directed acyclic graphs via bootstrap aggregating}

\author{Ru Wang and Jie Peng\footnote{correspondence author: jiepeng@ucdavis.edu}\\\\
\textit{Department of Statistics, University of California, Davis, CA 95616} \\
}
\date{}

\begin{document}

\maketitle
\begin{abstract}
Probabilistic graphical models are graphical representations of probability distributions. Graphical models have applications in many fields including biology, social sciences, linguistic, neuroscience. In this paper, we propose directed acyclic graphs (DAGs) learning  via bootstrap aggregating. The proposed procedure is named as \texttt{DAGBag}. Specifically, an ensemble of DAGs is first learned based on bootstrap resamples of the data and then an aggregated DAG is derived by minimizing the overall distance to the entire ensemble. A family of metrics based on the structural hamming distance is defined for the space of DAGs (of a given node set) and is used for aggregation. Under the high-dimensional-low-sample size setting, the graph learned on one data set often has excessive number of false positive edges due to over-fitting of the noise.
Aggregation  overcomes over-fitting through variance reduction and thus greatly reduces false positives. We also develop an efficient implementation of the hill climbing search algorithm of DAG learning which makes the proposed method computationally competitive for the high-dimensional regime.  The \texttt{DAGBag} procedure is implemented in the R package \texttt{dagbag}.
\end{abstract}


\section{Introduction}
 A \textit{directed acyclic graph  model} (a.k.a. \textit{Bayesian network model}) consists of a  \textit{directed acyclic graph (DAG)} $\mathcal{G}(\mathbb{V},\mathbb{E})$
and a probability distribution $\mathbb{P}$ over the node set $\mathbb{V}$  which admits recursive factorization according to  $\mathcal{G}$ \cite{lauritzen1996graphical, koller2009probabilistic}.  The edge set $\mathbb{E}$ represents direct probabilistic interactions among the nodes.
The graph $\mathcal{G}$ provides a compact and modular representation of  $\mathbb{P}$. It also facilitates modeling, interpretation and reasoning.

DAG models have many applications. DAGs have been used to infer causal relationships in various research domains \cite{pearl2000causality}, to process natural languages \cite{bishop2006pattern}, to develop medical intelligence systems, to construct genetic regulatory networks \cite{friedman2000using,pe2001inferring, sachs2005causal}, to learn expression quantitative trait loci (eQTL) \cite{neto2010causal}.

Graphical model learning is an active research field with a large body of literature \cite{koller2009probabilistic}. An important  topic of this field is to recover the underlying graph topology based on an \textit{independently and identically distributed (i.i.d.)} sample from the distribution $\mathbb{P}$ \cite{geiger1994learning, heckerman1995learning, heckerman2008tutorial,  scutari2009learning}. This is often referred to as graphical model \textit{structure learning}.

%Recall that  the graph $\mathcal{G}$ represents a set of conditional independence relations in the associated distribution %$\mathbb{P}$ \cite{lauritzen1996graphical}. Therefore, the essence of structure learning is to elucidate conditional independence %relationships of the nodes  based on an i.i.d. sample.

One motivation for structure learning is to distinguish between direct and indirect interactions.
Consider a simple system $X \rightarrow Y_1 \rightarrow Y_2$, where $X$ is a single nucleotide variant (SNV) and $Y_1, Y_2$ are gene expressions.
If we  adopt the most popular approach by practitioners which simply examines pairwise marginal correlations between SNVs and expression levels, we would declare $X$ as an {\sl eQTL} for both $Y_1,Y_2$. However, if we examine conditional independencies, we would find out that $X$ is only directly associated with $Y_1$, but is conditionally independent with $Y_2$ given $Y_1$. Thus
we could recover the underlying system and (correctly) declare that $X$ is an eQTL for $Y_1$, but not for $Y_2$.

Except for elucidating independencies, structure learning could also help  with parameter estimation, as the set of independencies imposes constraints on the parameter space. Moreover, after structure learning and parameter estimation, the learned model may be used for prediction of future instances.

In this paper, we consider structure learning of DAG models under the high-dimensional setting where the number of nodes $p$ is comparable to or larger than the sample size $n$. With a large number of nodes, the space of DAGs is huge as its size is super-exponential in $p$.
 This leads to computational challenges as well as excessive number of false positive edges due to over-fitting of the noise by commonly used DAG learning methods.

To tackle the computational challenges,  we develop an efficient implementation of the \textit{hill climbing search algorithm}, where at each search step we utilize information from the previous search step to speed up both score calculation and acyclic check.   For a graph with $p = 1000$ nodes, sample size $n = 250$, it  takes about $150$ seconds to conduct $2000$ search steps on a machine
with 8GB memory and 2quad CPU.

DAG structure learning procedures are usually highly variable, i.e., the learnt graph tends to change drastically with even small perturbation of the data.  To tackle this challenge, we propose
to use \textit{bootstrap aggregating (bagging)} \cite{Breiman96} to achieve variance reduction and consequently to reduce the number
of false positives. Specifically, an ensemble of DAGs is first learned based on bootstrap resamples
of the data and then an aggregated DAG is derived by minimizing the overall distance to the entire
ensemble. A family of metrics based on the \textsl{structural hamming distance} is defined for the space
of DAGs (of a given nodes set) and is used for aggregation.
The idea is to look for structures which are stable with respect to data perturbation. Since  this approach is inspired by bagging, it is named as \textit{DAGBag}. It is shown by simulation studies that, the proposed \texttt{DAGBag} procedure greatly reduces the number of false positives in edge detection while sacrificing little in power.
%Through model aggregation, we have been able to greatly reduce the number of false positives.

Although bagging is initially proposed for building stable prediction models, in recent years,  model aggregation techniques have been successfully applied to  variable selection in high-dimensional regression models  \cite{Bach08, wang2011random} and Gaussian graphical model learning \cite{meinshausen2010stability, li2011bootstrap}.

The idea of data perturbation  and model aggregation have been previously considered for DAG learning. \cite{friedman1999data} and \cite{imoto2002bootstrap} propose to measure the confidence for a graphical feature (e.g. an edge) through feature frequencies based on graphs learnt on bootstrap resamples.
 \cite{elidan2011bagged} propose to find a stable prediction model through bagged estimate of the log-likelihood.
\cite{elidan2002data} use data perturbation as a way to escape locally optimal solutions in the search algorithm.
\cite{broom2012model} study model averaging strategies for DAG structure learning under a Bayesian framework through edge selection frequency thresholding, even though there is no guarantee that the resulting graph is acyclic. None of these methods leads to an aggregated DAG  as we are proposing in this paper.

%Data perturbation has also been applied to learn expression networks \cite{pe2001inferring, friedman2000using} and HIV networks \cite{deforche2006analysis}.
%These papers mainly focus on the $p < n$ regime and often aim at building good prediction models.

The rest of the paper is organized as follows. In Section \ref{sec:DAG}, we give a brief overview of DAG models. In Section \ref{sec:search}, we discuss an efficient implementation of the hill climbing search algorithm for DAG structure learning. In Section \ref{sec:aggregation}, we propose aggregation strategies to improve DAG structure learning. In Section \ref{sec:numerical}, we present numerical studies which show that aggregation is an effective way to reduce false positive edges. We conclude the paper by a summary (Section \ref{sec:discussion}). Some details are deferred to Appendices.

\section{Directed Acyclic Graph Models}
\label{sec:DAG}
In this section, we give a brief overview of directed acyclic graph models. For more details, the readers are referred to \cite{lauritzen1996graphical} and \cite{koller2009probabilistic}.
\subsection{Directed acyclic graphs}
\begin{figure}
\begin{center}
\includegraphics[scale=0.75]{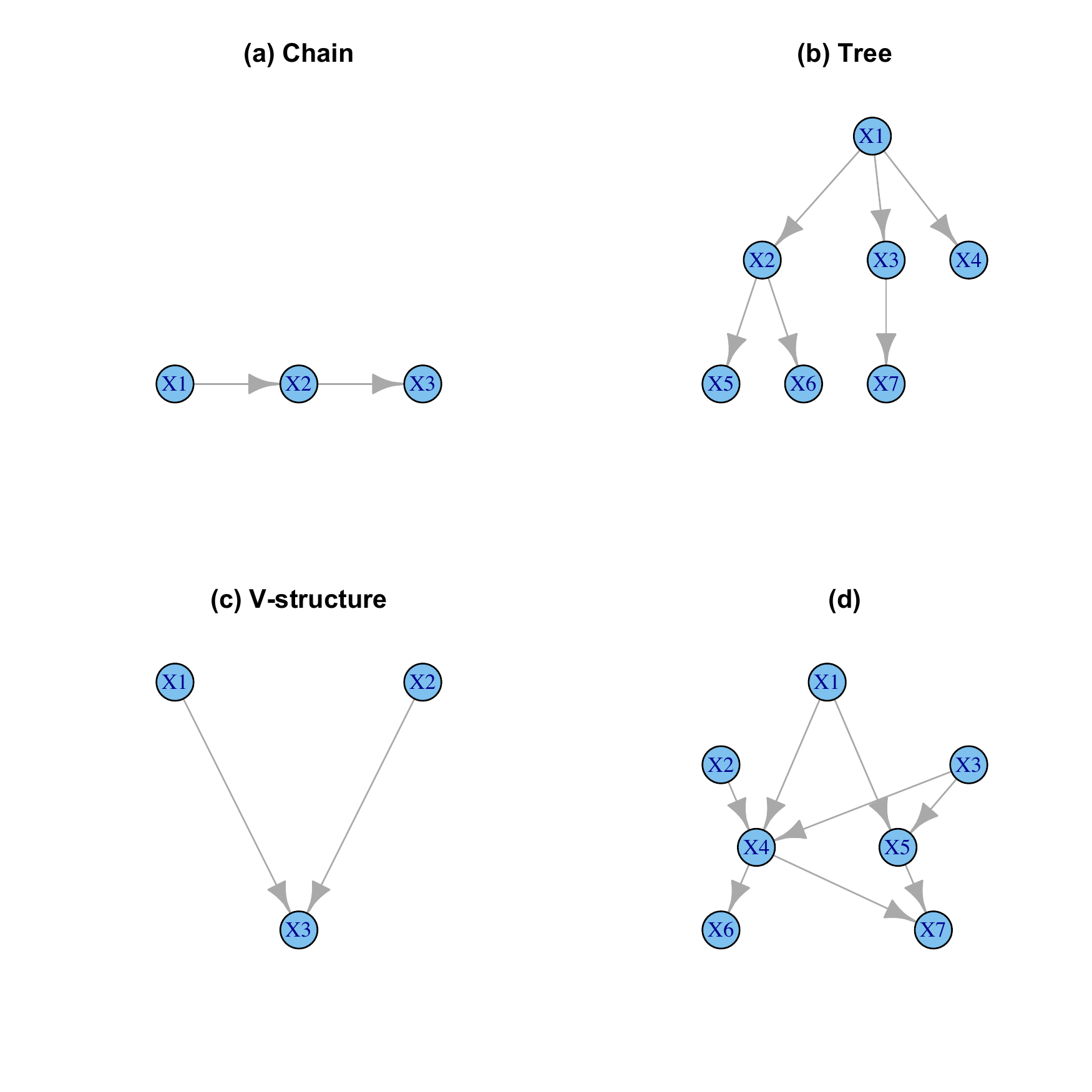}
\caption{Four examples of Directed Acyclic Graph. \label{fig:graph_example}}
%\caption{A DAG (left), its skeleton (middle) and its moral graph (right)}
\end{center}
\end{figure}

A directed acyclic graph $\mathcal{G}(\mathbb{V},\mathbb{E})$ consists of a node set $\mathbb{V}=\{x_1,\cdots,x_p\}$ and an edge set
$\mathbb{E}$ with directed edges of the form $x_i\rightarrow x_j$ ($x_i$ is called a \textit{parent} of $x_j$, and $x_j$ is called  a \textit{child} of $x_i$). As the name suggests, there is no cycle in a DAG, i.e., starting from any node, there is no directed path in the graph leading back to it. The most well known DAGs are tree graphs.  Figure \ref{fig:graph_example} gives four examples of DAG.

For each node $x_j$  ($j=1,\cdots p$), let $pa_j^{\mathcal{G}}=\{x_i \in \mathbb{V}: x_i\rightarrow x_j \in \mathbb{E}\}$ denote its parent set  and $ch_j^{\mathcal{G}}=\{x_k \in \mathbb{V}: x_j \rightarrow x_k  \in \mathbb{E}\}$ denote its children set.
In this paper, $pa_j^{\mathcal{G}}$ is also referred to as the \textit{neighborhood} of $x_j$. Note that, $\mathcal{G}$ is characterized by the parent sets, $\{pa_j^{\mathcal{G}}\}^p_{j=1}$. Therefore, DAG structure learning amounts to identifying the parent set for each node.
Another representation of  a DAG is by an \textit{adjacency matrix}: $\mathbb{A}=(A_{ij})$ where $A_{ij}=1$ if $x_i \rightarrow x_j  \in \mathbb{E}$, otherwise $A_{ij}=0$ ($1 \leq i, j \leq p$).

Two nodes $x_i, x_j$ are said to be \textit{adjacent} in $\mathcal{G}$, denoted by $x_i \sim x_j$, if either $x_i \rightarrow x_j \in \mathbb{E}$ or $x_j \rightarrow x_i \in \mathbb{E}$.
A \textit{v-structure} is a triplet of nodes  $(x_1,x_3,x_2)$, such that  $x_1 \rightarrow x_3 \in \mathbb{E}$,  $x_2 \rightarrow x_3 \in \mathbb{E}$, and $x_1, x_2$ are not adjacent.  Figure \ref{fig:graph_example} ( c ) shows a v-structure:  $x_1 \rightarrow x_3 \leftarrow x_2$. In this v-structure,  $x_1$ and $x_2$ are called \textit{co-parents} of $x_3$.

\begin{figure}
%\caption{\label{fig:moral_graph}}
\begin{center}
\includegraphics[scale=1.2]{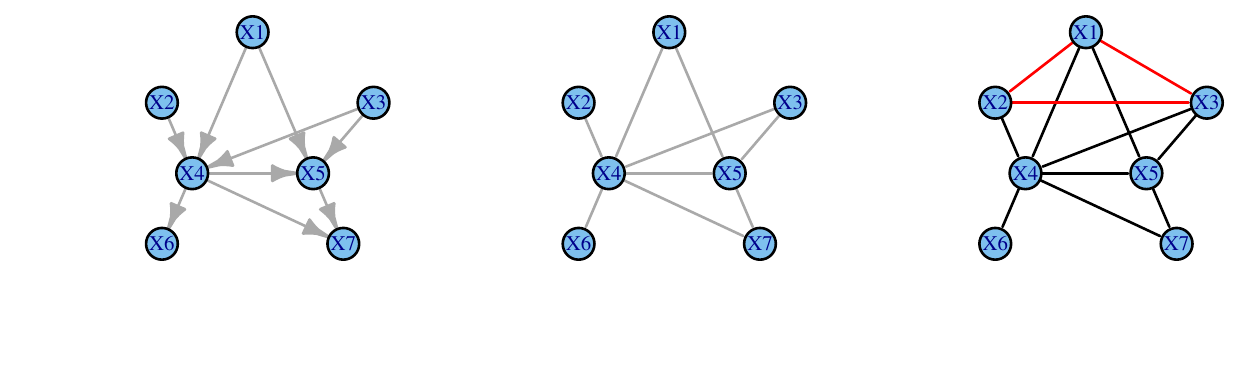}
\caption{A DAG with seven nodes (left), its skeleton graph (middle) and its moral graph (right). \label{fig:moral_graph}}
\end{center}
\end{figure}

There are two undirected graphs associated with a DAG $\mathcal{G}$, namely, a \textit{skeleton graph} which is obtained by discarding edge directions in $\mathcal{G}$, and  a  \textit{moral graph} which  is obtained by adding edges between co-parents in the v-structures and then removing edge directions. Figure \ref{fig:moral_graph} shows a DAG with seven nodes (left), its skeleton graph (middle), and its moral graph (right) with three additional edges due to connecting co-parents highlighted in red color.

An important notion is  \textit{d-separation}.
\begin{mydef}
A \textit{path} from a node $x$ to another node $y$ is a sequence $x=\alpha_0 \sim \alpha_1 \sim \cdots \sim \alpha_m=y$. A path is said to be \textit{blocked} by a subset of nodes $S$ if there exists a node on this path such that,
\begin{itemize}
\item  either it is in the set $S$, and the arrows on the path do not meet head-to-head at this node, or
\item neither the node, nor any of its descendants, is in the set $S$, and the arrows do meet head-to-head at this node.
\end{itemize}
(A node $x_j$ is called a \textit{descendant} of another node $x_i$ if there is a directed path from $x_i$ to $x_j$ and $x_i$ is called an \textit{ancestor} of $x_j$. )
\end{mydef}

\begin{mydef}
Let $A, B, S$ be disjoint subsets of the node set $\mathbb{V}$.
$A$ and $B$ are said to be \textit{d-separated} by $S$ if \textit{all} paths between $A$ and $B$ are \textit{blocked} by $S$.
\end{mydef}

For the DAG in Figure \ref{fig:moral_graph}, the set $S=(X_4, X_5)$ d-separates the sets $A=(X_1,X_2,X_3)$ and $B=(X_6,X_7)$.

\subsection{Factorization, I-map, P-map and I-equivalence}
Next, we  relate a probability distribution to a DAG which leads to the definition of DAG models.  For each node $x_i$ of a DAG $\mathcal{G}$, we attach a random variable (also denoted by $x_i$) to it. We then consider a probability distribution $\mathbb{P}$ over the node set $\mathbb{V}=\{x_1,\cdots, x_p\}$.
%With a slight abuse of notation, we use the same symbol $\mathbb{V}=\{x_1,\cdots, x_p\}$ to denote  the set of nodes and the %corresponding set of random variables.

\begin{mydef}
A distribution $\mathbb{P}$ is said to admit a \textit{recursive factorization (DF)} according to a DAG $\mathcal{G}$ if it has a p.d.f. of the form:
$$
p(\textbf{x})=\prod_{i=1}^p p(x_i|pa_i^{\mathcal{G}}), ~~ \textbf{x}=(x_1, \cdots, x_p) \in \mathcal{X},
$$
where $p(x_i|pa_i^{\mathcal{G}})$ is the conditional probability density of $x_i$ given its parents $pa_i^{\mathcal{G}}$.
\end{mydef}
We  can now formally define DAG models.
\begin{mydef}
A \textit{DAG model} is a pair $(\mathcal{G}(\mathbb{V}, \mathbb{E}), \mathbb{P})$, where $\mathcal{G}$ is a DAG and $\mathbb{P}$ is  a probability distribution over $\mathbb{V}$ which admits recursive factorization according to $\mathcal{G}$.
\end{mydef}
By definition, in a DAG model, the distribution $\mathbb{P}$ is specified by a set of local conditional probability distributions, namely, $\{p(x_i | pa_i^{\mathcal{G}}): i=1,\cdots, p\}$.

The factorization property leads to a compact representation of $\mathbb{P}$ if the graph $\mathcal{G}$ is sparse. In order to get a characterization of its independence properties, we need the notion of Markov property.
\begin{mydef}
A distribution $\mathbb{P}$ is said to obey the \textit{directed global Markov property $(DG)$} according to a DAG $\mathcal{G}$ if  whenever  $A$ and $B$ are \textit{d-separated} by $S$ in $\mathcal{G}$, there is $A$ and $B$ conditionally independent given $S$ under $\mathbb{P}$ , denoted by   $A \independent B| S ~~~ [\mathbb{P}]$.
\end{mydef}

It turns out that, DF and DG are equivalent \cite{lauritzen1996graphical}.
\begin{prop}
\label{prop: DF-DG}
Let $\mathcal{G}$ be a DAG and  $\mathbb{P}$ be a probability distribution over its node set which has a density with respect to a product measure $\mu$. Then   $\mathbb{P}$ admits recursive factorization according to $\mathcal{G}$ if and only if   $\mathbb{P}$ obeys directed global Markov property according to $\mathcal{G}$.
\end{prop}

We can  now  try to characterize the independence properties of a distribution by a DAG. We need the notions of \textit{I-map} and \textit{P-map}.
\begin{mydef}
Let $\mathcal{I}(\mathbb{P})$ denote  all  conditional independence assentations of the form $X_A \independent X_B | X_S$ that hold under a distribution $\mathbb{P}$. Let $\mathcal{I}(\mathcal{G})$ be the set of conditional independence characterized by a DAG $\mathcal{G}$, i.e.,
$\mathcal{I}(\mathcal{G}):=\{X_A \independent X_B| X_S: \hbox{$S$ d-separates $A$ and $B$ in $\mathcal{G}$} \}$.  If
$\mathcal{I}(\mathcal{G}) \subseteq \mathcal{I}(\mathbb{P})$, then $\mathcal{G}$ is called an \textit{I-map} of $\mathbb{P}$.
\end{mydef}

 Proposition \ref{prop: DF-DG} says that, $\mathbb{P}$ factorizes according to $\mathcal{G}$ if and only if
$\mathcal{G}$ is an I-map of $\mathbb{P}$. It is obvious that, the factorization property does not totally characterize the independence properties of  a distribution. To see this, consider the distribution $\mathbb{P}$ under which $x_i$'s are mutually independent, then $\mathbb{P}$ factorizes according to any DAG (with nodes $x_i$'s). However, intuitively its ``corresponding" DAG should be the empty graph. We thus have the following definition.

\begin{mydef}
If $\mathcal{I}(\mathbb{P}) = \mathcal{I}(\mathcal{G})$, i.e., all independencies in $\mathbb{P}$  are reflected by the d-separation properties in $\mathcal{G}$ and vice versa,
then $\mathcal{G}$ is said to be a {\sl perfect map (P-Map)} for $\mathbb{P}$.
\end{mydef}

 Unfortunately, a P-map does not necessarily exist for a distribution $\mathbb{P}$. However, it can be shown that \cite{koller2009probabilistic}, for almost all distributions $\mathbb{P}$ that factorize over a DAG $\mathcal{G}$, we have $\mathcal{I}(\mathbb{P}) = \mathcal{I}(\mathcal{G})$.  On the other hand, there could be more than one P-maps for a distribution $\mathbb{P}$, as a set of independencies may be compatible to  multiple DAGs.

\begin{mydef}
   Two DAGs $\mathcal{G}$ and $\tilde{\mathcal{G}}$ over the same set of nodes are said to be \textsl{I-equivalent} if $\mathcal{I}(\mathcal{G})=\mathcal{I}(\mathcal{\tilde{G}})$, i.e., they encode the same set of independencies.
\end{mydef}

 We have the following characterization of I-equivalence \cite{Verma1991equivalence}.
\begin{prop}
\label{prop: I-equiv}
Two DAGs are I-equivalent if and only if they have the same set of skeleton edges and v-structures.
\end{prop}
An immediate consequence of Proposition \ref{prop: I-equiv} is that I-equivalent DAGs have the same moral graph (but the converse is not true). The I-equivalent relation partitions the DAG space into equivalence classes. In the following, we use $\mathcal{E}(\mathcal{G})$ to denote the equivalence class that $\mathcal{G}$ belongs to.

\begin{figure}[h]
\caption{Four DAGs:  $\mathcal{G}_1$ -- upper left,  $\mathcal{G}_2$ -- upper right, $\mathcal{G}_3$-- bottom left,  $\mathcal{G}_4$ -- bottom right. \label{fig:equiv}}
\begin{center}
\includegraphics[scale=0.5]{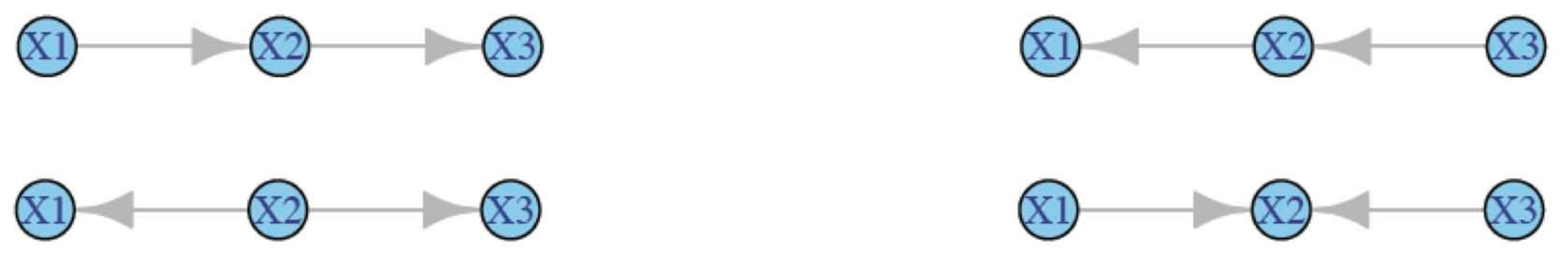}
\end{center}
\end{figure}

In Figure \ref{fig:equiv}, $\mathcal{G}_1,\mathcal{G}_2, \mathcal{G}_3$ are I-equivalent since they encode the same set of  independencies:
$\mathcal{I}=\{x_1 \independent x_3 \vert x_2\}$, as in all three graphs, the only d-separation property is:  $x_2$ d-separates $x_1$ from $x_3$.
  It is also clear that, they have the same moral graph:  $x_1 - x_2 - x_3$. On the other hand, these three DAGs are  not equivalent to $\mathcal{G}_4$ which has $\mathcal{I}=\{x_1 \independent x_3\}$ since in $\mathcal{G}_4$, $x_1$ and $x_3$ are de-separated by the empty set, but not by $x_2$. Its  moral graph is
$
\begin{array}{ccc}
x_1& - & x_3 \\
 \backslash & & /.  \\
& x_2&
\end{array}
$

It  is obvious that, it is  impossible to distinguish I-equivalent DAGs based solely on the distribution $\mathbb{P}$. The best one can hope for is to learn the I-equivalence class corresponding to $\mathbb{P}$.
Therefore, we  formally define the goal of DAG structure learning as follows.
\begin{mydef}
Given an i.i.d. sample $\mathbb{D}$ from a distribution $\mathbb{P}$, DAG structure learning is
 to recover the I-equivalence class of DAGs that are P-Maps    for $\mathbb{P}$ (assuming existence of P-maps).
\end{mydef}

\subsection{DAG structure learning}

In this subsection, we give a brief overview of popular DAG structure learning methods.
There are mainly three classes of methods for DAG learning, namely, \textit{score-based} methods, \textit{constraint-based} methods and \textit{hybrid} methods.

 In score-based methods, the DAG is viewed  as specifying a model and structure learning is approached   as a model selection problem.  Specifically,  a score-based method aims at  minimizing a pre-specified \textit{score} over the space of DAGs (defined on a given set of nodes). Commonly used scores include negative log-likelihood score, Akaike information criterion (AIC) score, Bayesian information criterion (BIC) score, Bayesian gaussian equivalent (BGe) score \cite{geiger1994learning}.

 As mentioned earlier, the DAG space is super-exponentially large (with respect to the number of nodes) and thus an exhaustive search for models with the optimal score is usually infeasible.  Therefore, greedy search algorithms are often employed.  One of the most popular search algorithms for DAG learning is the \textit{hill climbing algorithm}. At each search step, it conducts a local search among the graphs which are different from the current graph on a single edge \cite{russell2010artificial}. Other options include conducting the search in the space of I-equivalence classes of DAGs \cite{chickering2002learning} or using an ordering based algorithm \cite{teyssier2012ordering}.

 In constraint-based methods,  a DAG is viewed as a set of conditional independence constraints and
  the graph is learnt through conditional independence tests. One such method is the  \texttt{PC algorithm (PC-ALG)}  \cite{Verma1991equivalence, spirtes2001causation, kalisch2007estimating}. Other constraint-based methods  include  \texttt{Grow and Shrink (GS)} \cite{margaritis2003learning}, \texttt{Increament Association Markov Blanket (IAMB)} and its two variants \texttt{Fast IAMB and inter IAMB} \cite{yaramakala2005speculative}.
  %%% note: check citation on fast.iamb here

   In hybrid methods, certain local structures of  the graph such as the Markov blanket of each node are first learned, e.g., through independence tests and then such knowledge is used to   impose restrictions on the search space in a score-based method.   Hybrid methods  include \texttt{Max-Min Hill Climbing (MMHC)} \cite{tsamardinos2006max} and \texttt{L1MB} by \cite{schmidt2007learning}. %Another class of methods different from the above three concerns updating on the space of equivalent classes of DAGs \cite{chickering2002learning, chickering2003optimal}.

In this paper, we focus on score-based methods. We first develop an efficient implementation of the hill climbing search algorithm (Section \ref{sec:search}). We then propose aggregation strategies to improve the performance of DAG structure learning (Section \ref{sec:aggregation}). It should be noted that,  the proposed aggregation procedure \texttt{DAGBag} can be coupled with any DAG learning algorithms whether it is score-based or constraint-based or hybrid.

%We will also propose a hybrid method, where we use a sequence of learned moral graphs to constrain the search space in each step %(\texttt{spaceDAG}, Section \ref{sec:spaceDAG}).

\section{An Efficient Structure Learning Algorithm}
\label{sec:search}
In this section, we first discuss decomposable scores used in score-based methods and the hill climbing algorithm for searching DAGs with the optimal score. We  then propose an efficient implementation  which greatly speeds up the hill climbing search algorithm and thus makes high-dimensional DAG structure learning computationally feasible.

\subsection{Decomposable scores}
\label{subsec:score}
 In the rest of this paper, assume that we observe $n$ independently and identically distributed (i.i.d.) samples, $\mathbb{D}=\{X^1, \cdots, X^n\}$, from a $p$-dimensional multivariate distribution $\mathbb{P}$ with mean $\boldsymbol{\mu}$ and covariance $\Sigma$, where $X^k=(x^k_1, \cdots, x^k_p)^T$ denotes the $k^{th}$ sample ($k=1,\cdots, n$). We also assume that the distribution $\mathbb{P}$ has a P-map.
 As mentioned earlier, our goal  is to learn the I-equivalence class of DAGs  which are P-maps of $\mathbb{P}$.

 In the following, we use $X_i=\{x_i^1,\cdots, x_i^n\}$ to denote the data of node $x_i$ ($i=1,\cdots, p$). Moreover, we use $\mathbb{G}(\mathbb{V})$ to denote the space of DAGs defined on the node set $\mathbb{V}=(x_1,\cdots, x_p)$. We first introduce the notion of \textit{decomposable scores}.

 \begin{mydef}
 A score defined on the DAG space $\mathbb{G}(\mathbb{V})$ is called \textit{decomposable} if for any data $\mathbb{D}$ it has:
$$
score(\mathcal{G}: \mathbb{D})=\sum^p_{i=1} score(x_i|pa^\mathcal{G}_i:\mathbb{D}), ~~~ \mathcal{G} \in \mathbb{G}(\mathbb{V}),
$$
where $score_i:=score(x_i|pa^\mathcal{G}_i:\mathbb{D})$ only depends on the data of the node $x_i$ and  the data of its parent set $pa^\mathcal{G}_i$ ($i=1,\cdots p$). Hereafter, we refer to $score_i$ as the \textit{neighborhood score} of node $x_i$.
\end{mydef}
As we shall see in the next subsection, decomposable scores render efficient score updating in the search algorithm.

For the rest of this subsection,  we adopt the \textit{working assumption} that $\mathbb{P}$ is a multivariate normal distribution. We use $\theta$ to denote model parameters (in the Gaussian case, this includes the mean vector $\boldsymbol{\mu}$ and the covariance matrix $\Sigma$).
Then for a given DAG $\mathcal{G}$, by the factorization property, the negative maximum-log-likelihood -- the negative log-likelihood function evaluated at the restricted maximum likelihood estimator $\hat{\theta}_{\mathcal{G}}$ -- leads to a decomposable score:
\begin{eqnarray*}
score_{like}(\mathcal{G}: \mathbb{D})&=&-2\log \mathbb{P}_{\hat{\theta}_{\mathcal{G}}}(\mathbb{D})=\sum_{i=1}^p n \log(RSS_i/n),
\end{eqnarray*}
where $RSS_i=RSS(x_i|pa^\mathcal{G}_i: \mathbb{D})$ is the residual sum of squares by least-squares regression of $x_i$ onto $pa_i^\mathcal{G}$  ($i=1,\cdots p$).
%(Note, the latter equality holds only under Gaussianity.)

 Structure learning based on the likelihood score will over-fit the data since it always favors larger models -- distributions with less independence constraints which correspond to DAGs with more edges. Therefore, it is reasonable to consider scores that penalize for the model complexity. Two natural candidates are AIC and BIC which lead to the following decomposable scores:
\begin{eqnarray*}
score_{AIC}(\mathcal{G}: \mathbb{D})&=&-2\log \mathbb{P}_{\hat{\theta}_{\mathcal{G}}}(\mathbb{D})+ 2 |\mathbb{E}(\mathcal{G})|\\
&=&\sum_{i=1}^p \left ( n \log(RSS_i/n) + 2 |pa_i^\mathcal{G}|\right),
\end{eqnarray*}
where $|S|$ denotes the size of a set $S$, and

\begin{eqnarray*}
score_{BIC}(\mathcal{G}: \mathbb{D})&=&-2\log \mathbb{P}_{\hat{\theta}_{\mathcal{G}}}(\mathbb{D})+  |\mathbb{E}(\mathcal{G})| \log(n) \\
&=&\sum_{i=1}^p \left ( n \log(RSS_i/n) + |pa_i^\mathcal{G}| \log (n)\right).
\end{eqnarray*}

It can be shown that,  $score_{BIC}$ is \textit{model selection consistent} \cite{chickering2002optimal}, meaning that as sample size $n$ goes to infinity (and the number of nodes $p$ being fixed), the following holds with probability going to one: (i) if $\mathcal{G}^{*}$ is a P-map of $\mathbb{P}$, then it will minimize the score; (ii) any $\mathcal{\tilde{G}}$ that is not a P-map of $\mathbb{P}$ will have a strictly larger BIC value. Furthermore, $score_{BIC}$ is also \textit{locally consistent} \cite{chickering2002optimal}, meaning that the following holds with probability going to one: (i) the score decreases by adding an edge that eliminates an independence constraint that is not implied by $\mathbb{P}$; (ii) the score increases by adding an edge that does not eliminate such a constraint.  The local consistency property justifies the hill climbing search algorithm which conducts local updates of the model.

%%%%%
%%%note: need some reference here
%%%%%
\begin{comment}
It is known that, BIC is not a consistent model selection criteria when the model dimension $p$ diverges with $n$ sufficiently fast. Recently, various information criteria are proposed for consistent model selection under the large $p$ case by  adapting to the size of the model space. In this thesis, we thus also consider scores based on high-dimensional information criteria.
Two such examples are the extended Bayesian information criteria (eBIC) \cite{chen2008extended} and the generalized information criteria (GIC)  \cite{kim2012consistent}, which lead to the following scores,
\begin{eqnarray*}
\text{score}_{eBIC} (\mathcal{G}: \mathbb{D})&=&\sum_{i=1}^p  \left(n\log(RSS_i/n)+|pa^\mathcal{G}_i| \left(\log(n)+2\log(p)\right) \right)\\
\text{score}_{GIC} (\mathcal{G}: \mathbb{D})&=&\sum_{i=1}^p \left(n\log(RSS_i/n)+|pa^\mathcal{G}_i| \left(\log\log(n)\right) \log(p)\right).
\end{eqnarray*}
\end{comment}

Under the Bayesian framework, various scores based on posterior probabilities of DAG have been proposed. These include  the
Bayesian Gaussian equivalent (BGe) score  for multivariate normal distributions \cite{geiger1994learning}.
%We will compare the aforementioned scores in Chapter  \ref{chp:simulation} through numerical studies.
We will compare different scores in terms of their performance in DAG learning in Section \ref{sec:numerical}.
%%and Bayesian Dirichlet equivalent (BGe) score for multinomial distributions

\begin{comment}
Except for decomposability, another desirable property for a score is to have the same value on DAGs in the same I-equivalence class.
\begin{mydef}
A score $S(\mathcal{G}: \mathbb{D})$ is called score equivalent  if whenever two DAGs $\mathcal{G}, \mathcal{\tilde{G}}$ are I-equivalent, we have $S(\mathcal{G}: \mathbb{D})=S(\mathcal{\tilde{G}}: \mathbb{D})$ for all data $\mathbb{D}$.\
\end{mydef}
With score equivalence,  if we move to a graph with a better score, we are necessarily move into a different I-equivalence class. All aforementioned scores satisfy score equivalence.
\end{comment}
%%%%%%%%%%%%%%%%%%%%%%%%%%%%%%%%%%%%%%%
%%%%%%%%%%%%%%%%%%%%%%%%%%%%%%%%%%%%%%%%
\subsection{Hill climbing algorithm} %literature review
\label{subsec:hc}

Given  a score function, we want to find a DAG $\mathcal{G}^{\ast}$ that has the minimum score value among all DAGs, i.e.,
$$
\mathcal{G}^{\ast}=\min_{\mathcal{G} \in \mathbb{G}(\mathbb{V})} score(\mathcal{G}:\mathbb{D}).
$$
As mentioned earlier, due to the sheer size of the DAG space, an exhaustive search is infeasible for even moderate number of nodes. Therefore, heuristic search is often employed. A popular algorithm is the \textit{hill climbing algorithm} which is a greedy iterative search algorithm. It starts with an empty graph (or an user given initial graph). At each subsequent step, a best operation, the one among the set of all eligible operations  that results in the maximum reduction of the score, is used to update the current graph. The search stops when no operation is able to decrease the score anymore, which leads to a (local) optimal solution.

In the hill climbing algorithm for DAG learning, an \textit{operation} is defined as either the addition of an absent edge, or the deletion of an existing edge, or the reversal of an existing edge. Moreover, an operation is \textit{eligible} only if it does not result in cycles in the graph.  This algorithm is outlined in Table \ref{table:hc}.

 \begin{table}
\caption{Hill climbing algorithm for DAG structure learning. \label{table:hc}}
\begin{center}
\begin{tabular}{l}
\hline\hline
 {\bf Input:} data $\mathbb{D}$, node set $\mathbb{V}$, score function $score(\cdot:\cdot)$. \\
\hspace{5pt} \underline{Initial step}: $\mathcal{G}^{(0)}=$ empty graph; for $i=1,\cdots, p$, calculate $score_i(\mathcal{G}^{(0)}: \mathbb{D})$. \\

\hspace{5pt} \underline {$s^{th}$ Step} : Current graph $\mathcal{G}^{(s)}$.\\
\hspace{10pt} $\bullet$ {\it Acyclic check}: for each potential operation, check whether applying it to   $\mathcal{G}^{(s)}$ results in cycle or not.\\
\hspace{10pt} $\bullet$ {\it Score update}: for each eligible operation $O$, \\
\hspace{15pt} calculate the score change $\delta^{(s)}(O, \mathbb{D}):=score(O(\mathcal{G}^{(s)}):\mathbb{D})-score(\mathcal{G}^{(s)}: \mathbb{D})$, \\
\hspace{15pt}  where $O(\mathcal{G}^{(s)})$ denotes the graph resulting from applying $O$ to $\mathcal{G}^{(s)}$.\\
\hspace{10pt} $\bullet$ {\it Graph update}: calculate  $\delta^{(s)}_{\min}:=\min_{O} \delta^{(s)}(O, \mathbb{D})$. \\

\hspace{20pt} $\bullet$ If $\delta^{(s)}_{\min} \geq 0$, stop the search algorithm and set  $\mathcal{G}^{\ast}=\mathcal{G}^{(s)}$.\\

\hspace{20pt} $\bullet$ Otherwise,  choose the operation $O^{(s)}$ corresponding to $\delta^{(s)}_{\min}$. \\
\hspace{30pt} $\bullet$  Set  $\mathcal{G}^{(s+1)}=O^{(s)}(\mathcal{G}^{(s)})$. \\
 \hspace{30pt} $\bullet$  Proceed to step $s+1$. \\
{\bf Output:} $\mathcal{G}^{\ast}$, the list of best operations $O^{(s)}$'s, the corresponding list of score decreases $\delta^{(s)}_{\min}$'s.  \\ \hline\hline
\end{tabular}
\end{center}
\end{table}

The major computational cost of the hill climbing algorithm is from the acyclic check for each potential operation (i.e., check whether the operation leads to cycles or not) and calculating score change for each eligible operation at every search step.
Note that,  by definition, an operation only results in the change in one (for addition and deletion operations) or two (for reversal operations) neighborhoods of  the current graph. Consequently, for a decomposable score, an operation only changes the neighborhood score for up to two nodes. Therefore, the score updating step for each operation is computationally cheap as we only need to calculate the change of score for the neighborhood(s) involved in this operation. For instance, if operation $O$ is to add an edge $x_i \rightarrow x_j$, then applying $O$ on graph $\mathcal{G}$ leads to:
\begin{eqnarray*}
\delta(O, \mathbb{D})=score(O(\mathcal{G}):\mathbb{D})-score(\mathcal{G}: \mathbb{D})=score_j(O(\mathcal{G}):\mathbb{D})-score_j(\mathcal{G}: \mathbb{D}),
\end{eqnarray*}
as only the neighborhood of $x_j$ is changed by operation $O$.

Nevertheless, even with decomposable scores, under the high-dimensional setting, acyclic check and score updating are costly due to  the large number of potential operations ($\sim O(p^2)$) at each search step.  Moreover, when $p$ is large, it often takes many steps until the algorithm stops (i.e., to reach $\delta_{\min} \geq 0$). In the next subsection, we discuss efficient updating schemes which greatly speed up the search algorithm such that it may be used for high-dimensional DAG models learning.

%%%%%%%%%%%%%%%

\subsection{Efficient score updating and  acyclic check}
In this subsection, we describe an efficient implementation of the hill climbing search algorithm for decomposable scores, where information from the previous step is re-used to facilitate both score updating and acyclic check in the current step.

We illustrate the idea by an example. Suppose the best operation selected in the previous step is to add $x_i \rightarrow x_j$. Then at the current step, the following holds: (i) any operation that does not involve neighborhood of $x_j$ results in the same score change as in the previous step; (ii) any operation that results in cycles  in the previous step still leads to cycles;  (iii) an operation that does not result in cycles in the previous step may only lead to cycles in the current step through $x_i \rightarrow x_j$. Therefore acyclic check can be performed very efficiently for such operations.

In the following, we use $\mathcal{\tilde{G}}$ to denote the graph in the previous step,
$O^{\ast}$ to denote the selected operation by the previous step, and $\mathcal{G}^{\ast}=O^{\ast}(\mathcal{\tilde{G}})$ to denote the current graph.  We also use $\delta(O: \mathcal{G})$ to denote the change of score resulting from applying operation $O$ to graph $\mathcal{G}$. We summarize the score updating and acyclic check schemes in the following two propositions.

\begin{prop}\label{prop:score_update} Suppose $score(\cdot: \cdot)$ is a decomposable score.
For an (eligible) operation $O$, $\delta(O: \mathcal{\tilde{G}})=\delta(O: \mathcal{G}^{\ast})$, if one of the following holds:
\begin{itemize}
\item $O^{\ast}$  is one of the forms: ``add $x_i \rightarrow x_j$" or ``delete $x_i \rightarrow x_j$",  and $O$ is not one of the forms: ``add $x_k \rightarrow x_j$" , ``delete $x_k \rightarrow x_j$" , `` reverse $x_j \rightarrow x_k$ ".
    \item $O^{\ast}$ is of the form ``reverse $x_i \rightarrow x_j$", and $O$ is not one of the forms: ``add $x_k \rightarrow x_j$" , ``delete $x_k \rightarrow x_j$" , `` reverse $x_j \rightarrow x_k$ ", ``add $x_k \rightarrow x_i$" , ``delete $x_k \rightarrow x_i$",  `` reverse $x_i \rightarrow x_k$ ".
\end{itemize}
\end{prop}
In short, any operation that does not involve the neighborhoods changed by the selected operation in the previous step will lead to the same score change as in the previous step. This has been pointed out by \cite{koller2009probabilistic}.

%\begin{prop}
%Let $G$ and $G'$ be two DAGs and score be a decomposable score and $\delta(G:o)$ denotes the delta of score for operation $o$ on the %graph $G$.
%\begin{itemize}
%\item If $o$ is either "Add $i\rightarrow j$" or "Delete $i \rightarrow j$", and $Pa_j^G=Pa_j^{G'}$, then $\delta %(G:o)=\delta(G':o)$
%\item If $o$ is "Reverse $i\rightarrow j$", and $Pa_j^G=Pa_j^{G'}$ and $Pa_i^G=Pa_i^{G'}$, then $\delta (G:o)=\delta(G':o)$
%\end{itemize}
%\end{prop}
%

In the following, $de(x_i)$ denotes the set of descendent of node $x_i$ and $an(x_i)$ denotes the set of ancestors of node $x_i$.

\begin{prop} \label{prop:acyclic} The following holds for acyclic check.
\begin{itemize}
\item If $O^{\ast}$ is of the form ``add $x_{i^{\ast}} \rightarrow x_{j^{\ast}}$", then for an operation $O$ the  following holds:

\begin{itemize}
\item If $O$ does not lead to cycles in the previous step, then

\begin{itemize}
\item if $O$ is of the form ``add $x_i \rightarrow x_j$", and $i \in de(x_{j^\ast})$  and $j \in an(x_{i^\ast})$, then $O$ leads to a cycle.
\item if $O$ is of the form ``reverse  $x_i \rightarrow x_j$", and   $j \in de(x_{j^\ast})$  and $i \in an(x_{i^\ast})$, then $O$ leads to a cycle.
    \item if otherwise, $O$ remains acyclic.
\end{itemize}
\item If $O$ leads to a cycle in the previous step, it remains cyclic.

\end{itemize}

\item If $O^{\ast}$ is of the form ``delete $x_{i^{\ast}} \rightarrow x_{j^{\ast}}$",  then for an operation $O$ the  following holds:

\begin{itemize}
\item If $O$ does not lead to cycles in the previous step, it remains acyclic.
\item If $O$ leads to a cycle in the previous step, then
\begin{itemize}
\item if $O$ is of the form ``add $x_i \rightarrow x_j$", and $i \in de(x_{j^\ast})$  and $j \in an(x_{i^\ast})$, then we  need to check its acyclicity;
\item if $O$ is of the form ``reverse  $x_i \rightarrow x_j$", and  $j \in de(x_{j^\ast})$  and $i \in an(x_{i^\ast})$, then we need to check its acyclicity;
\item if otherwise,  $O$ remains cyclic.

\end{itemize}
\end{itemize}

\item If $O^{\ast}$ is of the form ``reverse $x_{i^{\ast}} \rightarrow x_{j^{\ast}}$", then for an operation $O$ the  following holds:
\begin{itemize}
\item If $O$ does not lead to cycles in the previous step, then
\begin{itemize}
\item if $O$ is of the form ``add $x_i \rightarrow x_j$", and  $i \in de(x_{i^\ast})$  and $j \in an(x_{j^\ast})$, then $O$ leads to a cycle.
\item if  $O$ is of the form ``reverse  $x_i \rightarrow x_j$", and $j \in de(x_{i^\ast})$  and $i \in an(x_{j^\ast})$, then $O$ leads to a cycle.
    \item  if otherwise,  $O$ remains acyclic.
\end{itemize}
\item If $O$ leads to a cycle in the previous step, then
\begin{itemize}
\item if $O$ is of the form ``add $x_i \rightarrow x_j$", and $i \in de(x_{j^\ast})$  and $j \in an(x_{i^\ast})$, then we  need to check its acyclicity;
\item if  $O$ is of the form ``reverse  $x_i \rightarrow x_j$", and $j \in de(x_{j^\ast})$  and $i \in an(x_{i^\ast})$, then we  need to check its acyclicity;
\item if otherwise, $O$ remains cyclic.
  \end{itemize}
\end{itemize}

\end{itemize}
\end{prop}

 It is clear from the above two propositions that at each search step, only for a fraction of  operations we need to calculate their score change and check the acyclic status. For the majority of operations, score change and acyclic status remain the same as those in the previous step and thus do not need to be re-assessed.

 The implementation with these efficient updating schemes greatly speeds up the search algorithm. For example, for a graph with $p=1000$ nodes, sample size $n=250$, conducting $2000$ search steps takes $\sim 150$ seconds under our implementation and $\sim 2,500$ seconds under the \texttt{hc} function in \texttt{bnlearn} R package \cite{scutari2009learning} (version 3.3, published on 2013-03-05). The above numbers are user times averaged over $10$ simulation replicates performed on a machine with 8GB memory and 2quad CPU.

We defer additional implementation details including \textit{early stopping} and \textit{random restart} into the Appendix (Section \ref{app:alg}).

\section{Bootstrap Aggregating for DAG Learning}
\label{sec:aggregation}

In this section, we propose the \texttt{DAGBag} procedure which is inspired by \textit{bootstrap aggregating -- Bagging}  \cite{Breiman96}.
Bagging  is originally proposed to get an aggregated prediction rule based on multiple versions of prediction rules built on bootstrap resamples of the data. It is most effective in improving highly variable learning procedures such as classification trees through variance reduction. Recently, the aggregation idea has been applied  to learn stable structures in high-dimensional regression models and Gaussian graphical models \cite{meinshausen2010stability,wang2011random, li2011bootstrap}.

\subsection{\texttt{DAGBag}}
\label{subsec:DAGbag}
As with many structure learning procedures, DAG learning procedures  are  usually highly variable -- the learned graph changes drastically with even small perturbation of the data. Therefore, we propose to utilize the aggregation idea in DAG learning to search for stable structures. In particular, through aggregation we are able to alleviate over-fitting and greatly reduce the number of  false positive edges.

In \texttt{DAGbag}, an ensemble of DAGs $\mathbb{G}^{e}=\{\mathcal{G}_b\}_{b=1}^B$ are first learned based on bootstrap resamples $\{\mathbb{D}_b\}_{b=1}^B$ of the original data $\mathbb{D}$ by a DAG learning procedure, e.g., a score-based method. Then an aggregated DAG is learned based on this ensemble. Aggregation of a collection of DAGs is nontrivial because the notion of average is not straightforward on the DAG space. Here, we generalize the idea of median by searching for a DAG  that minimizes an average distance to the DAGs in the ensemble:

$$
\mathcal{G}^*={\rm argmin}_{\mathcal{G} \in \mathbb{G}(\mathbb{V})}score_{d}(\mathcal{G}: \mathbb{G}^{e}), $$
where  $$score_{d}(\mathcal{G}: \mathbb{G}^{e}):= \frac{1}{B}\sum^B_{b=1} d(\mathcal{G},\mathcal{G}_b)$$
with $d(\cdot,\cdot)$ being a distance matric on the DAG space.  As a general recipe, the aggregated DAG $\mathcal{G}^\ast$ may be learned by the hill climbing search algorithm.  However, as shown in the next subsection,  for the family of distance metrics based on the structural hamming distance, the search can be conducted in a much more  efficient manner.

The  \texttt{DAGBag} procedure is outlined in Table  \ref{table:DAGBag}. %%`Fr$\acute{e}$chet mean"
 \begin{table}
\caption{\texttt{DAGBag} procedure. \label{table:DAGBag}}
\begin{center}
\begin{tabular}{l} \hline\hline
 {\bf Input:} data $\mathbb{D}$, node set $\mathbb{V}$,  DAG learning procedure $\mathcal{L}$, distance measure  $d(\cdot,\cdot)$  on the DAG space $\mathbb{G}(\mathbb{V})$, \\
\hspace{30pt}  number of bootstrap samples $B$.  \\
\hspace{10pt} \underline{Bootstrapping}:  obtain $B$ bootstrap samples of the data: $\mathbb{D}_1,\cdots, \mathbb{D}_B$.\\
\hspace{10pt} \underline{Ensemble of DAGs}:  for $b=1,\cdots, B$, learn a DAG $\mathcal{G}_b$ based on  resample $\mathbb{D}_b$  by learning procedure $\mathcal{L}$.\\
\hspace{50pt} Obtain an ensemble of DAGs: $ \mathbb{G}^{e}:=\{\mathcal{G}_1, \cdots, \mathcal{G}_B\}$.\\

\hspace{10pt}\underline{Aggregation}: learn  an aggregated DAG:\\
\hspace{50pt} $
\mathcal{G}^*={\rm argmin}_{\mathcal{G} \in \mathbb{G}(\mathbb{V})}score_{d}(\mathcal{G}: \mathbb{G}^{e})$, where  $score_{d}(\mathcal{G}: \mathbb{G}^{e}):= \frac{1}{B}\sum^B_{b=1} d(\mathcal{G},\mathcal{G}_b)$.
 \\
 {\bf Output:} aggregated DAG $\mathcal{G}^*$. \\
\hline\hline
\end{tabular}
\end{center}
\end{table}

\begin{comment}
In the \texttt{DAGBag} procedure, the default learning procedure $\mathcal{L}$ is set to be the BIC-score based method as described in Section \ref{sec:search}. The resamples are obtained through random sampling  with replacement of the $n$ samples. The number of resamples $B$ is set to be $100$. In Chapter \ref{chp:simulation}, we study the effect of number of resamples and find that $B=100$ is sufficient for most cases. However, for challenging scenarios such as when the graph is relatively dense, a larger number of $B$ (say $500$) is beneficial.
As a general recipe, the aggregated DAG $\mathcal{G}^\ast$ may be learned by the hill climbing search algorithm.  As shown in the next subsection,  for some distance measures, the search may be conducted in a very  efficient manner.
\end{comment}

\subsection{Structural Hamming distances and aggregation scores}
\label{subsec:shd}
One crucial aspect of the \texttt{DAGBag} procedure is the distance metric $d(\cdot, \cdot)$ on DAG space which dictates how the ensemble of DAGs should be aggregated.
In this subsection, we discuss distance metrics based on the \texttt{structural Hamming distance (SHD)} and the properties of their corresponding aggregation scores.

In information theory, the \textit{Hamming distance} between two 0-1 vectors of equal length is the minimum number of substitutions needed to convert one vector to another.  This can be generalized to give a distance measure between DAGs with the same set of nodes.
\begin{mydef}
A structural Hamming distance between $\mathcal{G}, \mathcal{\tilde{G}} \in \mathbb{G}(\mathbb{V})$ is defined as :
$$
d(\mathcal{G}, \mathcal{\tilde{G}}):=\text{the minimum number of operations needed to covert $\mathcal{G}$ to $ \mathcal{\tilde{G}}$.}
$$
\end{mydef}
It is obvious that, this definition leads to valid distance measures as: (i) $d(\mathcal{G}, \mathcal{\tilde{G}}) \geq 0$, ``$=0$" if and only if $\mathcal{G} = \mathcal{\tilde{G}}$; (ii) $d(\mathcal{G}, \mathcal{\tilde{G}})=d(\mathcal{\tilde{G}}, \mathcal{G})$; (iii) $d(\mathcal{G}, \mathcal{\tilde{G}}) \leq d(\mathcal{G}, \mathcal{G}^{\ast})+d(\mathcal{G}^{\ast}, \mathcal{\tilde{G}})$.

As in the hill climbing search algorithm, we may define an operation as,  the addition of an absent edge, or the  deletion of an existing edge,  or the reversal of an existing edge, which does not lead to cycles. In this subsection, we propose variants of SHD on DAG space where the difference among these variants lies in  how reversal operations are counted.

Specifically,  the addition or the deletion of an edge is always counted as one unit of operation. If the reversal of an edge is counted as two units of operations, then we have the following distance measure:
 \begin{eqnarray*}
 d_{SHD}(\mathcal{G},\mathcal{\tilde{G}})&=&\sum_{i=1} ^p \sum_ {j=1} ^p |\mathbb{A}(i,j)-\mathbb{\tilde{A}}(i,j)| = \sum_{1 \leq i<j \leq p} |\mathbb{A}(i,j)-\mathbb{\tilde{A}}(i,j)| + |\mathbb{A}(j,i)-\mathbb{\tilde{A}}(j,i)| , \\
 &=&\sum_{i=1} ^p \sum_ {j=1} ^p \left(\mathbb{A}(i,j)-\mathbb{\tilde{A}}(i,j)\right)^2,
\end{eqnarray*}
where $\mathbb{A}, \mathbb{\tilde{A}}$ denote the adjacency matrices of $\mathcal{G}$ and $\mathcal{\tilde{G}}$, respectively.
Note that $d_{SHD}(\mathcal{G},\mathcal{\tilde{G}})$ is both the $\ell_1$ distance and the $\ell_2$ distance between the two adjacency matrices.

If the reversal of an edge  is also counted as one unit of operation, then we have the following distance measure:
\begin{eqnarray*}
d_{adjSHD}(\mathcal{G},\mathcal{\tilde{G}})=\sum_{1 \leq i<j \leq p} \max\{|\mathbb{A}(i,j)-\mathbb{\tilde{A}}(i,j)|, |\mathbb{A}(j,i)-\mathbb{\tilde{A}}(j,i)|\}.
\end{eqnarray*}

More generally, one may count the reversal of an edge as $\alpha(> 0)$ units of operations which leads to the following family of distances.
 \begin{mydef}
 \label{def:gshd}
 For $\alpha>0$, the generalized structural Hamming distance $d_{GSHD(\alpha)}$ on the DAG space $\mathbb{G}(\mathbb{V})$ is defined as:
\begin{eqnarray*}
d_{GSHD(\alpha)}(\mathcal{G},\mathcal{\tilde{G}})=\sum_{1 \leq i<j \leq p} S_{ij}(\mathbb{A}, \mathbb{\tilde{A}}; \alpha), ~~ \mathcal{G},\mathcal{\tilde{G}} \in \mathbb{G}(\mathbb{V}),
\end{eqnarray*}
where $S_{ij}(\mathbb{A}, \mathbb{\tilde{A}}; \alpha)$ is defined as following:
\begin{itemize}
\item If $|\mathbb{A}(i,j)-\mathbb{\tilde{A}}(i,j)| + |\mathbb{A}(j,i)-\mathbb{\tilde{A}}(j,i)|=0$, then $S_{ij}(\mathbb{A}, \mathbb{\tilde{A}}; \alpha)=0$;
    \item If $|\mathbb{A}(i,j)-\mathbb{\tilde{A}}(i,j)| + |\mathbb{A}(j,i)-\mathbb{\tilde{A}}(j,i)|=1$, then $S_{ij}(\mathbb{A}, \mathbb{\tilde{A}}; \alpha)=1$;
        \item If $|\mathbb{A}(i,j)-\mathbb{\tilde{A}}(i,j)| + |\mathbb{A}(j,i)-\mathbb{\tilde{A}}(j,i)|=2$, then $S_{ij}(\mathbb{A}, \mathbb{\tilde{A}}; \alpha)=\alpha$.
\end{itemize}
\end{mydef}

It is easy to see that, $d_{SHD}$ corresponds to $\alpha=2$ and $d_{adjSHD}$ corresponds to $\alpha=1$.   In the literature, $d_{SHD}$ and $d_{GSHD(0.5)}$ have been used to compare a learned DAG to the ``true" DAG as performance evaluation criteria
 of DAG learning procedures in numerical studies \cite{tsamardinos2006max,perrier2008finding}.

In $d_{SHD}$, a reversely oriented direction  is penalized twice as much as a missing or an extra skeleton edge. Since edge directions are not always identifiable, it is reasonable to penalize  a reversely oriented direction less severely. This is supported by our numerical results where aggregation based on $d_{adjSHD}$ (or $d_{GSHD(\alpha)}$ with an $\alpha<2$) usually outperforms aggregation based on $d_{SHD}$. In general, with larger $\alpha$, the aggregated DAG retains less edges and tends to have smaller false positive edges as well as less correct edges.

In the following, we use \texttt{score.SHD}, \texttt{score.adjSHD} and \texttt{score.GSHD$(\alpha)$} to denote aggregation scores based on $d_{SHD}$, $d_{adjSHD}$ and $d_{GSHD(\alpha)}$, respectively.

In Proposition \ref{prop:score_shd} below, we derive an expression of \texttt{score.SHD} in terms of edge \textit{selection frequencies (SF)}.
\begin{mydef}\label{def:sf}
Given an ensemble of DAGs: $\mathbb{G}^e=\{\mathcal{G}_b: b=1,\cdots B\}$,  the selection frequency (SF) of a directed edge $e$ is defined as
$$
p_e:= \#\{\mathcal{G}_b: e \in \mathbb{E}(\mathcal{G}_b)\}/B,$$
where $\mathbb{E}(\mathcal{G}_b)$ denotes the edge set of DAG $\mathcal{G}_b$.
\end{mydef}

\begin{prop}\label{prop:score_shd}
Given an ensemble of DAGs: $\mathbb{G}^e=\{\mathcal{G}_b: b=1,\cdots B\}$,  the aggregation score under $d_{SHD}$ is
\begin{eqnarray*}
score.SHD(\mathcal{G}: \mathbb{G}^e)
= \sum_{e \in \mathbb{E}(\mathcal{G})}(1-2p_e)+C,
\end{eqnarray*}
where
$$
C=\frac{1}{B}\sum_{b=1}^B \sum_{i=1}^p \sum_{j=1}^p \mathbb{A}_b(i,j)=\sum_{i=1}^p \sum_{j=1}^p p_{x_i \rightarrow x_j},
$$
is a constant which only depends on the ensemble $\mathbb{G}^e$, but does not depend on $\mathcal{G}$.
\end{prop}
By Proposition \ref{prop:score_shd}, $score.SHD$ is \textit{super-decomposable} in that it is an additive function of the selection frequencies of individual edges.  Moreover, the DAG $\mathcal{G}^\ast$ that minimizes \texttt{score.SHD} could only contain edges with selection frequency larger than $50\%$ and it should contain as many such edges  as possible. Indeed, as shown by Proposition \ref{prop:score_SHD_hc}, the hill climbing algorithm is simplified to the algorithm described in Table \ref{table:agg_score_SHD}.

\begin{table}
\caption{Hill climbing algorithm with aggregation \texttt{score.SHD}. \label{table:agg_score_SHD}}
\begin{center}
\begin{tabular}{l}
\hline\hline
 {\bf Input:}  an ensemble of DAGs: $\mathbb{G}^e=\{\mathcal{G}_b: b=1,\cdots B\}$.\\
\hspace{5pt} \underline{Calculate selection frequency} for all possible edges.\\
\hspace{5pt} \underline{Order edges} with selection frequency $>50\%$. \\
\hspace{5pt} \underline{Add edges} sequentially  according to SF and stop when SF $\leq 0.5$. \\
\hspace{10pt} $\bullet$ Initial step: $\mathcal{G}^{(0)}$= empty graph, $\mathbb{C}=$ empty set.  \\
\hspace{10pt} $\bullet s^{th}$ step: current graph $\mathcal{G}^{(s)}$, current operation $O$: ``add the edge with the $s^{th}$ largest SF". \\
\hspace{20pt} $\bullet$ If $O$ passes the acyclic  check, then $\mathcal{G}^{(s+1)}=O(\mathcal{G}^{(s)})$, i.e., add this edge. \\
\hspace{30pt} If $O$ does not pass the  acyclic check, then $\mathcal{G}^{(s+1)}=\mathcal{G}^{(s)}$, i.e., does not add this edge,\\
 \hspace{30pt} and add this edge to  $\mathbb{C}$.\\
\hspace{20pt} $\bullet$ If the $(s+1)^{th}$ largest SF $> 50\%$, proceed to step $s+1$.  \\
\hspace{30pt} Otherwise, set $\mathcal{G}^{\ast}=\mathcal{G}^{(s+1)}$ and stop the algorithm. \\
{\bf Output:} $\mathcal{G}^{\ast}$, and the set of ``cyclic edges" $\mathbb{C}$. \\ \hline\hline
\end{tabular}
\end{center}
\end{table}

\begin{prop}\label{prop:score_SHD_hc}
In each step of the hill climbing algorithm with \texttt{score.SHD}, the following operations (assuming eligible) will not decrease the score:
\begin{itemize}
\item add an edge with selection frequency $\leq 0.5$;
\item delete an edge in the current graph;
\item reverse an edge in the current graph.
\end{itemize}
Moreover, among eligible edge addition operations, the one corresponding to the largest selection frequency will lead to the most decrease of \texttt{score.SHD}. Therefore, the hill climbing search algorithm can be conducted as described in Table \ref{table:agg_score_SHD}.

\end{prop}

 In addition, it is shown in Proposition \ref{prop:score_SHD_alg} that, the algorithm in Table  \ref{table:agg_score_SHD} leads to a global optimal solution  if the set of ``cyclic edges" $\mathbb{C}$ (defined in Table \ref{table:agg_score_SHD}) has at most one edge.
\begin{prop}\label{prop:score_SHD_alg}
Given an ensemble of DAGs: $\mathbb{G}^e=\{\mathcal{G}_b: b=1,\cdots B\}$ defined on node set $\mathbb{V}$, the DAG $\mathcal{G}^{\ast}$ obtained by the  algorithm in Table \ref{table:agg_score_SHD} reaches the minimum \texttt{score.SHD} value:
$$
score.SHD(\mathcal{G}^\ast: \mathbb{G}^e)= \min_{\mathcal{G} \in \mathbb{G}(\mathbb{V})} score.SHD(\mathcal{G}: \mathbb{G}^e),
$$
provided that $|\mathbb{C}| \leq 1$.
\end{prop}
The proofs of Propositions \ref{prop:score_shd}, \ref{prop:score_SHD_hc}  and \ref{prop:score_SHD_alg} are given in the Appendix (\ref{app:proof}).
%%%%%%%%%%%%%%%%%%%%%%%%%%%%%%%%%%%%%%%%%%%%%%%%
%%%%%%% for our simulations, most of time |C|=0, and occasionally |C|=1. Have not yet observed |C|>1.
%%%%%%%%%%%%%%%%%%%%%%%%%%%%%%%%%%%%%%%%%%%

%%%%%%%%%%%%%%%%%%%%%%%GSHD
All the above results for \texttt{score.SHD} can be generalized to \texttt{score.GSHD$(\alpha)$}. For this purpose, we need the definition of \textit{generalized selection frequency}.
\begin{mydef}
\label{def:gsf}
Given an ensemble of DAGs: $\mathbb{G}^e=\{\mathcal{G}_b: b=1,\cdots B\}$ and an $\alpha>0$,  the generalized selection frequency (GSF) of a directed edge $e$ is defined as
$$
gp_e(\alpha):=p_e+(1-\frac{\alpha}{2})p_{e^*},
$$
where $e^{*}$ denotes the edge with the reversed direction of $e$.
\end{mydef}

\texttt{score.GSHD} can be expressed in terms of generalized selection frequencies.
\begin{prop}\label{prop:score_gshd}
Given an ensemble of DAGs: $\mathbb{G}^e=\{\mathcal{G}_b: b=1,\cdots B\}$ and an $\alpha>0$,  the aggregation score under $d_{GSHD(\alpha)}$ is
\begin{eqnarray*}
score.GSHD(\alpha)(\mathcal{G}: \mathbb{G}^e)
= \sum_{e \in \mathbb{E}(\mathcal{G})}(1-2gp_e(\alpha))+C,
\end{eqnarray*}
where
$$
C=\frac{1}{B}\sum_{b=1}^B \sum_{i=1}^p \sum_{j=1}^p \mathbb{A}_b(i,j)=\sum_{i=1}^p \sum_{j=1}^p p_{x_i \rightarrow x_j},
$$
is a constant which only depends on the ensemble $\mathbb{G}^e$, but does not depend on $\mathcal{G}$.
\end{prop}

Also the hill climbing search algorithm with \texttt{score.GSHD} can be simplified to the procedure described in Table \ref{table:agg_score_GSHD} (Proposition \ref{prop:score_GSHD_hc}).
Note that, only when $\alpha=2$,  the GSF  of an edge $e$ does not depend on the SF of the reversed edge $e^{*}$, and the search under the corresponding score \texttt{score.SHD} only depends on the SF of individual edges. For $\alpha$ other than $2$, the search depends on the SF of pairs of reversely oriented edges through GSF. In particular, the larger $\alpha$ is, the more  an edge would be penalized  by the extent of how frequently the reversed edge appeared in the ensemble.
\begin{table}
\caption{Hill climbing algorithm with aggregation \texttt{score.GSHD$(\alpha)$}. \label{table:agg_score_GSHD}}
\begin{center}
\begin{tabular}{l}
\hline\hline
 {\bf Input:}  an ensemble of DAGs: $\mathbb{G}^e=\{\mathcal{G}_b: b=1,\cdots B\}$, an $\alpha>0$.\\
\hspace{5pt} \underline{Calculate generalized selection frequency $gp_e(\alpha)$} for all possible edges.\\
\hspace{5pt} \underline{Order edges} with generalized selection frequency $>50\%$. \\
\hspace{5pt} \underline{Add edges} sequentially  according to GSF and stop when GSF $\leq 0.5$. \\
\hspace{10pt} $\bullet$ Initial step: $\mathcal{G}^{(0)}$= empty graph, $\mathbb{C}=$ empty set.  \\
\hspace{10pt} $\bullet s^{th}$ step: current graph $\mathcal{G}^{(s)}$, current operation $O$: ``add the edge with the $s^{th}$ largest GSF". \\
\hspace{20pt} $\bullet$ If $O$ passes the acyclic  check, then $\mathcal{G}^{(s+1)}=O(\mathcal{G}^{(s)})$, i.e., add this edge. \\
\hspace{30pt} If $O$ does not pass the  acyclic check, then $\mathcal{G}^{(s+1)}=\mathcal{G}^{(s)}$, i.e., does not add this edge,\\
 \hspace{30pt} and add this edge to  $\mathbb{C}$.\\
\hspace{20pt} $\bullet$ If the $(s+1)^{th}$ largest GSF $> 50\%$, proceed to step $s+1$.  \\
\hspace{30pt} Otherwise, set $\mathcal{G}^{\ast}=\mathcal{G}^{(s+1)}$ and stop the algorithm. \\
{\bf Output:} $\mathcal{G}^{\ast}$, and the set of ``cyclic edges" $\mathbb{C}$. \\ \hline\hline
\end{tabular}
\end{center}
\end{table}

\begin{prop}\label{prop:score_GSHD_hc}
In each step of the hill climbing algorithm with \texttt{score.GSHD$(\alpha)$}, the following operations (assuming eligible) will not decrease the score:
\begin{itemize}
\item add an edge with generalized selection frequency $\leq 0.5$;
\item delete an edge in the current graph;
\item reverse an edge in the current graph.
\end{itemize}
Moreover, among eligible edge addition operations, the one corresponding to the largest generalized selection frequency will lead to the most reduction of \texttt{score.GSHD$(\alpha)$}. Therefore, the hill climbing search algorithm can be conducted as described in Table \ref{table:agg_score_GSHD}.

In addition, $\mathcal{G}^\ast$ reaches the minimum \texttt{score.GSHD$(\alpha)$} value:
$$
score.GSHD(\alpha)(\mathcal{G}^\ast: \mathbb{G}^e)= \min_{\mathcal{G} \in \mathbb{G}(\mathbb{V})} score.GSHD(\alpha)(\mathcal{G}: \mathbb{G}^e),
$$
provided that $|\mathbb{C}| \leq 1$.
\end{prop}
The proofs of Propositions \ref{prop:score_gshd} and \ref{prop:score_GSHD_hc} are given in the Appendix (Section \ref{app:proof}).

\section{Numerical Study}
\label{sec:numerical}
In this section, we conduct an extensive simulation study to examine the proposed \texttt{DAGBag} procedure and compare it to several existing DAG learning algorithms.

\subsection{Simulation setting}

Given the true data generating DAG $\mathcal{G}$ and a sample size $n$, $n$ i.i.d. samples are generated according to the \textit{Gaussian linear mechanism} corresponding to $\mathcal{G}$:
 $$
 x_i=\sum_{j \in pa_i^{\mathcal{G}}} \beta_{ij} x_j +\epsilon_i,~~~ i=1,\cdots, p,
 $$
 where $\epsilon_i$s are independent Gaussian random variables with mean zero and variance $\sigma^2_i$.

 The coefficients $\beta_{ij}$s in the linear mechanism are uniformly generated from $\mathcal{B}=[-0.5,-0.3] \cup [0.3,0.5]$.  The error variances $\sigma^2_i$s are chosen such that for each node the corresponding \textit{signal-to-noise-ratio (SNR)}, defined as the ratio between the standard deviation of the signal part and that of the noise part, is in a given range $\mathcal{R}$.   Here we consider two settings, namely, a high SNR case $\mathcal{R}=[0.5, 1.5]$ and a low SNR case $\mathcal{R}=[0.2, 0.5]$.

The generated data are standardized to have sample mean zero and sample variance one before applying any DAG structure learning algorithm. For \texttt{DAGBag}, the number of bootstrap resamples is set to be $B=100$ and the learning algorithm applied to each resample is the score-based method with BIC score.
For each  simulation setting, $100$ independent replicates are generated.

%%% notes: \textcolor{red}{The data generation process is outlined in Table \ref{table:data_gen}.}

We consider DAGs with different dimension and complexity as well as different sample sizes. The data generating graphs are shown in Figures \ref{fig:tree} to \ref{fig:ultra_large} in the Appendix (Section \ref{app:figures_network}).
Statistics of each graph and the simulation parameters are  given in Table \ref{tab:graph_des}.

\begin{table}[http]
\small
\begin{center}
\caption{Graph statistics and simulation parameters. \label{tab:graph_des}}
\begin{tabular}{rrrrrrrr}
  \hline\hline
DAG & $p$ & $|\mathbb{E}|$ & V-struct & Moral $|\mathbb{E}|$ & $n$ & SNR  & max-step$^{*}$\\
  \hline
%Tree-Like Graph& 23 & 30 & 7 & 37 & 30, 50, 100&\\
Empty Graph & 1000 &0 &0 &0 &250 &[0.5,1.5]&2000\\\hline
Tree Graph & 100 &99&0&99&50&[0.5,1.5]&500\\\hline
 &  &  &  & & 50,102,&[0.2,0.5]&\\
Sparse Graph& 102 & 109 & 77 & 184& 200, 500,&[0.5,1.5]&500\\
&  &  &  &  & 1000, 5000&&\\
Dense Graph & 104 & 527 & 1675 & 1670 & 100&[0.2,0.5]&1000\\
& & & & & &[0.5,1.5]&\\

Large Graph& 504 & 515 & 307 & 808 & 100, 250&[0.2,0.5]&1000\\
& & & & & &[0.5,1.5]&\\
%Large Graph with& 511 & 445 &96 & 540 & 100, 250 & [0.5, 1.5]&1500 \\
%less v-structures& &  & & & & & \\\hline
Extra-Large Graph & 1000 & 1068 & 785 & 1823 & 100, 250&[0.5,1.5]&5000\\
%Double-Large Graph & 1791 & 1824 & 1330 & 3134 &350&5000\\
Ultra-Large Graph & 2639 & 2603 & 1899&4481 &250&[0.5,1.5]&10000\\
%power law Graph & 500 & & & 495 & 250& \\
\hline\hline
\end{tabular}
\\$^*$ maximum number of steps allowed in the search algorithm.
\end{center}
\end{table}

\subsection{Performance evaluation}
We report results on skeleton edge, v-structure and \textit{moral edge} (edges in the moral graph)  detection
by tables and figures. The reason to report results on these objects is because they are identifiable: I-equivalent DAGs have same sets of skeleton edges, v-structures and moral edges.

In the tables in  Appendix \ref{sec:appC}, for each method, the numbers of skeleton edges, v-structures and moral edges of the learned graph, denoted by ``Total E", ``Total V" and ``Total M", as well as the numbers of correctly identified skeleton edges, correctly identified v-structures and correctly identified moral edges, denoted by ``Correct E", ``Correct V", ``Correct M", are reported. All numbers are averaged over the results on $100$ replicates. The standard deviations are given in the parenthesis.

\begin{comment}
We also report tables with power and FDR in Appendix \ref{app:tables_fdr} which have a one-to-one correspondence with tables in Appendix \ref{app:tables}. For each replicate, power is defined as:
$$power=\frac{\hbox{number of true edges in the estimated DAG}}{ \hbox{number of total edges in the true DAG}},$$
and false positive proportion (fdp) is defined as:
$$fdp=\frac{\hbox{number of null edges in the estimated DAG}}{\hbox{ number of total edges in the estimated DAG}}, ~~ \frac{0}{0}:=0.$$
In the tables, power and FDR are the average power and fdp over $100$ replicates, respectively, and numbers in the parenthesis are standard deviations.
\end{comment}

 Under a given setting, if the learned graph of one method has similar or less number of edges than the learned graph of another method, while at the same time it identifies more correct edges, then it is fair to say that this method outperforms the other one because of less false positives and higher power.
  However, quite often, the method with less number of edges also identifies less correct edges, making methods comparison difficult. Therefore, besides  recording the results at the learned graphs, we also draw \textit{learning curves} for edge/v-structure detection to facilitate  methods comparison.

   For methods using a search algorithm, at each search step, we record the number of skeleton edges (v-structures) and the number of correct skeleton edges (v-structures) of the current graph. We then draw a ``Total E" (``Total V") versus ``Correct E" (``Correct V") curve across search steps. The end point of each curve denotes the result of the leaned graph (i.e., the place at which the search algorithm stops). For \texttt{MMHC} (by R package \texttt{bnlearn}) and \texttt{PC-Alg} (by R package \texttt{pcalg}), there is a tuning parameter $\alpha$ denoting the significance level used by the independence tests. Therefore, their learning curves are driven by this parameter: we run the algorithm on  a series of $\alpha$ and draw the learning trajectory  based on results across $\alpha$'s.

   If the learning curve of one method lies above that of another method, then the former method is better among the two, since it detects more correct edges (v-structures) at each given number of total edges (v-structures).

\subsection{Results and findings}
\subsubsection*{False positive reduction by aggregation}
To illustrate the effectiveness of the aggregation methods in  false positive reduction,
we first consider an empty graph with $p=1000$ nodes, $|\mathbb{E}|=0$ edge and sample size $n=250$.  As can been seen from Table \ref{tab:empty_network_n250}, aggregation methods result in very few false positive edges, whereas the non-aggregation methods, namely, \texttt{score}, \texttt{MMHC} and \texttt{PC-Alg}, all have very large number of  false positives. Same patterns are observed on the same graph with a larger sample size $n=500$ (results omitted).

We then consider
 a  graph with $p=504$ nodes, $|\mathbb{E}|=515$ edges and $|\mathbb{VS}|=307$ v-structures (Figure \ref{fig:large}), under two sample sizes, $n=100$ and $n=250$, with $SNR \in [0.5, 1.5]$.  Figures \ref{fig:dense_n100_0515_lc_chp2} and \ref{fig:dense_n250_0515_lc_chp2} show the \textit{learning curves} (driven by the updating steps in the search algorithm) in terms of skeleton edge detection and v-structure detection. The end point of each curve indicates the place where the corresponding algorithm stops (i.e., the learned graph). For the purpose of a better graphical presentation, the curves corresponding to \texttt{score} are cut short (since it stops much later than other methods) with the corresponding numbers given in the caption. Tables \ref{tab:indep_boot_large100} and \ref{tab:indep_boot_large250} in Appendix \ref{sec:appC} show the total number of skeleton edges/v-structures and correct number of skeleton edges/v-structures in the learned graph by each method.

 As can be seen from these figures, the aggregation methods have learning curves (colored curves) well above those of the non-aggregation \texttt{score} method (black curves) in both skeleton edge detection and v-structure detection, demonstrating their superior performance. Moreover, aggregation methods stop much earlier than the \texttt{score} method due to the (implicit) model regularization resulted from aggregation. This is why aggregation methods have much reduced number of false positives.

 Among \texttt{score.GSHD$(\alpha)$} methods, \texttt{adjSHD} ($\alpha=1$), \texttt{GSHD$(0.5)$} and  \texttt{GSHD$(1.5)$} have very similar learning curves, with the ones with smaller $\alpha$ stop later (i.e., more edges, higher false positive rate and higher power). \texttt{SHD} ($\alpha=2$) has inferior learning curves in edge detection and slightly better learning curves in v-structure detection compared with other \texttt{score.GSHD} methods with smaller $\alpha$. It also stops much earlier, demonstrating that over-penalization of edge direction discrepancy is detrimental for both learning efficiency (represented by learning curves) and power (represented by the position of the end point).

We also consider a \texttt{Tree Graph} with $p=100$ nodes (Figure \ref{fig:tree}), where we compare the aggregation based methods with the \texttt{minimum spanning tree (MST)} algorithm \cite{kruskal1956shortest,prim1957shortest}. \texttt{MST} is  the method of choice if we know aprior that the graph is a tree.   When sample size is $n=50$ (Table \ref{tab:tree_n50_indep_boot}), the aggregation methods have less power than \texttt{MST}, but their false positive rates remain low. When sample size is increased to $n=100$ (results omitted), the aggregation methods  are able to achieve a performance similar to that of \texttt{MST}.

Other main observations are listed below. More detailed results can be found in  Appendix \ref{sec:appB} (learning curves) and Appendix \ref{sec:appC} (tables).
\begin{itemize}
\item For the high SNR setting $(SNR \in [0.5, 1.5])$, aggregation methods have superior  learning curves compared to the non-aggregating  methods, in both skeleton edge detection and v-structure detection.
    \item For the low SNR setting  $(SNR \in [0.2, 0.5])$, under \texttt{Sparse Graph} (Figure \ref{fig:sparse}) and \texttt{Dense Graph} (Figure \ref{fig:dense}) (both graphs have $p \sim 100$) with $n \sim 100$, all methods tend to have similar learning curves in terms of skeleton edge detection and all of them perform poorly in v-structure detection due to lack of power (thus corresponding learning curves are not shown). While under the \texttt{Large Graph} ($p \sim 500$), the aggregation methods have better learning curves for both SNR ranges under both $n=100$ and $n=250$.

   \item  For all settings, aggregation based methods stop much earlier than the non-aggregation \texttt{score} method. This is due to model regularization induced by the aggregation process and it leads to much reduced number of false positives.
\item In terms of skeleton edge detection, \texttt{adjSHD}, \texttt{GSHD(1.5)},\texttt{GSHD(0.5)} have superior learning curves compared to \texttt{SHD}. The differences are more pronounced in the high SNR cases.

    \item In terms of v-structure detection, all aggregation based methods have somewhat similar learning curves. The main difference among these  methods lie in where the search algorithm stops.
        %Specifically, \texttt{CPSF} and \texttt{LN.CPSF} tend to have slightly better learning curves than other methods. They %also tend to stop later and thus are a bit more powerful in v-structure detection. This is probably due to the use of %pairwise edge selection frequencies in these two methods, which is more efficient for detecting complex structures %involving more than one edge.

      %  \item The \texttt{DAGRand} procedure is able to slightly improve the aggregation results, demonstrated by the slightly %better learning curves of \texttt{DAGRand} (coupled with \texttt{adjSHD}) compared to those of \texttt{adjSHD}. This can %also be seen by the results under the \texttt{Dense Graph} with $n=500, SNR \in [0.5, 1.5]$ (Table %\ref{tab:indep_boot_verydense500}, Figure \ref{fig:dense_0515_n500_lc}), where \texttt{DAGRand} is able to improve the %aggregation performance particularly in v-structure detection (less ``Total V", more ``Correct V").
\end{itemize}
\textsl{In summary, aggregation based methods are particularly competitive for either high SNR or high-dimensional settings.}

\subsubsection*{Effect of sample size on aggregation methods}
Here we study effect of sample size on DAG learning procedures. By comparing results on  \texttt{Large Graph} with both SNR ranges: $n=100$ vs. $n=250$ (Table \ref{tab:indep_boot_large100} vs. Table \ref{tab:indep_boot_large250}; Table \ref{tab:indep_boot_large100_0205} vs. Table \ref{tab:indep_boot_large250_0205}), it appears that sample size $n$ does not have much effect on the false positive rate of the aggregation methods. Even under small sample sizes, the aggregation methods are able to maintain a low false positive rate. The effect of sample size is mainly in power: the smaller $n$, the less power in edge/v-structure detection.

 Due to the consistency of the BIC score, we expect all methods (including \texttt{score}) will converge to a graph in the true I-equivalence class when $n$ goes to infinity. We  conduct a simulation on \texttt{Spare Graph} with $SNR \in [0.5, 1.5]$ under a series of sample sizes $n=50, 102, 200, 500, 1000, 5000$ to demonstrate such a convergence. Figure \ref{adjshd_differ_size} shows the skeleton edge detection learning curves of \texttt{score} and \texttt{adjSHD} under different sample sizes. It can be seen that, the curves corresponding to  larger sample sizes lie above those corresponding to smaller sample sizes. Moreover, with $n$ increasing,  the curves become closer to the diagonal line where number of false positives becomes zero. These results  are also given in Table \ref{tab:sparse_agg_diff_n}.

\subsubsection*{Effect of graph topology on aggregation methods}
Here we study the effect of graph topology on aggregation methods (focusing on \texttt{adjSHD} for ease of comparison and illustration). We first compare the results under the \texttt{Sparse Graph} which has $p \sim100, |\mathbb{E}| \sim 100$ (Figure \ref{fig:sparse}) to the results under the \texttt{Dense Graph} which has $p \sim 100, |\mathbb{E}| \sim 500$ and many more v-structures (Figure \ref{fig:dense}).
For simulations with $n \sim 100$ and $SNR \in [0.5, 1.5]$ (Table \ref{tab:indep_boot}  vs. Table \ref{tab:indep_boot_verydense100}),  the  power is much smaller under the \texttt{Dense Graph} due to its higher \textit{intrinsic dimension}/more complex model. However, the false positive rate is only slightly higher under the \texttt{Dense Graph}.
This again demonstrates the effectiveness of the aggregation methods in false positive reduction  even when the underlying model is complex.

We then compare the results under \texttt{Large} ($p \sim 500$, Figure \ref{fig:large}), \texttt{Extra-large} ($p \sim 1000$, Figure \ref{fig:extra_large}) and \texttt{Ultra-large} ($p \sim 2600$, Figure \ref{fig:ultra_large}) graphs. All these graphs have similar edge/node ratio in that $p \sim |\mathbb{E}|$. However they have increasing number of nodes. For simulations with $n = 250$ and $SNR \in [0.5, 1.5]$ (Tables \ref{tab:indep_boot_large250}, \ref{tab:extra_large_250},\ref{tab:super_large_250}),  both false positive rates and powers  are quite similar across these three networks. This indicates that, the \textit{nominal dimension}  alone  does not have much effect on performance of aggregation methods.

%% contrast with performance of score; hard to do, since for the socre the number of steps are capped by max-step. this would distort the fdr.

\begin{comment}
Next, we compare the results under  \texttt{Large Graph}  and \texttt{Large Graph with less v-structures} (Figure \ref{fig:large_lessv}). The major difference between these two graphs is that the former has $307$ v-structures and the latter has only $96$ v-structures. For simulations with $n = 100$ and $n=250$ and $SNR \in [0.5, 1.5]$ (Table \ref{tab:indep_boot_large100} vs. Table \ref{tab:lessv_network_n100}; Table \ref{tab:indep_boot_large250} vs. Table \ref{tab:lessv_network_n250}), false positive rates are very similar under these two graphs, but the more complex \texttt{Large Graph} leads to much less power.
\end{comment}

\textsl{
In summary, for aggregation methods, false positive rate is  not much affected by the complexity of the underlying model, while the power is lower for more complex models. Moreover, the complexity of the model is not much affected by the nominal dimension (i.e., the number of nodes), but rather factors including the intrinsic dimension, degree of sparsity of the network, the number of v-structures, etc.
}

\subsubsection*{Robustness to distributional assumptions}

To investigate the robustness of the score-based procedures (aggregation or not) to the Normality assumption, instead of sampling the residual terms of the linear mechanisms from Normal distributions, they  are sampled from mean-centered student t-distributions (df $=3$ and df $=5$) and Gamma distributions (shape parameter $=1$, scale parameter $=2$).
 %The probability density  curves are shown in Figure \ref{fig:robust_dist}. The variance of the residual terms are still determined %by the specified SNR range as described in Section \ref{sec:simu_glm}.

The results  under the \texttt{Sparse Graph} with $p=102$ nodes, $|\mathbb{E}|=109$ edges, and sample size $n=102, SNR \in [0.5, 1.5]$ are shown in Table \ref{tab:bic_robust}. We conclude that the score-based procedures are robust with respect to distributional assumptions.

%\begin{figure}[htbp]
%\begin{center}
%\includegraphics[scale=0.4]{robust_dists.pdf}
%\caption{Probability density curves of standard normal distribution, t-distribution with $df=3$ and $df=5$ and  Gamma distribution %with $shape=1, scale=2$.  \label{fig:robust_dist}}
%\end{center}
%\end{figure}

\subsubsection*{Comparison of scores}
Here, we compare different scores by a simulation under \texttt{Sparse Graph} with $p=n=102$. The scores under consideration include negative maximum log-likelihood score (\texttt{like}), \texttt{BIC}, \texttt{eBIC}, \texttt{GIC} and score equivalent Gaussian posterior density (\texttt{BGe}, by the \texttt{bnlearn} R package) .

Extended Bayesian information criterion (eBIC) \cite{chen2008extended} and the generalized information criterion (GIC)  \cite{kim2012consistent} are two criteria for high-dimensional setting which penalize more on model complexity compared with BIC:
\begin{eqnarray*}
\text{score}_{eBIC} (\mathcal{G}: \mathbb{D})&=&\sum_{i=1}^p  \left(n\log(RSS_i/n)+|pa^\mathcal{G}_i| \left(\log(n)+2\log(p)\right) \right)\\
\text{score}_{GIC} (\mathcal{G}: \mathbb{D})&=&\sum_{i=1}^p \left(n\log(RSS_i/n)+|pa^\mathcal{G}_i| \left(\log\log(n)\right) \log(p)\right).
\end{eqnarray*}
For the \texttt{BGe} score, there is a parameter ``iss", equivalent sample size,  that needs to be specified. Here we consider $iss=3$ and $iss =10$ ($10$ is the default in \texttt{bnlearn}).

It can be seen from Table \ref{tab:decomposable_scores} that,  in terms of the performance of the (non-aggregation) \texttt{score} method, \texttt{like} and \texttt{BGe} with $iss =10$ result in most false positives.  For \texttt{like}, this is because it does not penalize on model complexity and thus leads to larger models with more edges. As for \texttt{Bge}, although $iss=10$ is the default value for the prior sample size, it appears to be too liberal under this simulation setting.

\texttt{GIC} and \texttt{BGe} with $iss=3$ seemingly have the best \texttt{score} results: less total edges  and more correct edges.
\texttt{eBIC} penalizes the most on model complexity (in this case, the factor is $13.9$ for \texttt{eBIC}, compared with $7.1$ for \texttt{GIC} and $4.6$ for \texttt{BIC}). Therefore, it has least false positives, but also the least power.

 After model aggregation,  \texttt{BIC}, \texttt{like}, \texttt{GIC}  all perform very similarly. \texttt{BGe} with $iss=10$ is not as competitive as \texttt{BGe} with  $iss=3$, as the former detects considerably more total edges/v-structures with only slightly more correct edges/v-structures. \texttt{eBIC} is still under-powered, but with lower false positive rates than others.
 %Figure \ref{fig:BGe_sparse} shows that the learning curves on edge detection by \texttt{BIC}, \texttt{BGe}($iss=3$), \texttt{eBIC} %and \texttt{GIC} scores are very similar. The difference mainly lies in where the search stops.

 Coupled with aggregation procedures, the advantage of \texttt{BIC} over  \texttt{like} mainly lies in computation, as on each bootstrap resample, we expect the search algorithm to stop earlier with \texttt{BIC} score. The drawback of  \texttt{BGe} is  that it is not straightforward to choose a good $iss$ value in practice.

 %More comparisons among \texttt{BIC}, \texttt{eBIC} and \texttt{GIC} can be found in Tables \ref{tab:decomp_scores_large100} and %Table \ref{tab:decomp_scores_large250} under the \texttt{Large Graph}, which suggest similar patterns as observed here.

As an alternative to the aforementioned scores, \cite{elidan2011bagged} proposed a bagged estimate for log-likelihood score. For each eligible operation, the likelihood score change  is averaged across bootstrap resamples. The operation that results in the most (bootstrap averaged) score improvement is selected.  This method may be beneficial for building good predictive models, but it results in too many false positives in terms of structure learning. Table \ref{tab:bagged_est} shows its performance under the \texttt{Spare Graph}, which is similar as that of  \texttt{BGe} with $iss=10$ in the same setting.

 \textsl{In summary, \texttt{BIC} score coupled with  \texttt{DAGBag} is a good choice.}

\section{Summary}
\label{sec:discussion}
In this paper, we made two major contributions in DAG structure learning. First, we propose an efficient implementation of the
popular hill climbing search algorithm. Second,
we propose an aggregation framework \texttt{DAGBag} which aggregates an ensemble of DAGs learned on bootstrap resamples.
We also propose a family of distance metrics on the DAG space based on the structural hamming distance and investigate their properties and corresponding aggregation scores.
Through
extensive simulation studies, we show that the \texttt{DAGBag} procedure is able to greatly reduce number of false positives compared to the non-aggregation procedure it couples with. It should be noted that, the \texttt{DAGBag} procedure is a general procedure which may be coupled with any DAG learning algorithm. We implement the \texttt{DAGBag} procedure in an R package \texttt{dagbag} which is available on \texttt{http://anson.ucdavis.edu/$\sim$jie/software.html}.

%%%%%%%%%%%

%%%%%%%%%%%%%%%%%%

\clearpage
\newpage
\appendix

\section{Technical Details}
\label{sec:appnedix}
\renewcommand{\thesection}{A-\arabic{section}}
\renewcommand{\theequation}{A-\arabic{equation}}
\renewcommand{\thetable}{A-\arabic{table}}
\renewcommand{\thefigure}{A-\arabic{figure}}
\setcounter{figure}{0}
\setcounter{table}{0}
\setcounter{section}{0}
\subsection{Implementation details}

\label{app:alg}
\subsubsection*{Early stopping}
As mentioned earlier, when the number of nodes $p$ is large, it usually takes many steps  for the search algorithm to achieve a non-negative minimum score change: $\delta_{\min} \geq 0$. We observe that, the best operations selected in the later steps of the search often lead to only tiny decreases of the score. Indeed, these  are mostly ``wrong" operations, for example the addition of a \textit{null edge} (i.e., an edge not in the true model).  This phenomenon is particularly severe for the $p>n$ case due to over-fitting of the noise. It not only results in excessive and un-necessary computational cost, but also leads to large number of false positives in edge detection.

Therefore, we propose an early stopping rule which stops the search algorithm when the decrease of the score is smaller than a threshold $\epsilon$, i.e., we stop the search algorithm whenever  $\delta_{\min} >-\epsilon$. In our implementation,
the default value of $\epsilon$ is set as  $10^{-6}$.  Even through this is a very simple early stopping rule,   it not only
speeds up the search algorithm, but also reduces the number of false positives. In this sense, it may also be viewed as a  regularization scheme.

%%% based on an old result. OK for illustration purpose.
We use a numerical example to illustrate the effectiveness of this early stopping rule. We generate $50$ i.i.d. samples based on a Gaussian linear mechanism  according to a graph with $102$ nodes and $109$ edges. With the early stopping rule, the algorithm detects $483$ skeleton edges with $84$ of them being correct edges, while without early stopping, the algorithm detects $875$ skeleton edges with $92$ being correct.
%In a Graph with $p=123$ nodes and $|\mathbb{E}|=208$ edges, when sample size $n=50$,  these numbers are $108$ correct edges out of  $488$ total edges %with early stopping, and $131$ correct edges out of $1534$ total edges without early stopping.

When the number of nodes $p$ is much larger than the sample size $n$, even with the above early stopping rule, the search algorithm often still runs many steps before it stops.  To avoid too much computational burden, we may set a \texttt{max-step} and the search algorithm will be stopped if the number of steps reaches max-step whether it is deemed converged or not.

%%%%%%%%%%%%%%%%%%%%%%%%%%%%%%%%%%%%%%%%%%
\subsubsection*{Random restart}
Because hill climbing algorithm is a greedy search algorithm which usually leads to local optimal solutions, one may want to  utilize techniques for  global optimal search such as random restart \cite{scutari2009learning}, data perturbation \cite{elidan2002data} and simulated annealing \cite{de2000approximating}.
In random restart, after the hill climbing algorithm stops, we may perturb the learnt DAG structure through randomly adding new edges or randomly deleting and reversing existing edges (under the acyclic constraint). We then restart the search starting from the perturbed structure. This process is repeated several times and in the end, the structure with the smallest score is selected. Since under the high-dimensional setting, the learnt DAG often contains many false positive edges, we may revise the above procedure by only allowing deletion and reversal in the perturbation.

\cite{elidan2002data}  comment on random restart and simulated annealing in DAG learning that these techniques randomly perturb the structure of local optima and are unlikely to lead to improvement in the score.
 This is confirmed by our numerical experiments,  where we observe that neither score nor edge detection is improved through random restart, indicating that the local optima achieved by hill climbing is very stable and not easy to jump out of at least by simple methods.

\subsubsection*{Blacklist and whitelist}
Sometimes, we may have prior knowledge about the presence or absence of certain edges.
For example, in constructing expression quantitative trait loci networks,
it is reasonable to assume that only genotype nodes may point to expression nodes but not vice versa.
If available, such information may be used to constrain the model space through a \textit{blacklist} and/or a \textit{whitelist}. The former forbids the presence of some edges and the latter guarantees the presence of some edges. Implementation of the hill climbing search algorithm with such information is straightforward: in each search step, an \textit{eligible} operation is now an operation which is not forbidden by
the blacklist and passes the acyclic check.  Moreover, all edges in the whitelist would be in the initial graph and will never be deleted or reversed in the subsequent steps.

\subsection{Proofs}
\label{app:proof}
\begin{comment}
\noindent {\bf Proof of Proposition \ref{prop:DAG-moral-MRF}}. By Proposition \ref{prop:DAG-MRF}, $ \mathbb{E}(\mathcal{\tilde{G}}) \subset \mathbb{E}(\mathcal{G}^m)$. On the other hand, if $x_i \sim_{\mathcal{G}^m} x_j$, then either $x_i \sim_{\mathcal{G}} x_j $ or $x_i$ and $x_j$ are co-parents in $\mathcal{G}$. For both cases, $x_i$ and $x_j$ is not d-separated by  $x_{\mathbb{V} \backslash \{i,j\}}$ in $\mathcal{G}$. Since $\mathcal{G}$ is a P-map for $\mathbb{P}$, we have  $x_{i}$ and  $x_{j}$ are conditionally dependent given $x_{\mathbb{V} \backslash \{i,j\}}$. Then by definition, $x_i \sim _{\mathcal{\tilde{G}}} x_j$. End of proof. \\
\end{comment}

\noindent{\bf Proof of Proposition \ref{prop:score_shd}}. By definition of aggregation score
\begin{eqnarray*}
score.SHD(\mathcal{G}: \mathbb{G}^e)&=&\frac{1}{B} \sum_{b=1}^B d_{SHD}(\mathcal{G}, \mathcal{G}_b)=\frac{1}{B} \sum_{b=1}^B \sum_{(i,j)} |\mathbb{A}(i,j)-\mathbb{A}_b(i,j)|\\
&=&  \frac{1}{B}  \sum_{(i,j) \in \mathbb{E}(\mathcal{G})} \sum_{b=1}^B  (1- \mathbb{A}_b(i,j))+\frac{1}{B}  \sum_{(i,j) \notin \mathbb{E}(\mathcal{G})} \sum_{b=1}^B  \mathbb{A}_b(i,j),
\end{eqnarray*}
where $(i,j)$ denotes the edge $x_i \rightarrow x_j$.  By definition of selection frequency,
\begin{eqnarray*}
\frac{1}{B}  \sum_{ (i,j) \in \mathbb{E}(\mathcal{G})} \sum_{b=1}^B  \mathbb{A}_b(i,j) = \sum_{ (i,j) \in \mathbb{E}(\mathcal{G})} p_{(i,j)}.
\end{eqnarray*}
Moreover,
$$
\frac{1}{B}  \sum_{ (i,j) \in \mathbb{E}(\mathcal{G})} \sum_{b=1}^B  \mathbb{A}_b(i,j) + \frac{1}{B}  \sum_{(i,j) \notin \mathbb{E}(\mathcal{G})} \sum_{b=1}^B  \mathbb{A}_b(i,j) =\frac{1}{B}\sum_{b=1}^B \sum_{i=1}^p \sum_{j=1}^p \mathbb{A}_b(i,j)=: C.
$$
Therefore,
$$
score.SHD(\mathcal{G}: \mathbb{G}^e)=\sum_{(i,j) \in \mathbb{E}(\mathcal{G})}(1-p_{(i,j)})+ C - \sum_{(i,j) \in \mathbb{E}(\mathcal{G})}p_{(i,j)}.
$$
End of proof. \\

\noindent {\bf Proof of Proposition \ref{prop:score_SHD_hc}}. Let $\mathcal{G}$ denote the current graph.

(i) If the operation $O$ is to add an edge $e$ where $p_e\leq 0.5$. Then by Proposition \ref{prop:score_shd}, the change of \texttt{score.SHD} is
$$
score.SHD(O(\mathcal{G}): \mathbb{G}^e)-score.SHD(\mathcal{G}: \mathbb{G}^e) = 1-2p_e \geq 0.
$$

(ii) If the operation $O$ is to delete an existing edge $e$ in the current graph $\mathcal{G}$, then the change of score  is
$$
score.SHD(O(\mathcal{G}): \mathbb{G}^e)-score.SHD(\mathcal{G}: \mathbb{G}^e) = 2p_e-1 > 0,
$$
since $p_e >0.5$ as shown in (i).

(iii) If the operation $O$ is to reverse an existing edge $e$ in the current graph $\mathcal{G}$, then the change of score  is
$$
score.SHD(O(\mathcal{G}): \mathbb{G}^e)-score.SHD(\mathcal{G}: \mathbb{G}^e) = (1-2p_{e^{*}})-(1-2p_e) >0,
$$
where $e^{*}$ is the reversed edge from $e$. The above is because, $p_e >0.5$ as shown in (i). Moreover, by $p_e+p_e^{*} \leq 1$ as $e, e^*$ can not appear simultaneously  in a DAG, we have $p_{e^{*}} <0.5$.

(iv) For addition operations,  it is obvious from (i)  that, the one with the largest selection frequency leads to  the most score decrease.  End of proof. \\

\noindent {\bf Proof of Proposition \ref{prop:score_SHD_alg}}. We want to show that, if $|\mathbb{C}| \leq 1$, then  for any $\mathcal{G} \in \mathbb{G}(\mathbb{V})$, there is
$$
score.SHD(\mathcal{G}^\ast: \mathbb{G}^e) \leq score.SHD(\mathcal{G}: \mathbb{G}^e) ~~~ (\ast).
$$

Let $\mathbb{E}^{+}=\{x_i \rightarrow x_j: p_{(i,j)}>0.5, 1 \leq i, j \leq p\}$  be the set of all possible edges having SF $>0.5$.
Let $\mathbb{G}(\mathbb{V})^{+}$ be the subset of DAGs  with edges all in $\mathbb{E}^{+}$.

We only need to proof $(\ast)$ for DAGs in  $\mathbb{G}(\mathbb{V})^{+}$. To see why this is the case, for a  DAG $\mathcal{G}$,  let $\mathbb{E}^{-}(\mathcal{G})$ denote the set of its edges with SF $\leq 0.5$. Let $O$ be ``delete all edges in $\mathbb{E}^{-}(\mathcal{G})$". Then $O(\mathcal{G})$ is still a DAG since no cycle will be created by dropping edges and all its edges have SF $>0.5$, so $O(\mathcal{G}) \in \mathbb{G}(\mathbb{V})^{+}$. Moreover, by Proposition \ref{prop:score_shd},
$$
score.SHD(O(\mathcal{G}): \mathbb{G}^e)-score.SHD(\mathcal{G}: \mathbb{G}^e)=\sum_{e \in \mathbb{E}^{-}(\mathcal{G})}(2p_e-1) \leq 0.
$$

Moreover, $(\ast)$ holds for $\mathcal{G} \in \mathbb{G}(\mathbb{V})^{+}$ with $\mathbb{E}(\mathcal{G}) \subseteq \mathbb{E}(\mathcal{G}^{\ast})$, since
$$
score.SHD(\mathcal{G}^{\ast}: \mathbb{G}^e)-score.SHD(\mathcal{G}: \mathbb{G}^e) = \sum_{e \in \mathbb{E}(\mathcal{G}^{\ast}) \backslash \mathbb{E}(\mathcal{G})} (1-2p_e) \leq 0,
$$
as by definition, all edges in $ \mathcal{G}^{\ast}$ have SF $>0.5$.

As in Algorithm Table \ref{table:agg_score_SHD}, order the edges in  $\mathbb{E}^{+}$ by their selection frequencies from the largest to the smallest and denote the edges passing the acyclic check by ``a" and those failing the acyclic check by ``c".

If there is no ``c" edge, i.e., $|\mathbb{C}|=0$, then $\mathbb{E}(\mathcal{G}^{\ast})=\mathbb{E}^{+}$. Thus for any DAG $\mathcal{G} \in \mathbb{G}(\mathbb{V})^{+}$,  we have $\mathbb{E}(\mathcal{G}) \subseteq \mathbb{E}(\mathcal{G}^{\ast})$ and $(\ast)$ holds.

If there is one ``c" edge, i.e.,  $|\mathbb{C}|=1$, denoted by $c^*$, let $\mathbb{E}_1^{+}$ be the set of ``a" edges having SF greater or equal to that of  $c^{*}$ and
 $\mathbb{E}_2^{+}$ be the rest of the ``a" edges. Then $\mathbb{E}(\mathcal{G}^{\ast})= \mathbb{E}_1^{+}  \cup  \mathbb{E}_{2}^{+}$.
 Consider a DAG $\mathcal{G} \in \mathbb{G}(\mathbb{V})^{+}$ such that $\mathbb{E}(\mathcal{G})$ is not a subset of  $\mathbb{E}(\mathcal{G}^{\ast})$, so it has to contain the $c^{*}$ edge. Let $\mathbb{E}(\mathcal{G})^{+}$ be the  subset of edges in $\mathbb{E}(\mathcal{G})$ ordered before $c^{*}$ and $\mathbb{E}(\mathcal{G})^{-}$ be the subset of edges ordered after $c^{*}$. So  $\mathbb{E}(\mathcal{G})=  \mathbb{E}(\mathcal{G})^{+} \cup \{c\} \cup \mathbb{E}(\mathcal{G})^{-}$.

Due to the acyclic constraint, at least one edge $e^*$ in  $\mathbb{E}_1^{+}$ can not be in $\mathbb{E}(\mathcal{G})$, so $\mathbb{E}(\mathcal{G})^{+} \subseteq  \mathbb{E}_1^{+} \backslash e^*$. Moreover, $\mathbb{E}(\mathcal{G})^{-} \subseteq  \mathbb{E}_{2}^{+}$. Therefore,
\begin{eqnarray*}
&& score.SHD(\mathcal{G}^{\ast}: \mathbb{G}^e)-score.SHD(\mathcal{G}: \mathbb{G}^e)  \\
&\leq& (1-2p_{e^*})-(1-2p_{c^*})+ \sum_{e \in \mathbb{E}_{2}^{+} \backslash \mathbb{E}(\mathcal{G})^{-}} (1-2p_e) \leq 0,
\end{eqnarray*}
since $p_{e^*} \geq p_{c^*}$ and $p_e > 0.5$. End of proof. \\

\noindent{\bf Proof of Proposition \ref{prop:score_gshd}}. To help with the proof, we use Table \ref{table:Sij} to
 show the value of $S_{ij}(\mathbb{A}, \mathbb{\tilde{A}}; \alpha)$ for $1 \leq i<j \leq p$. For an adjacency matrix $\mathbb{A}$ and a given pair $(i,j)$ with $i<j$, let $(1,0)$ denote the case where $\mathbb{A}(i,j)=1, \mathbb{A}(j,i)=0$, $(0,1)$  denote the case where $\mathbb{A}(i,j)=0, \mathbb{A}(j,i)=1$ and $(0,0)$  denote the case where $\mathbb{A}(i,j)=0, \mathbb{A}(j,i)=0$ (note $(1,1)$ is not possible due to the acyclic constraint).

\begin{table}
\caption{$S_{ij}(\mathbb{A}, \mathbb{\tilde{A}}; \alpha)$ with $1 \leq i<j \leq p$ and $\alpha>0$. \label{table:Sij}}
\begin{center}
\begin{tabular}{c|ccc}\hline
$\mathbb{\tilde{A}} \backslash  \mathbb{A}
$ & $(1,0)$ &$(0,1)$& $(0,0)$\\\hline
$(1,0)$& $0$&$\alpha$&$1$\\\hline
$(0,1)$& $\alpha$&$0$&$1$\\\hline
$(0,0)$& $1$&$1$&$0$\\\hline
\end{tabular}
\end{center}
\end{table}

By definition:
\begin{eqnarray*}
score.GSHD(\alpha)(\mathcal{G}: \mathbb{G}^e)&=&\frac{1}{B} \sum_{b=1}^B \sum_{1 \leq i<j \leq p} S_{ij}(\mathbb{A}, \mathbb{A}_b; \alpha)\\
&=& \sum_{i<j: x_i \rightarrow x_j \in \mathbb{E}(\mathcal{G})} \left(\alpha p_{ji}+p_{ij}^0\right)\\
&+& \sum_{i<j: x_j \rightarrow x_i \in \mathbb{E}(\mathcal{G})} \left(\alpha p_{ij}+p_{ij}^0\right)\\
&+& \sum_{i<j: x_i \rightarrow x_j \notin \mathbb{E}(\mathcal{G}), x_j \rightarrow x_i \notin \mathbb{E}(\mathcal{G})} \left(p_{ij}+p_{ji}\right),
\end{eqnarray*}
where $p_{ij}$ denotes the selection frequency of edge $x_i \rightarrow x_j$, and $p_{ij}^0=p_{ji}^0:=1-p_{ij}-p_{ji}$ is the frequency that neither  $x_i \rightarrow x_j$ nor  $x_j \rightarrow x_i$ got selected.

Note that,
\begin{eqnarray*}
\sum_{i<j: x_i \rightarrow x_j \notin \mathbb{E}(\mathcal{G}), x_j \rightarrow x_i \notin \mathbb{E}(\mathcal{G})} \left(p_{ij}+p_{ji}\right) &=& \sum_{1 \leq i<j \leq p} \left(p_{ij}+p_{ji}\right) \\
&-& \sum_{i<j: x_i \rightarrow x_j \in \mathbb{E}(\mathcal{G})} \left(p_{ij}+p_{ji}\right) -  \sum_{i<j: x_j \rightarrow x_i \in \mathbb{E}(\mathcal{G})} \left(p_{ij}+p_{ji}\right).
\end{eqnarray*}
Therefore
 \begin{eqnarray*}
score.GSHD(\alpha)(\mathcal{G}: \mathbb{G}^e)&=& \sum_{i<j: x_i \rightarrow x_j \in \mathbb{E}(\mathcal{G})} \left(\alpha p_{ji}+1-2(p_{ij}+p_{ji})\right)\\
&+& \sum_{i<j: x_j \rightarrow x_i \in \mathbb{E}(\mathcal{G})} \left(\alpha p_{ij}+1-2(p_{ij}+p_{ji})\right)\\
&+&\sum_{1 \leq i<j \leq p} \left(p_{ij}+p_{ji}\right)\\
&=&\sum_{i<j: x_i \rightarrow x_j \in \mathbb{E}(\mathcal{G})} \left(1-2 \left(p_{ij}+(1-\frac{\alpha}{2})p_{ji}\right)\right)\\
&+&\sum_{i<j: x_j \rightarrow x_i \in \mathbb{E}(\mathcal{G})} \left(1-2 \left(p_{ji}+(1-\frac{\alpha}{2})p_{ij}\right)\right)\\
&+&\sum_{1 \leq i<j \leq p} \left(p_{ij}+p_{ji}\right).\\
\end{eqnarray*}
By definitions  of the constant $C$ and the generalized selection frequency, we complete the proof. \\

\noindent {\bf Proof of Proposition \ref{prop:score_GSHD_hc}}.  Let $\mathcal{G}$ denote the current graph.

(i) If the operation $O$ is to add an edge $e$ where $gp_e\leq 0.5$. Then by Proposition \ref{prop:score_gshd}, the change of \texttt{score.GSHD$(\alpha)$} is
$$
score.GSHD(\alpha)(O(\mathcal{G}): \mathbb{G}^e)-score.GSHD(\alpha)(\mathcal{G}: \mathbb{G}^e) = 1-2gp_e \geq 0.
$$

(ii) If the operation $O$ is to delete an existing edge $e$ in the current graph $\mathcal{G}$, then the change of score  is
$$
score.GSHD(\alpha)(O(\mathcal{G}): \mathbb{G}^e)-score.GSHD(\alpha)(\mathcal{G}: \mathbb{G}^e) = 2gp_e-1 > 0,
$$
since $gp_e >0.5$ as shown in (i).

(iii) For addition operations,  it is obvious from (i)  that, the one with the largest generalized selection frequency leads to  the most score decrease.

(iv) Consider an operation $O$ to reverse an existing edge $e$ in the current graph $\mathcal{G}$. Assume that it is eligible. Also assume that there is no reversal operation got selected in any of the previous steps.  Then the change of score  is
\begin{eqnarray*}
score.GSHD(\alpha)(O(\mathcal{G}): \mathbb{G}^e)-score.GSHD(\alpha)(\mathcal{G}: \mathbb{G}^e) &=& (1-2gp_{e^{*}})-(1-2gp_e)\\
&=&2(gp_e - gp_e^{*}),
\end{eqnarray*}
where $e^{*}$ is the reversed edge from $e$.  Since at the current step, reversing $e$ is an eligible operation (i.e., not causing cycle), then at a previous step $s$ where edge $e$ got added, adding edge $e^{*}$ must also be an eligible operation since there is no deletion (due to (ii)) and reversal operation (by assumption) got selected before this step. Since at step $s$, the operation ``add $e$" got selected over the operation ``add $e^{*}$", it means that $gp_e \geq gp_e^{*}$ due to (iii). Therefore, the score change is non-negative. This also means that the operation ``reverse $e$" will not be selected at the current step. By induction, no eligible reversal operation will lead to score decrease.

The last part of the Proposition can be proved by the same arguments as in the proof of Proposition \ref{prop:score_SHD_alg} and thus the details are omitted here.  End of proof. \\

\clearpage
\newpage

\renewcommand{\thesection}{B-\arabic{section}}
\renewcommand{\theequation}{B-\arabic{equation}}
\renewcommand{\thetable}{B-\arabic{table}}
\renewcommand{\thefigure}{B-\arabic{figure}}
\setcounter{figure}{0}
\setcounter{table}{0}
\setcounter{section}{0}
\section{Figures}
\label{sec:appB}

\subsection*{DAGs used in the simulation study}
\label{app:figures_network}

\begin{figure}
\begin{center}
\includegraphics[scale=0.75]{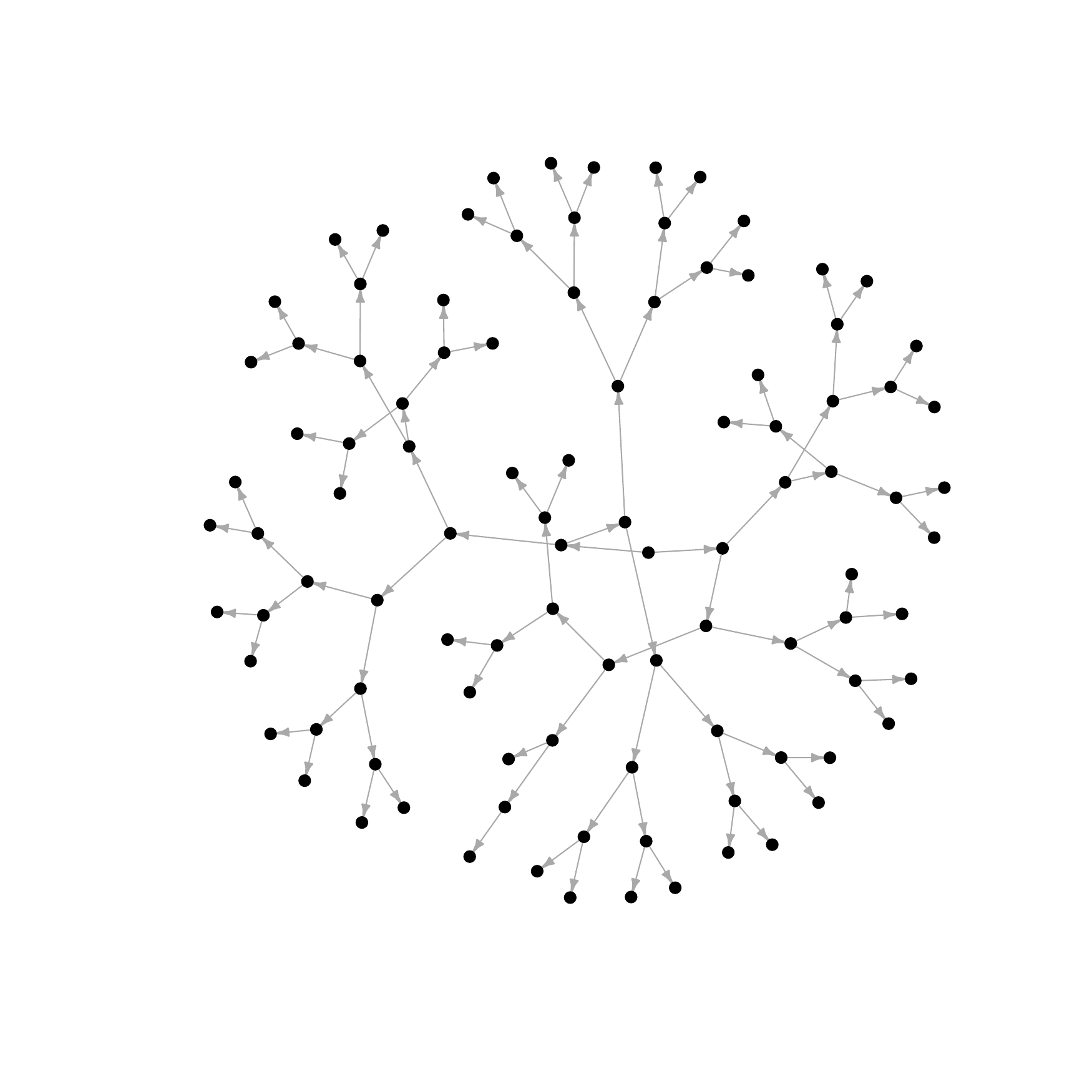}
\caption{Tree Graph with $p=100, |\mathbb{E}|=99$. \label{fig:tree}}
\end{center}
\end{figure}

\begin{figure}
\begin{center}
\includegraphics[scale=0.6]{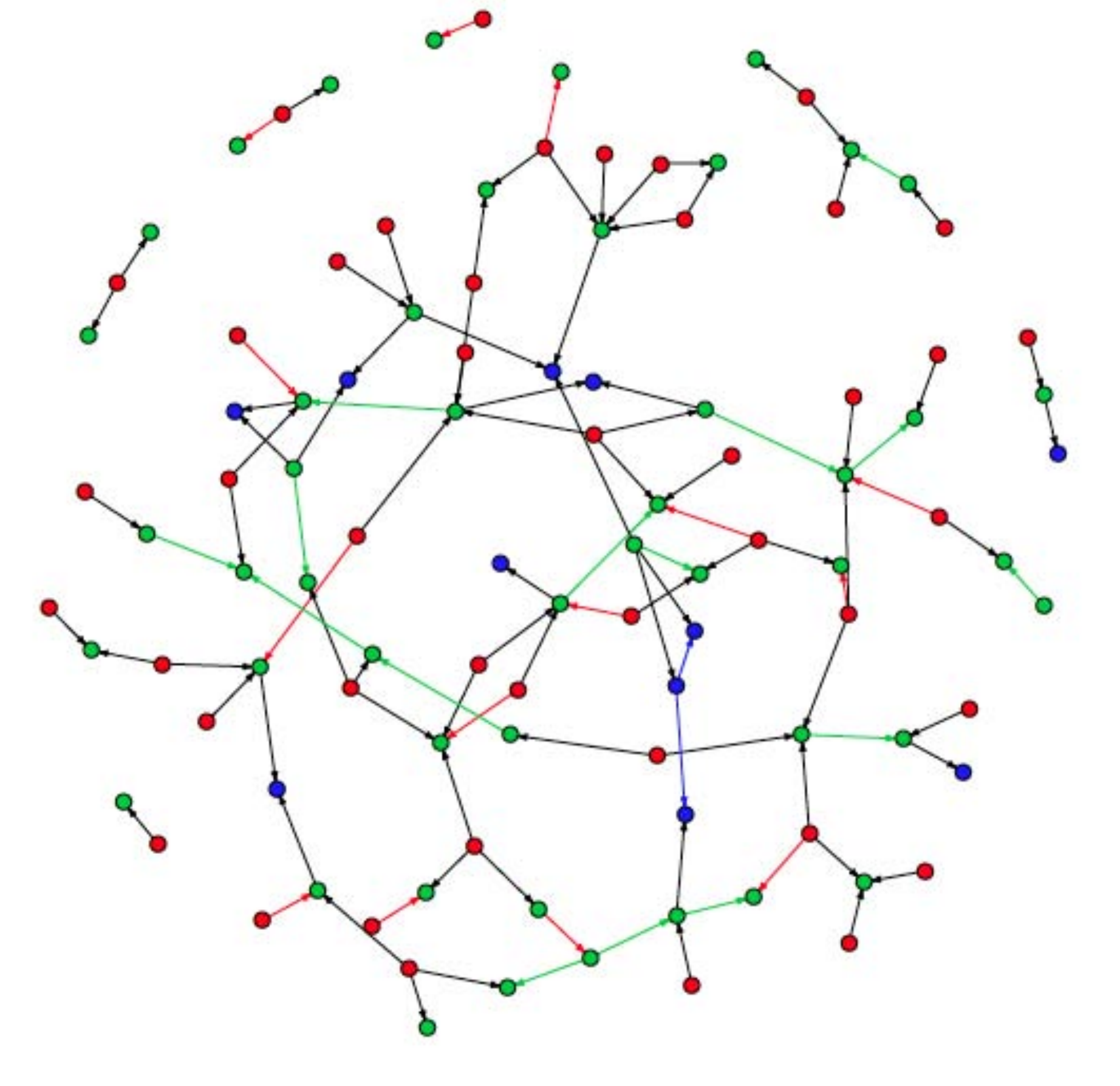}
\caption{Sparse Graph with $p=102, |\mathbb{E}|=109, |\mathbb{VS}|=77$, Moral $|\mathbb{E}|=184$. \label{fig:sparse}}
\end{center}
\end{figure}

\begin{figure}
\begin{center}
\includegraphics[scale=0.8]{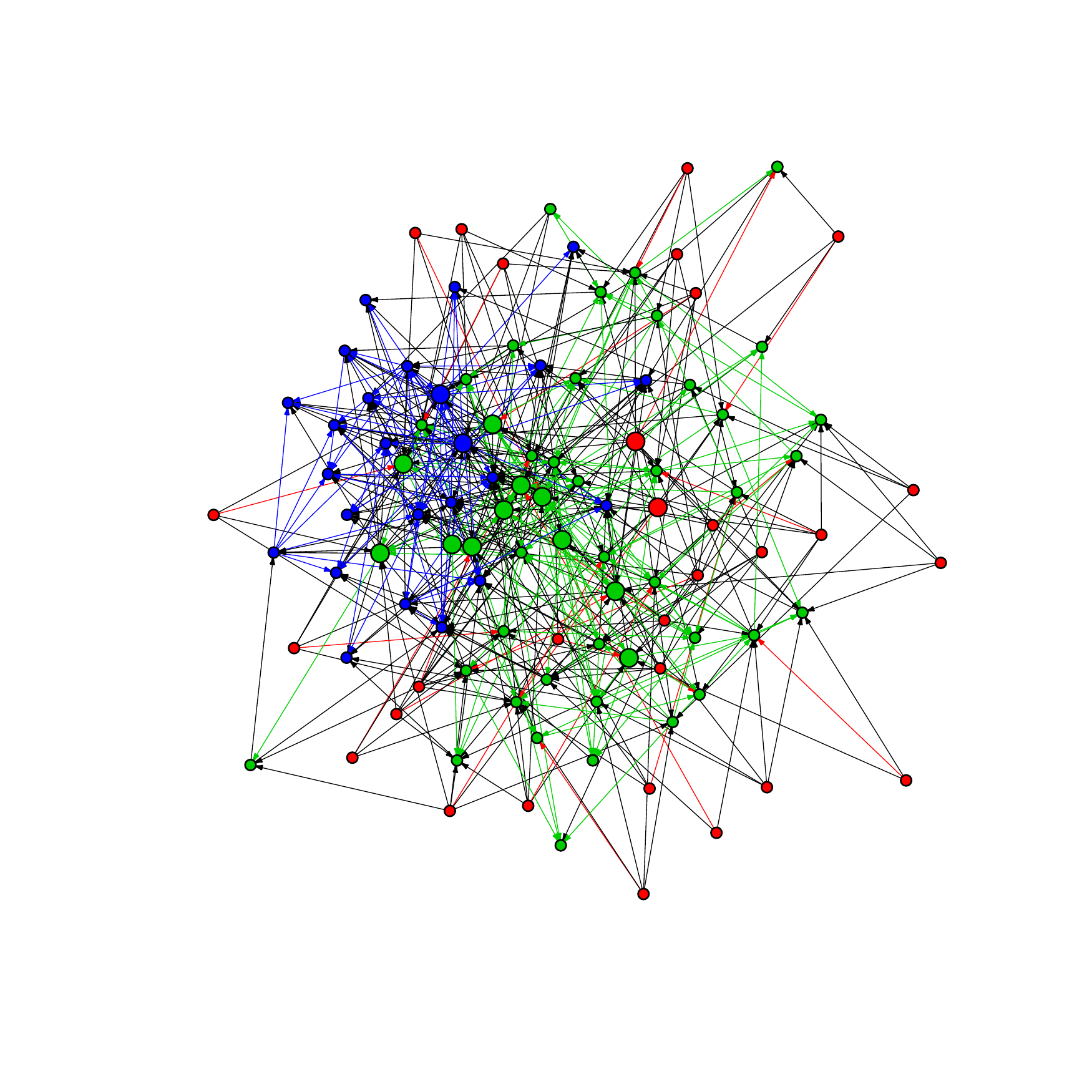}
\caption{Dense Graph with $p=104, |\mathbb{E}|=527, |\mathbb{VS}|=1675$, Moral $|\mathbb{E}|=1670$.  \label{fig:dense}}
\end{center}
\end{figure}

\begin{figure}
\begin{center}
\includegraphics[scale=0.8]{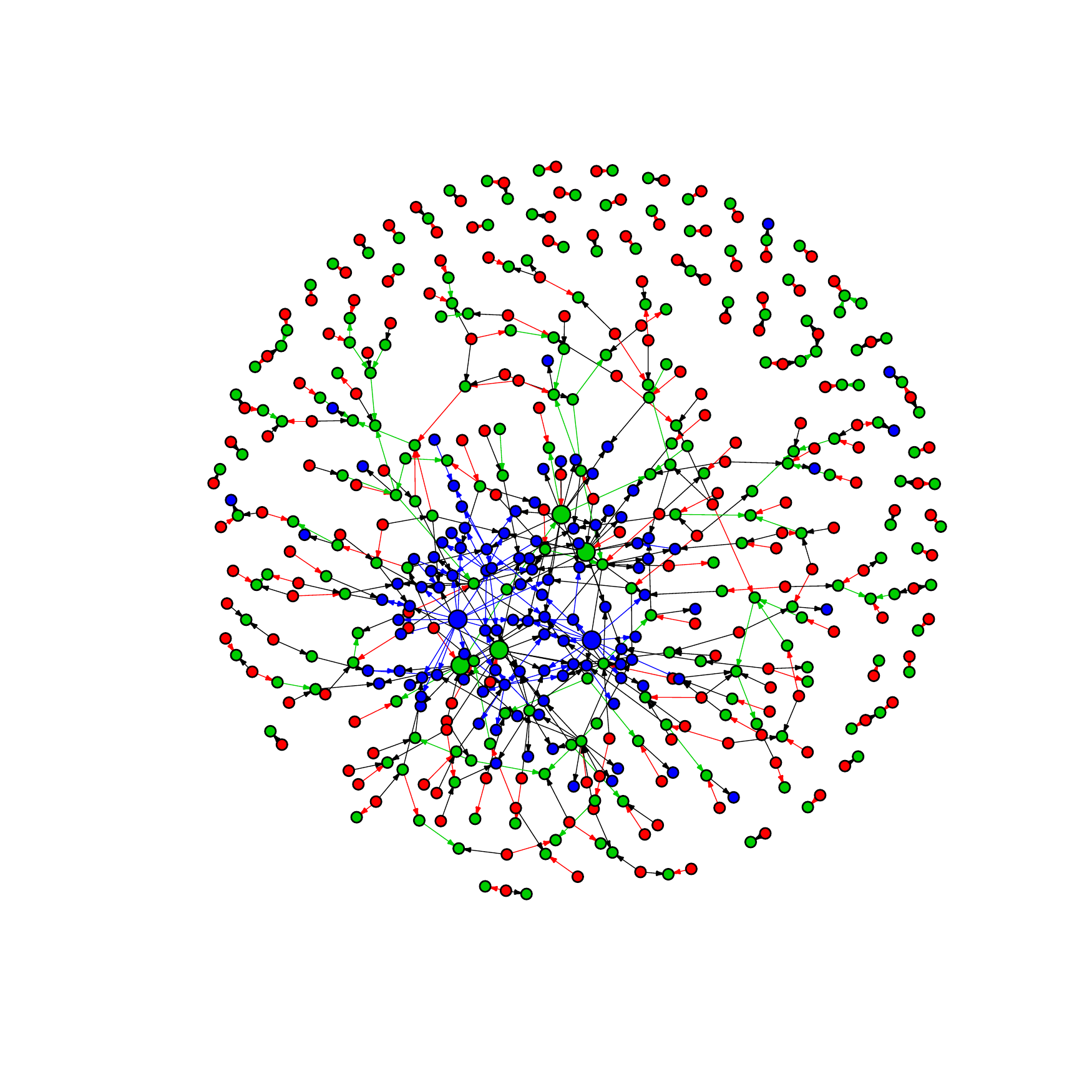}
\caption{Large Graph with $p=504, |\mathbb{E}|=515, |\mathbb{VS}|=307$, Moral $|\mathbb{E}|=808$. \label{fig:large}}
\end{center}
\end{figure}

%\begin{figure}
%\begin{center}
%\includegraphics[scale=0.8]{large_network_lessv.pdf}
%\caption{Large Graph with less v-structures, $p=511, |\mathbb{E}|=445, |\mathbb{VS}|=96$, Moral $|\mathbb{E}|=540$. %\label{fig:large_lessv}}
%\end{center}
%\end{figure}

\begin{figure}
\begin{center}
\includegraphics[scale=0.6]{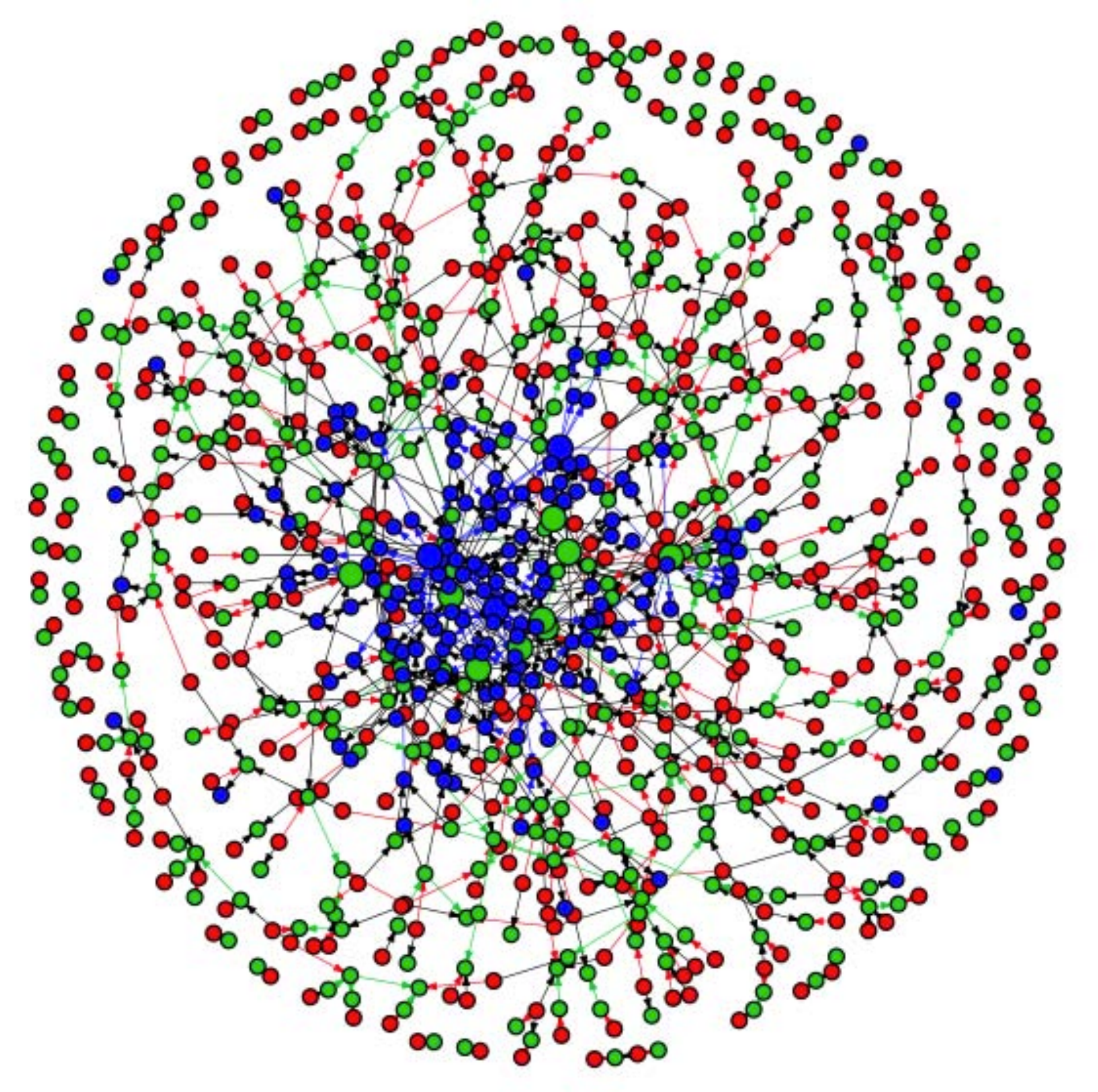}
\caption{Extra-large Graph, $p=1000, |\mathbb{E}|=1068, |\mathbb{VS}|=785$, Moral $|\mathbb{E}|=1823$. \label{fig:extra_large}}
\end{center}
\end{figure}

\begin{figure}
\begin{center}
\includegraphics[scale=0.8]{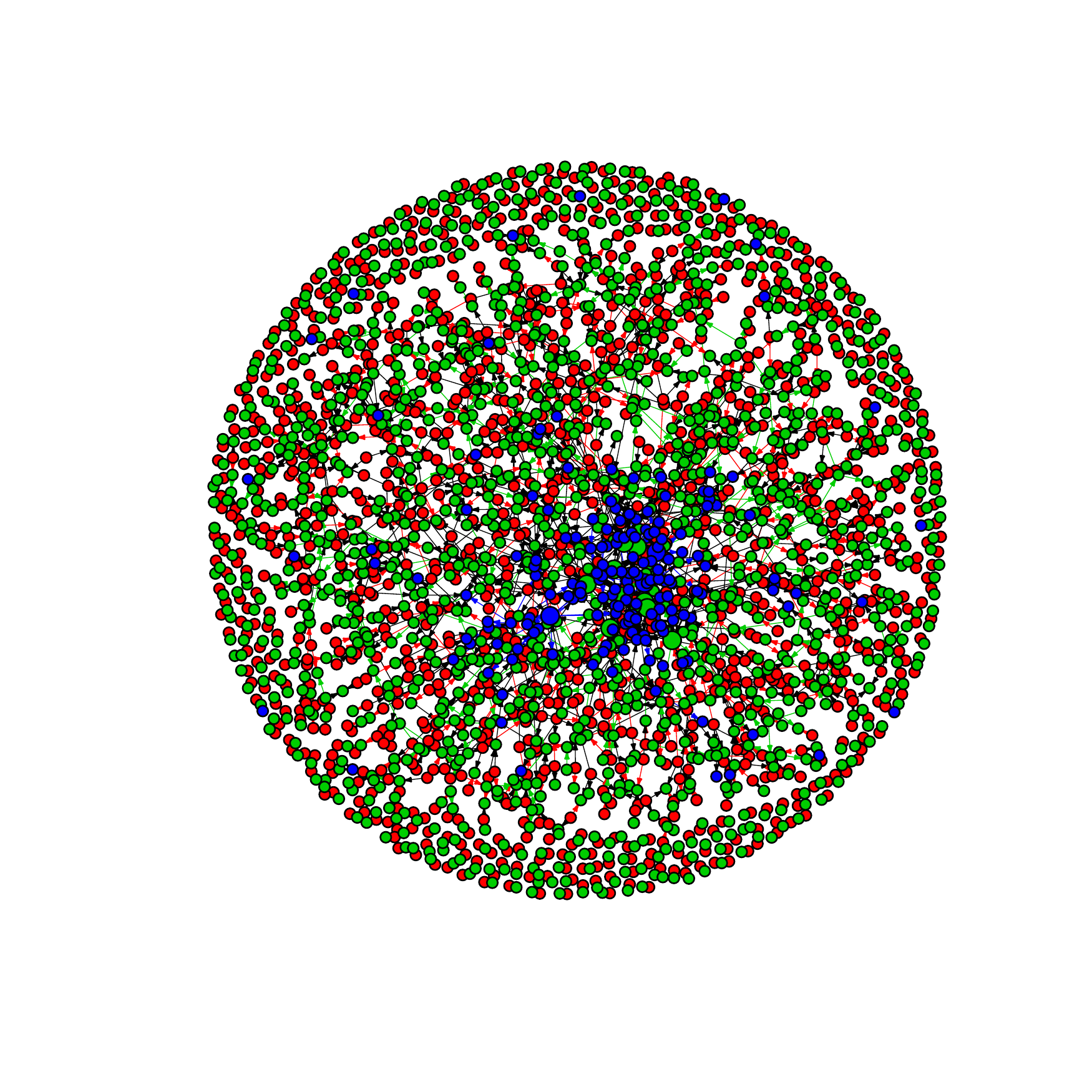}
\caption{Ultra-large Graph, $p=2639, |\mathbb{E}|=2603, |\mathbb{VS}|=1899$, Moral $|\mathbb{E}|=4481$. \label{fig:ultra_large}}
\end{center}
\end{figure}

\clearpage
\newpage
\subsection*{Learning curves}

\begin{figure}
\begin{center}
\includegraphics[scale=0.4]{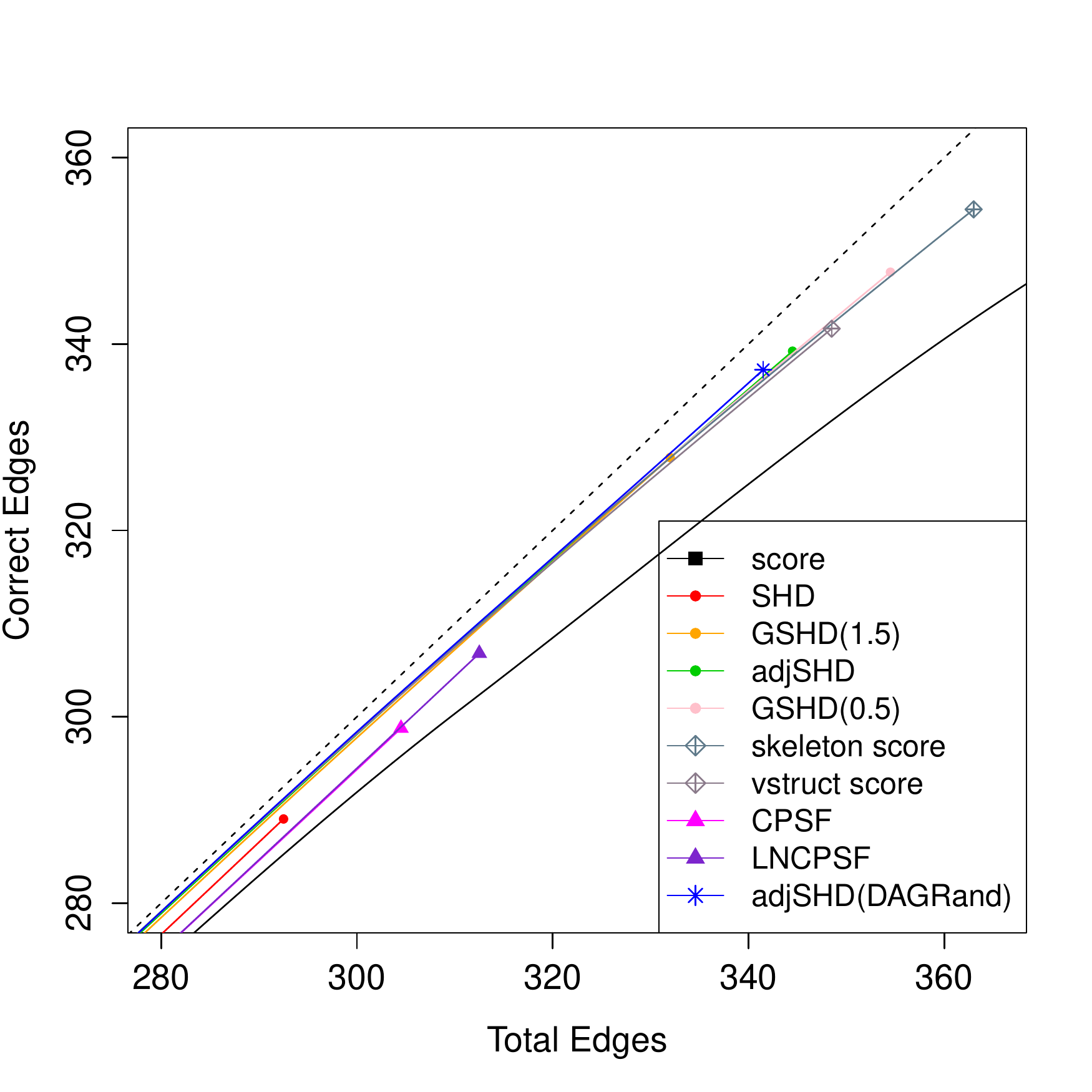}
\includegraphics[scale=0.4]{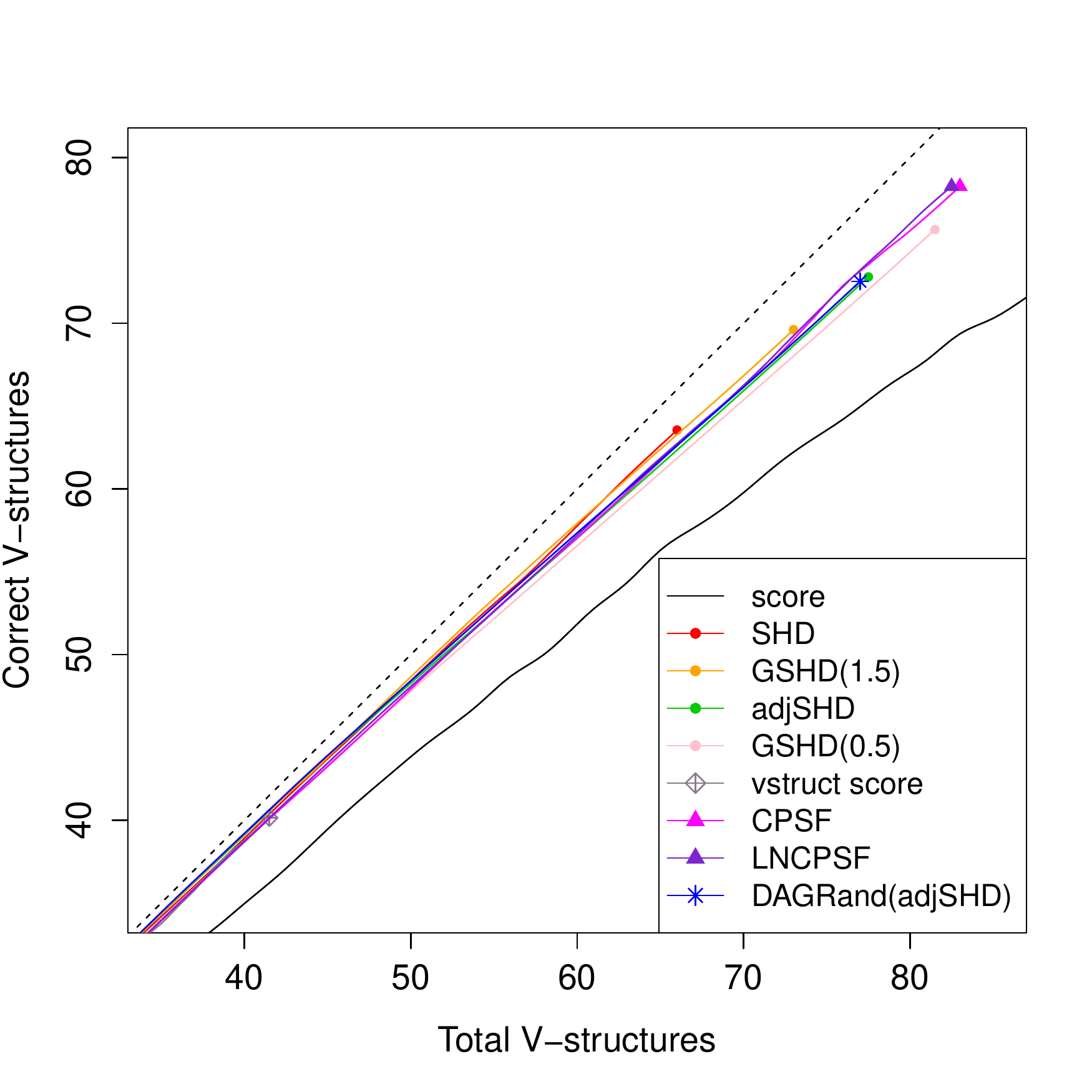}
\caption{Learning Curves for \texttt{Large Graph} with $p=504$ nodes, $|\mathbb{E}|=515$ edges, $|\mathbb{VS}|=307$ v-structures and sample size $n=100$, SNR $\in[0.5, 1.5]$. Skeleton detection (left): \texttt{score} stops at $(990, 430)$; v-structure detection (right): \texttt{score} stops at $(1151,123)$. Dotted line indicates the diagonal line where $\#$ of total $= \#$ of correct. \label{fig:dense_n100_0515_lc_chp2}}
\end{center}
\end{figure}

\begin{figure}
\begin{center}
\includegraphics[scale=0.4]{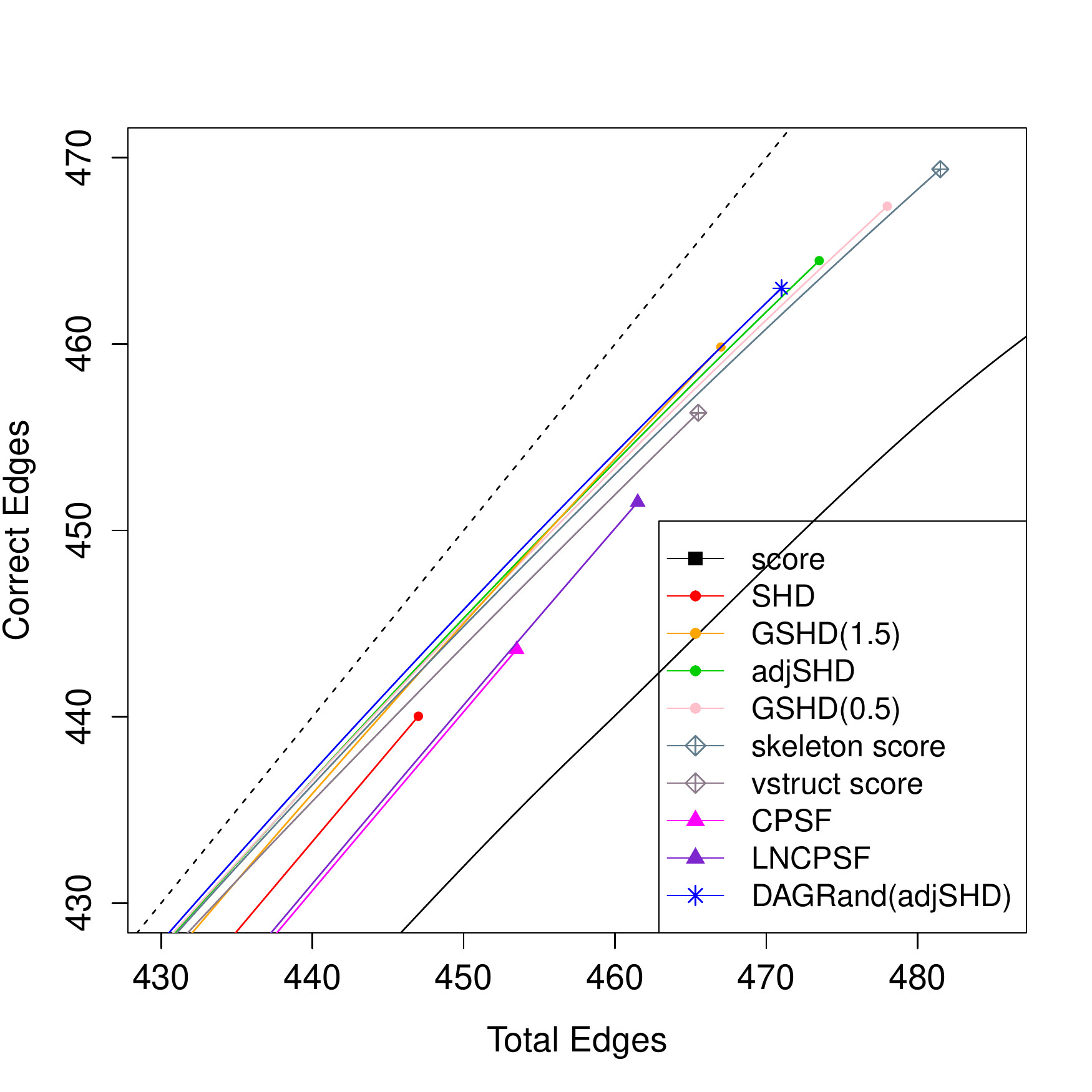}
\includegraphics[scale=0.4]{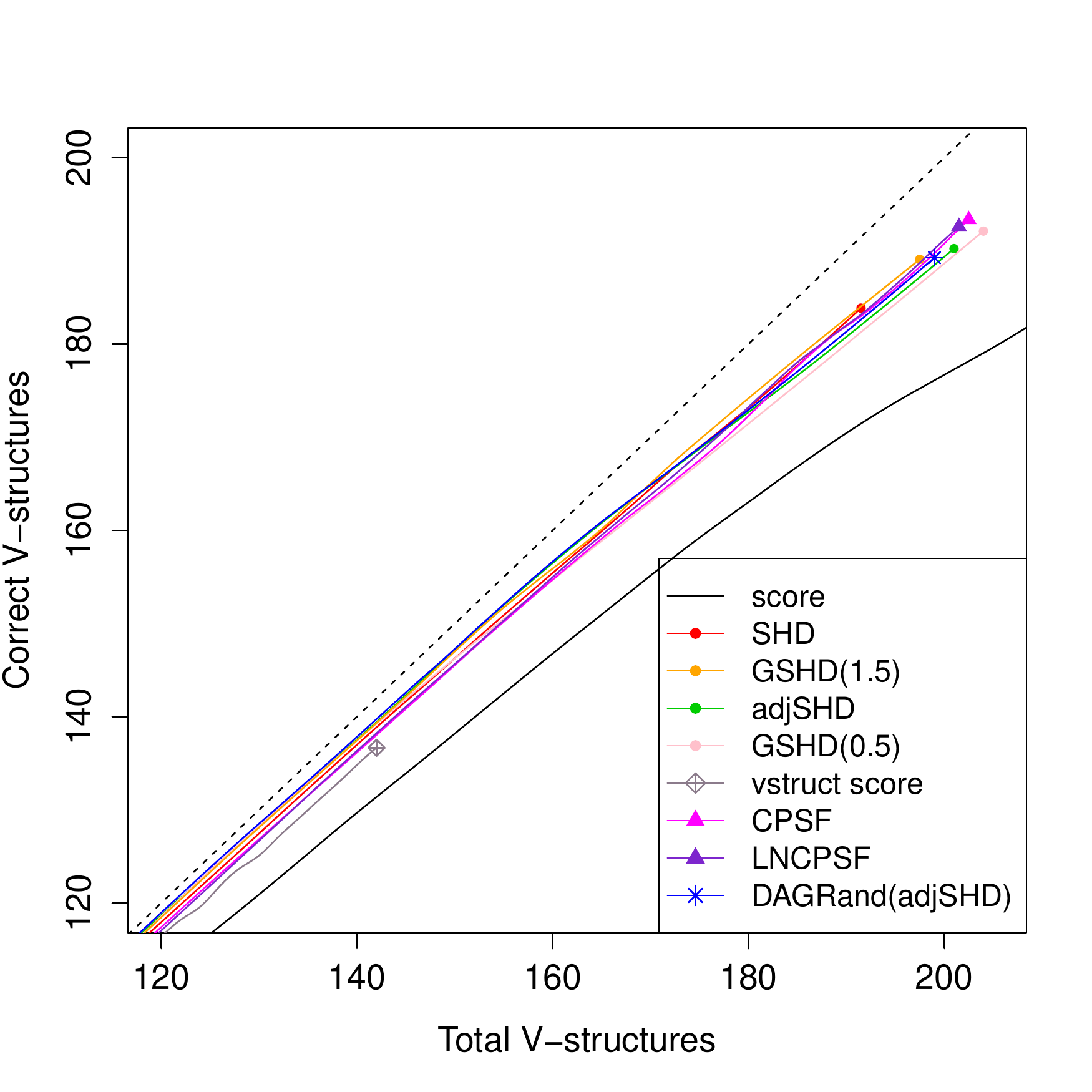}
\caption{Learning Curves for \texttt{Large Graph} with $p=504$ nodes, $|\mathbb{E}|=515$ edges, $|\mathbb{VS}|=307$ v-structures and sample size $n=250$, SNR $\in[0.5, 1.5]$. Skeleton detection (left): \texttt{score} stops at $(971, 494)$; v-structure detection (right): \texttt{score} stops at $(992,212)$. Dotted line indicates the diagonal line where $\#$ of total $= \#$ of correct.  \label{fig:dense_n250_0515_lc_chp2}}
\end{center}
\end{figure}

\begin{figure}
\begin{center}
\includegraphics[scale=0.4]{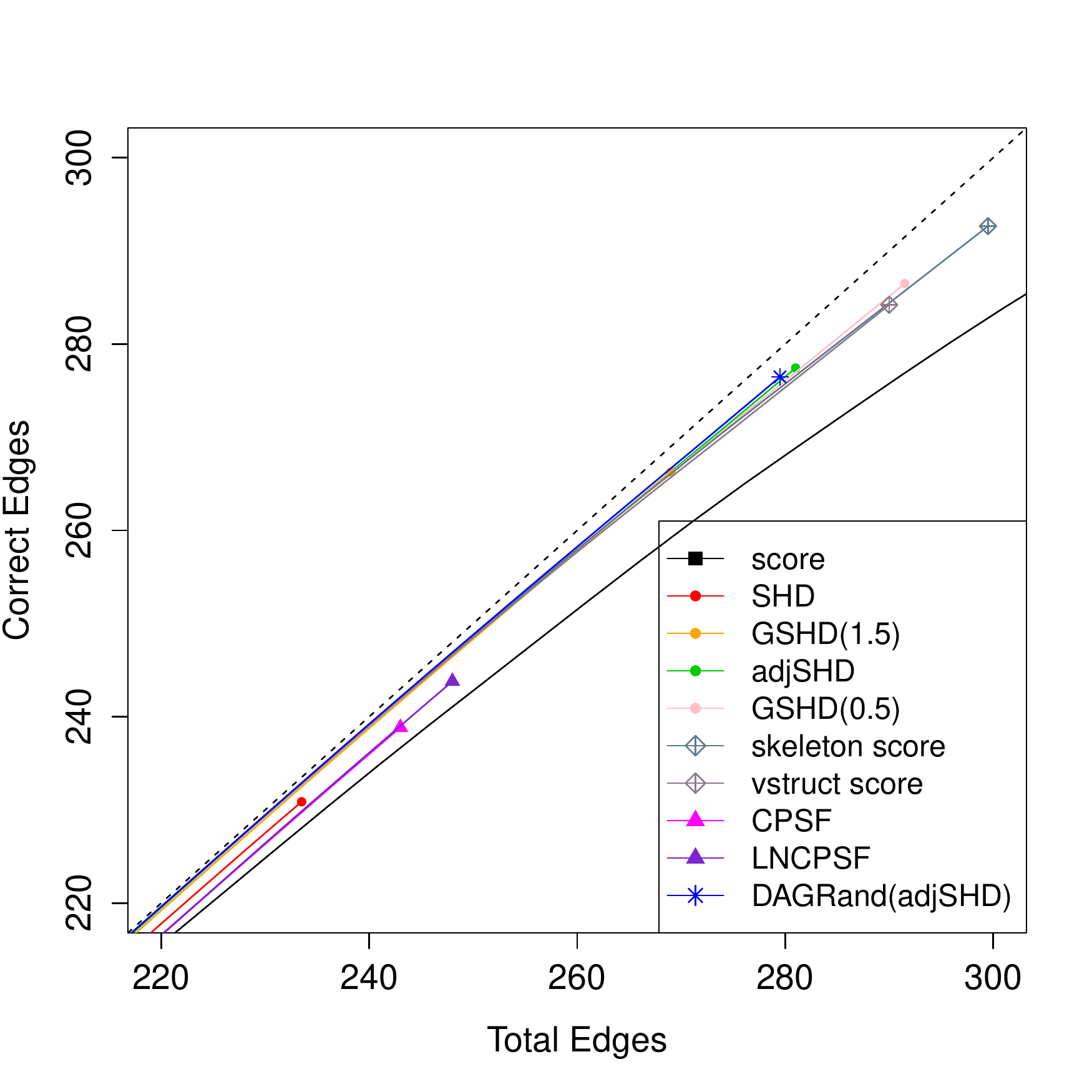}
\includegraphics[scale=0.4]{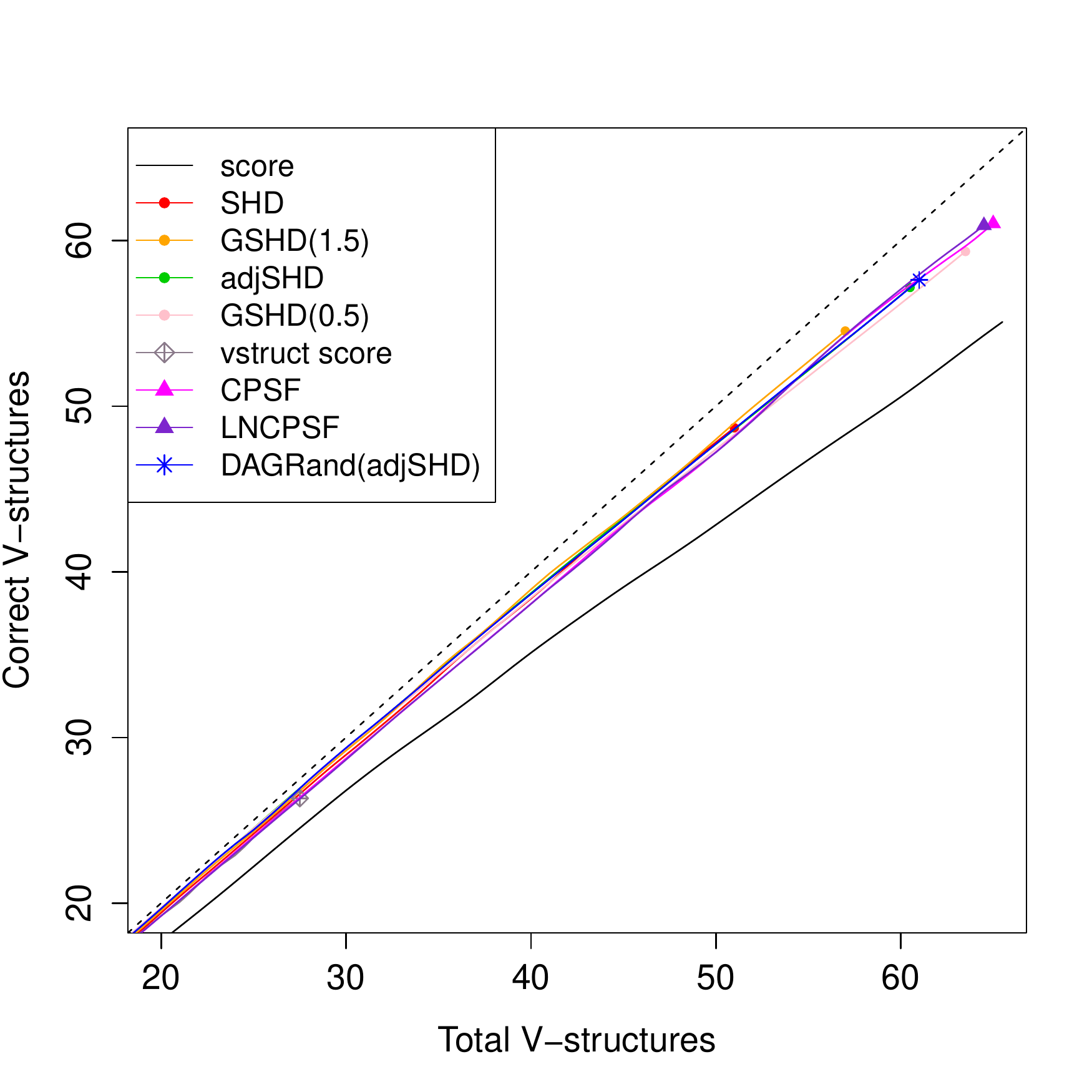}
\caption{Learning curves for Large Graph with $p=504, |\mathbb{E}|=515, n=100$, SNR $\in[0.2, 0.5]$, skeleton edge curves (left), v-structure curves (right), \texttt{score} stops at $(991.51,371.45)$ on skeleton edge curve and $(1232.6,94.71)$ on v-structure curve.\label{fig:dense_n100_0205_lc}}
\end{center}
\end{figure}

\begin{figure}
\begin{center}
\includegraphics[scale=0.4]{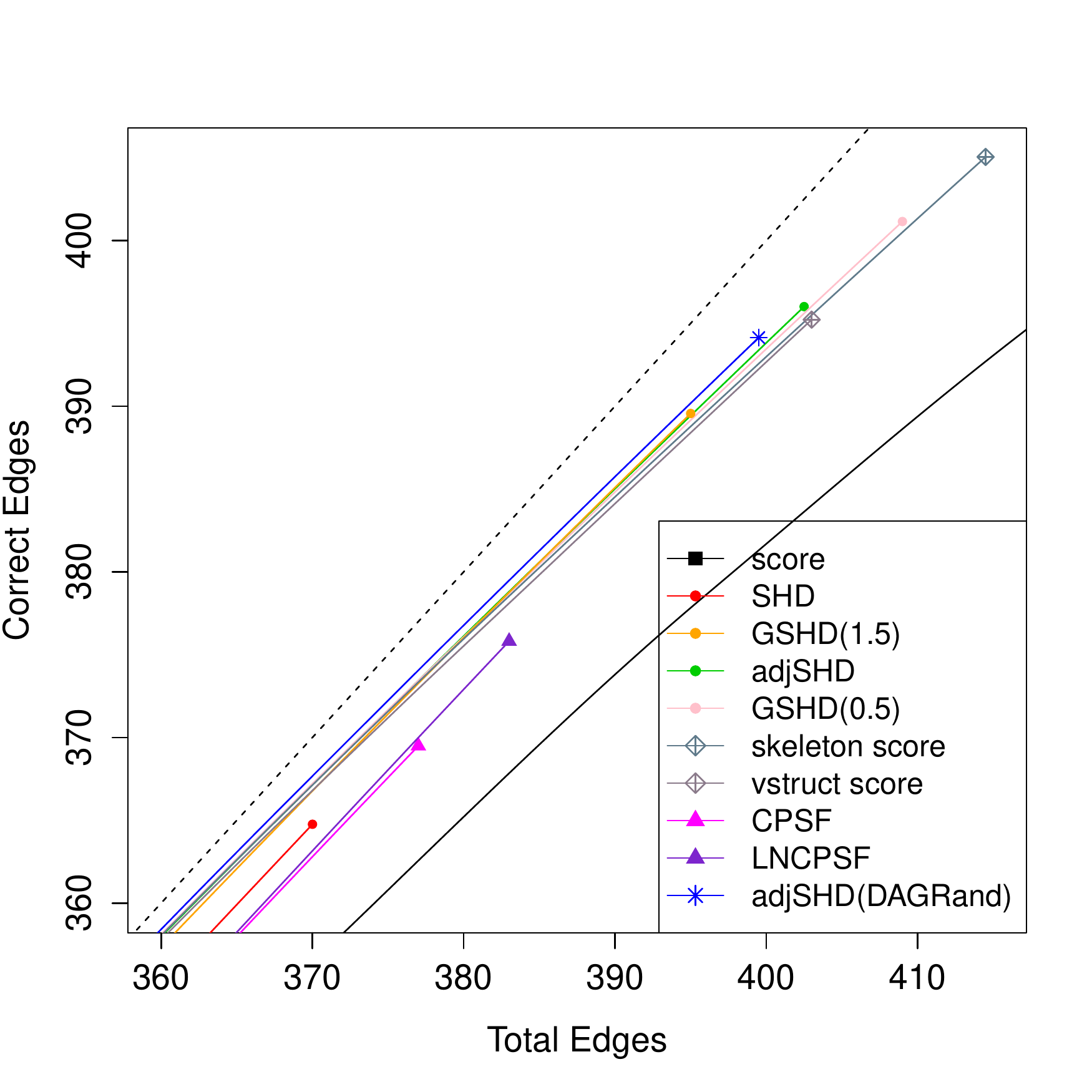}
\includegraphics[scale=0.4]{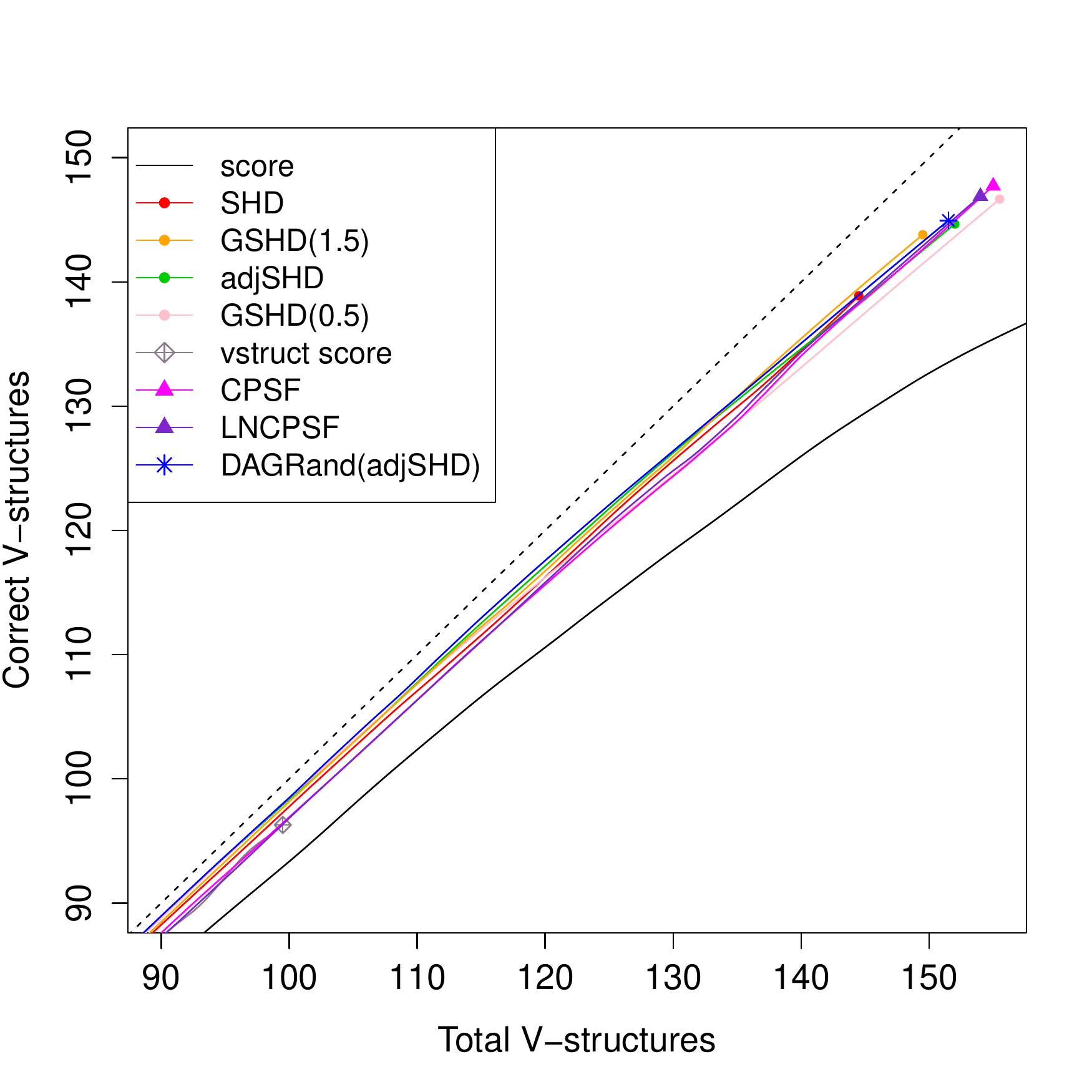}
\caption{Learning curves  for Large Graph with $p=504, |\mathbb{E}|=515, n=250$, SNR $\in[0.2, 0.5]$, skeleton edge curves (left), v-structure curves (right), \texttt{score} stops at $(978.49,451.3)$ on skeleton edge curve and $(1036.51,165.86)$ on v-structure curve. \label{fig:dense_n250_0205_lc}}
\end{center}
\end{figure}

\begin{figure}
\begin{center}
\includegraphics[scale=0.4]{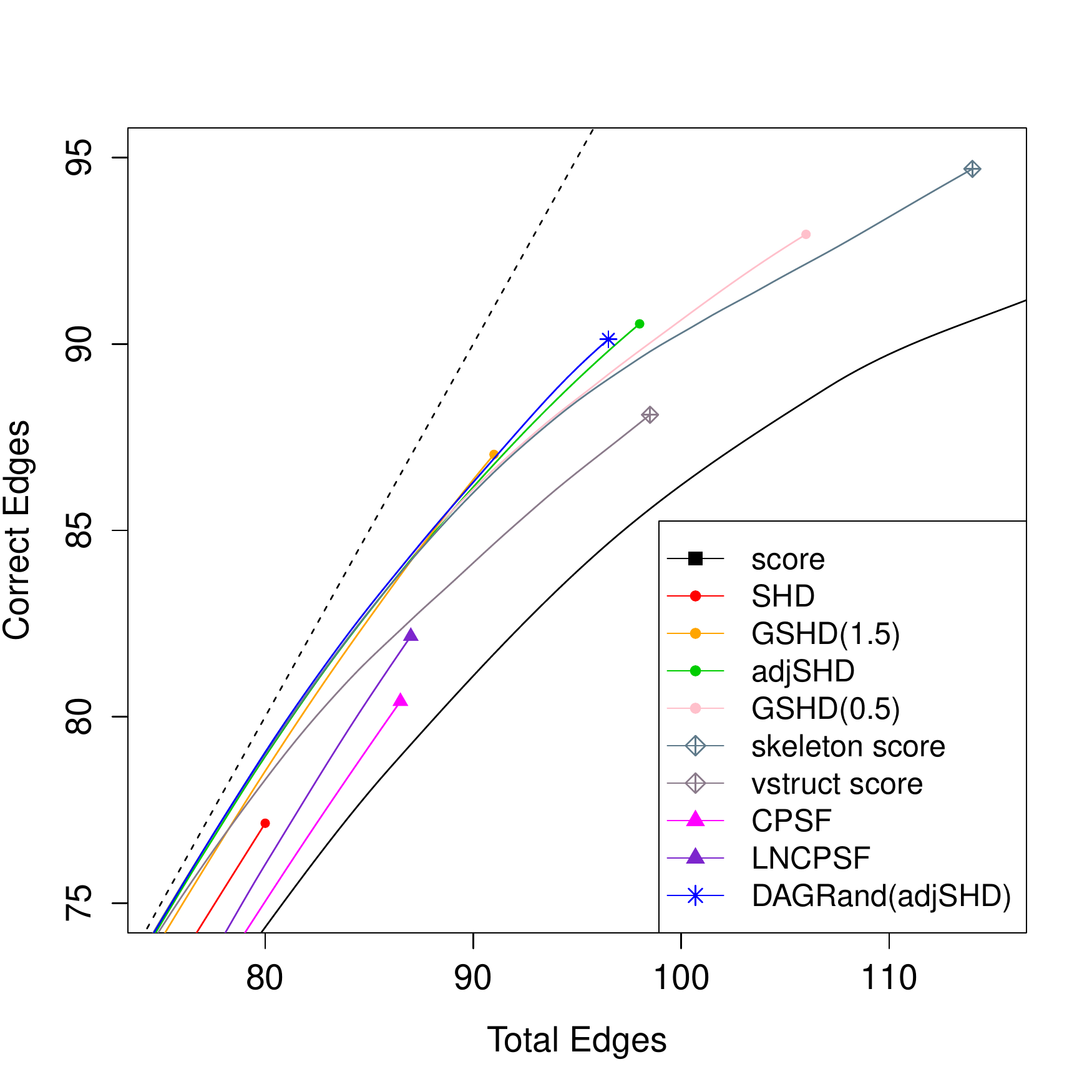}
\includegraphics[scale=0.4]{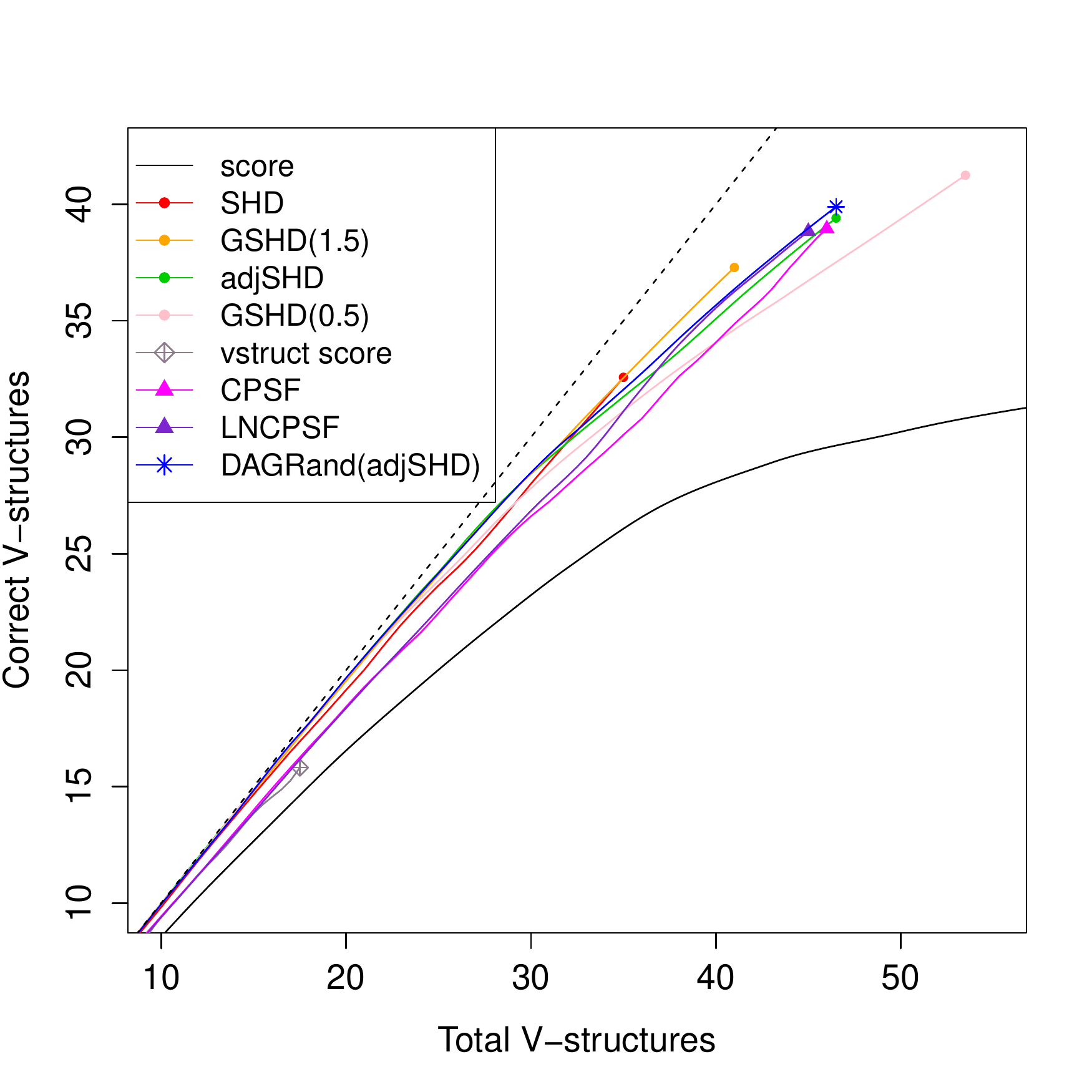}
\caption{Learning curves for Sparse Graph with $p=102, |\mathbb{E}|=109, n=102$, SNR $\in[0.5,1.5]$, skeleton edge curves (left), v-structure curves (right), \texttt{score} stops at $(336.79, 99.77)$ on skeleton edge curve and $(672.46,36.07)$ on v-structure curve. \label{fig:sparse_lc}}
\end{center}
\end{figure}

\begin{figure}
\begin{center}
\includegraphics[scale=0.4]{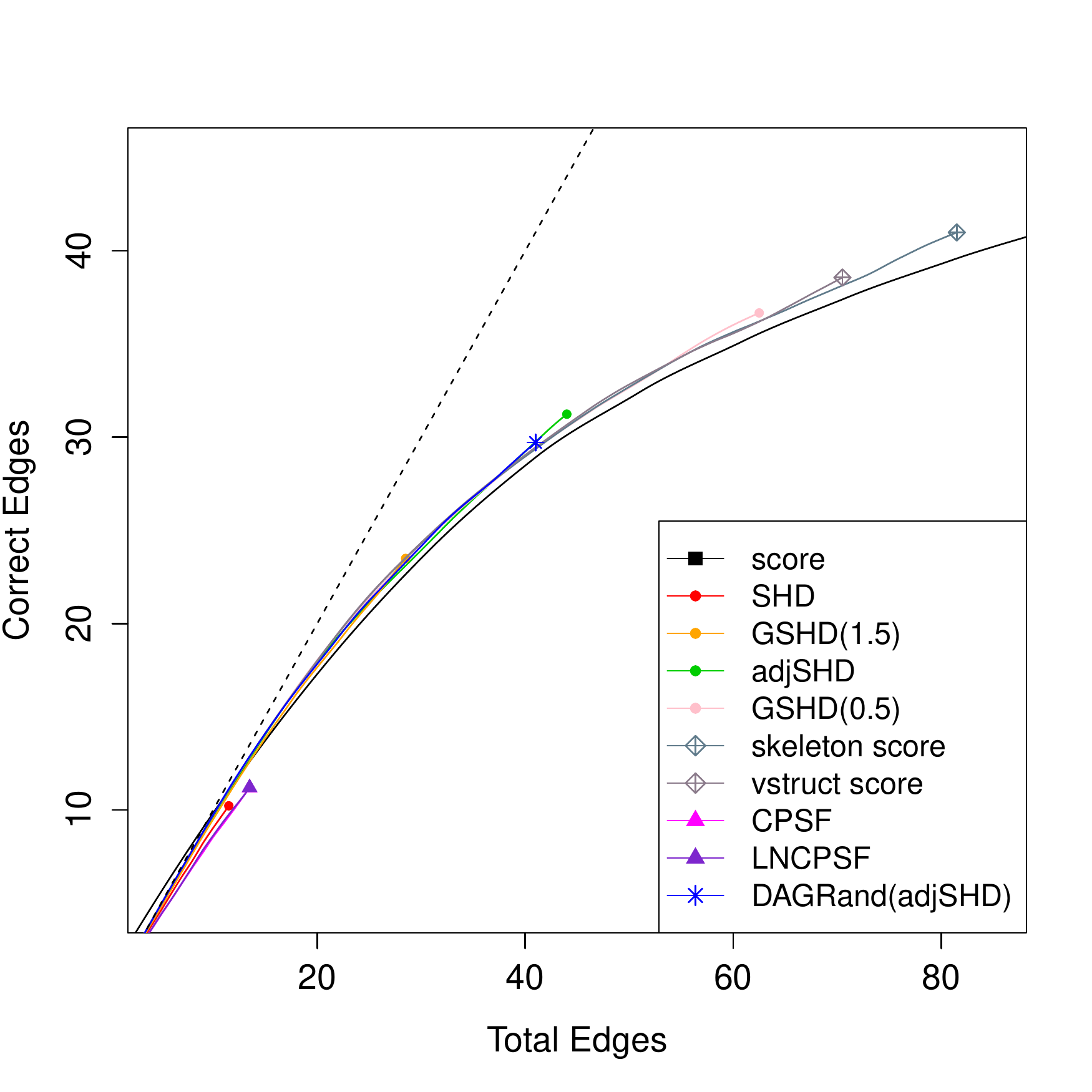}
\caption{Skeleton edge learning curve for Sparse Graph with $p=102, |\mathbb{E}|=109, n=102$, SNR $\in[0.2, 0.5]$,  \texttt{score} stops at $(307.6,58.08)$.  \label{fig:sparse_0205_lc}}
\end{center}
\end{figure}

\begin{figure}[htbp]
\begin{center}
\includegraphics[scale=0.4]{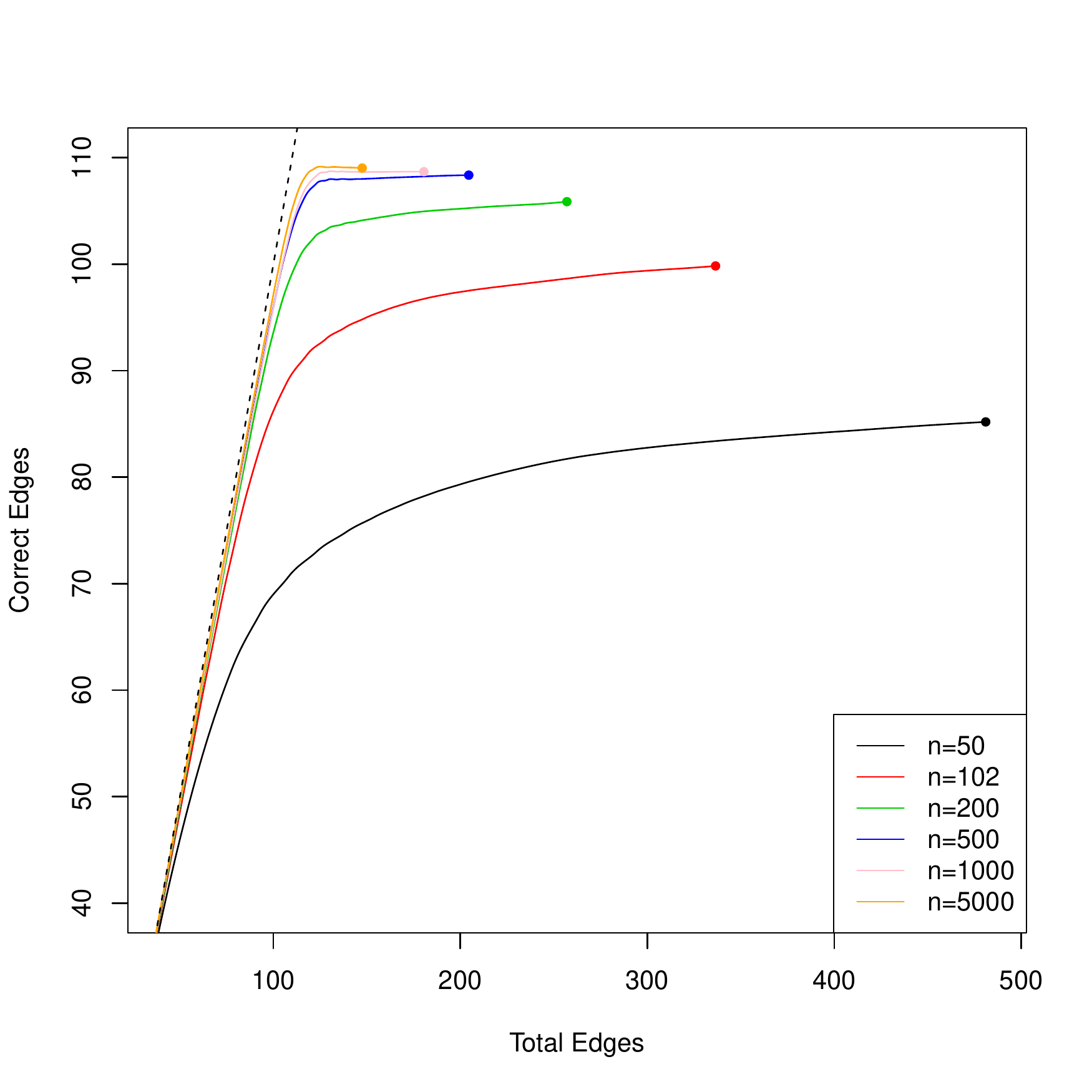}
\includegraphics[scale=0.4]{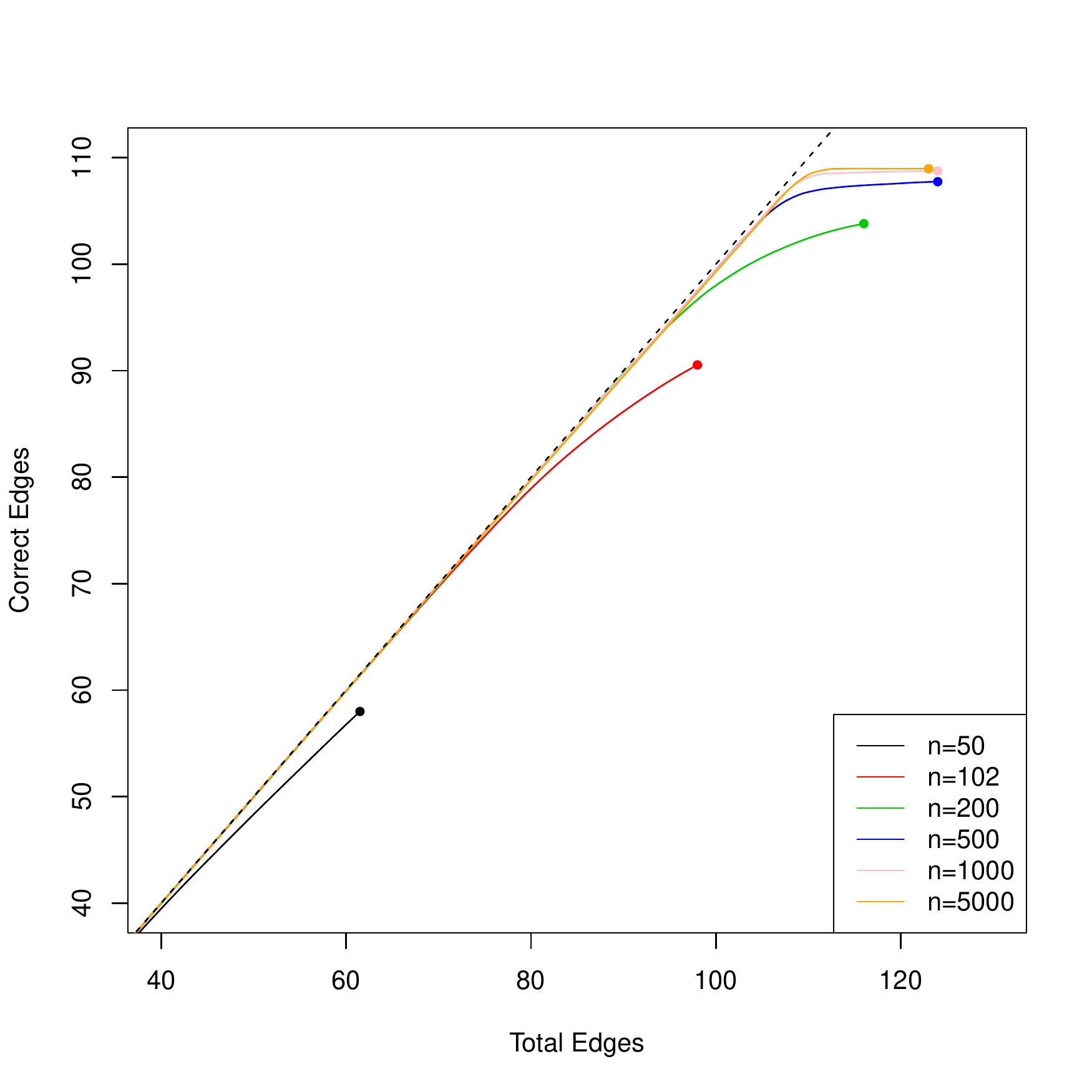}
\caption{Skeleton edge learning curves of \texttt{score}  (left) and \texttt{adjSHD} (right) under Sparse Graph with $p=102, |\mathbb{E}|=109, \text{SNR} \in [0.5, 1.5]$, and  $n=50, 102, 200, 500, 1000, 5000$.\label{adjshd_differ_size}}
\end{center}
\end{figure}

\begin{figure}
\begin{center}
\includegraphics[scale=0.4]{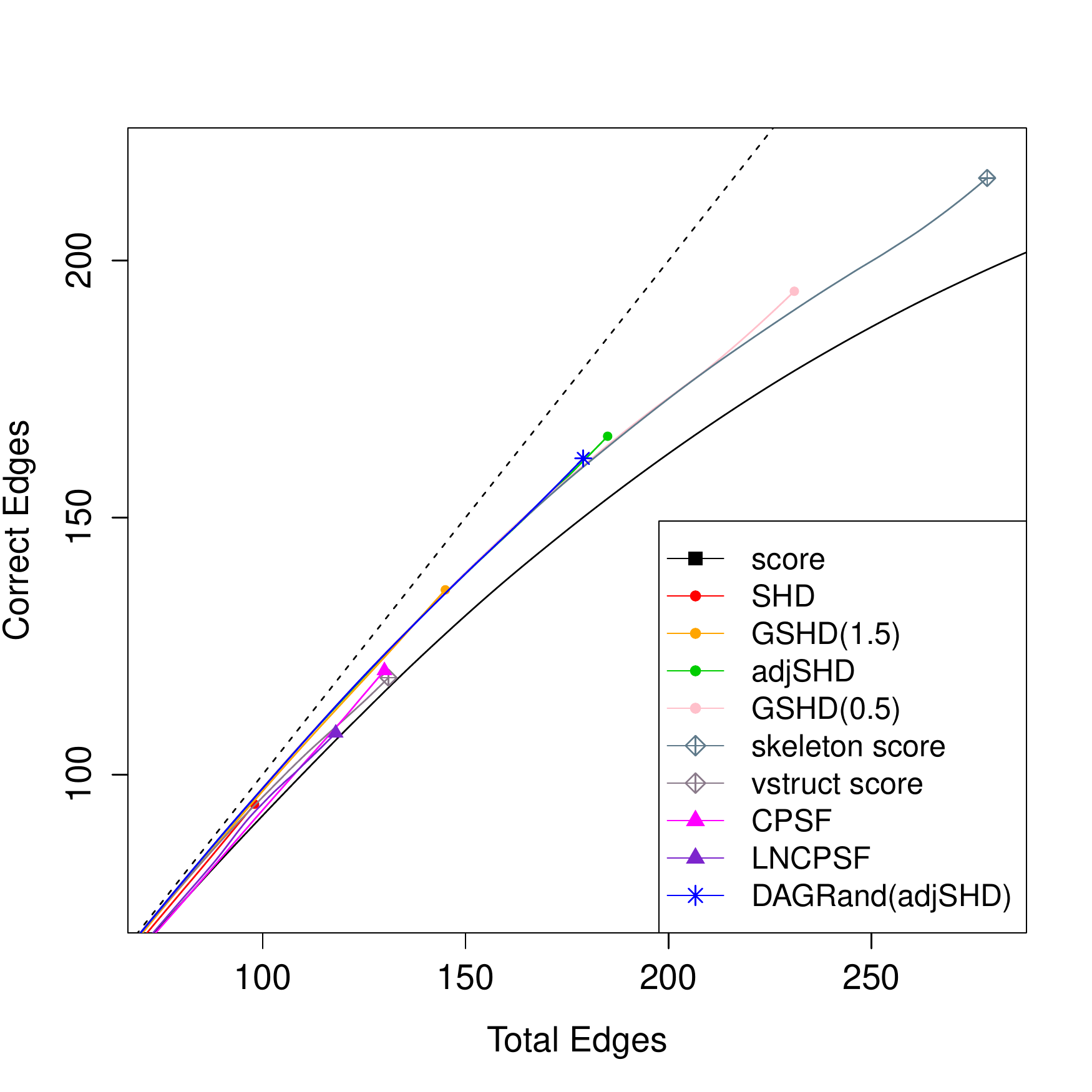}
\includegraphics[scale=0.4]{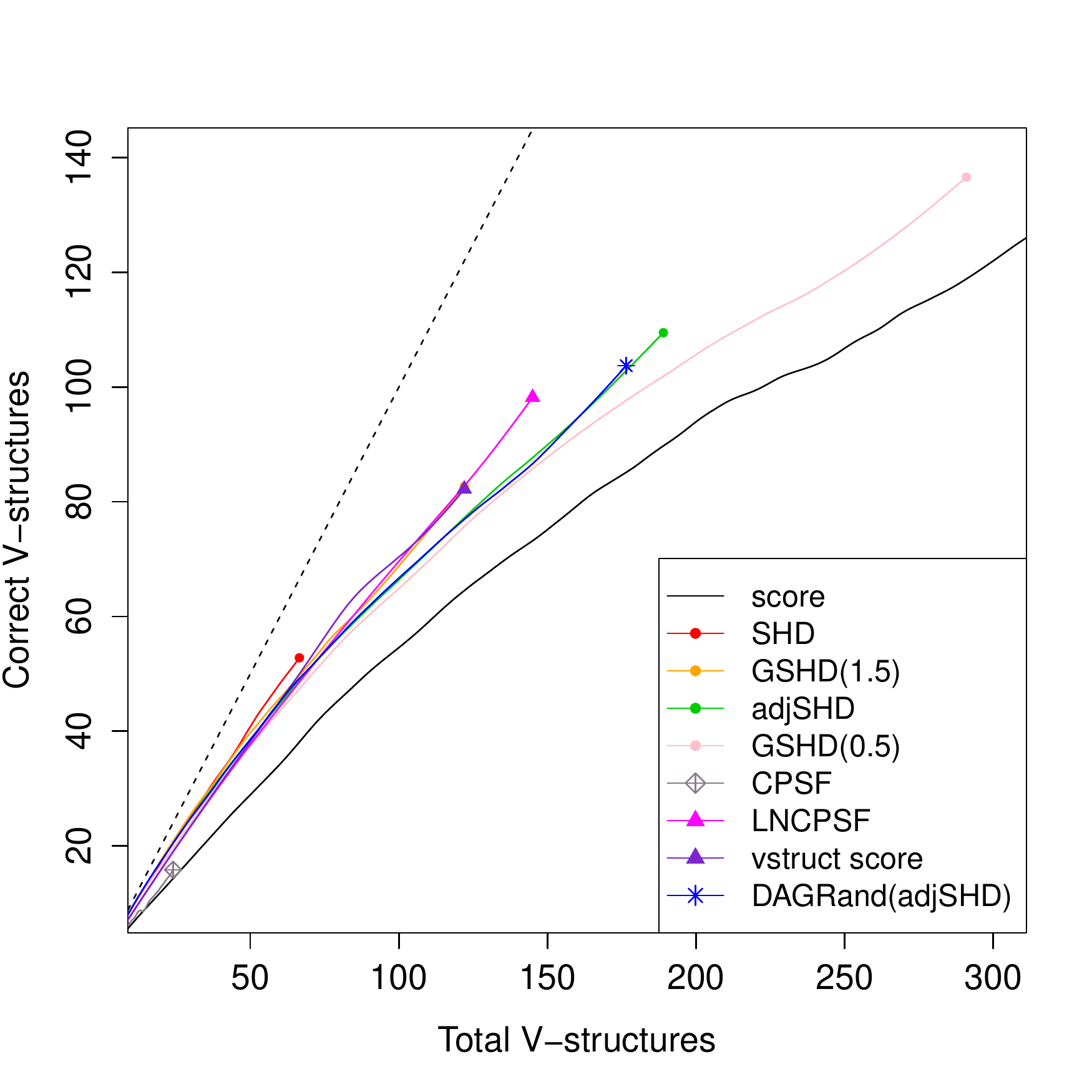}
\caption{Learning curves for Dense Graph with $p=104, |\mathbb{E}|=527, n=100$, SNR $\in[0.5, 1.5]$, skeleton edge curves (left), v-structure curves (right), \texttt{score} stops at $(524.69,254.70)$ on skeleton edge curve and $(1541.32,206.58)$ on v-structure curve. \label{fig:dense_0515_lc}}
\end{center}
\end{figure}

\begin{figure}
\begin{center}
\includegraphics[scale=0.4]{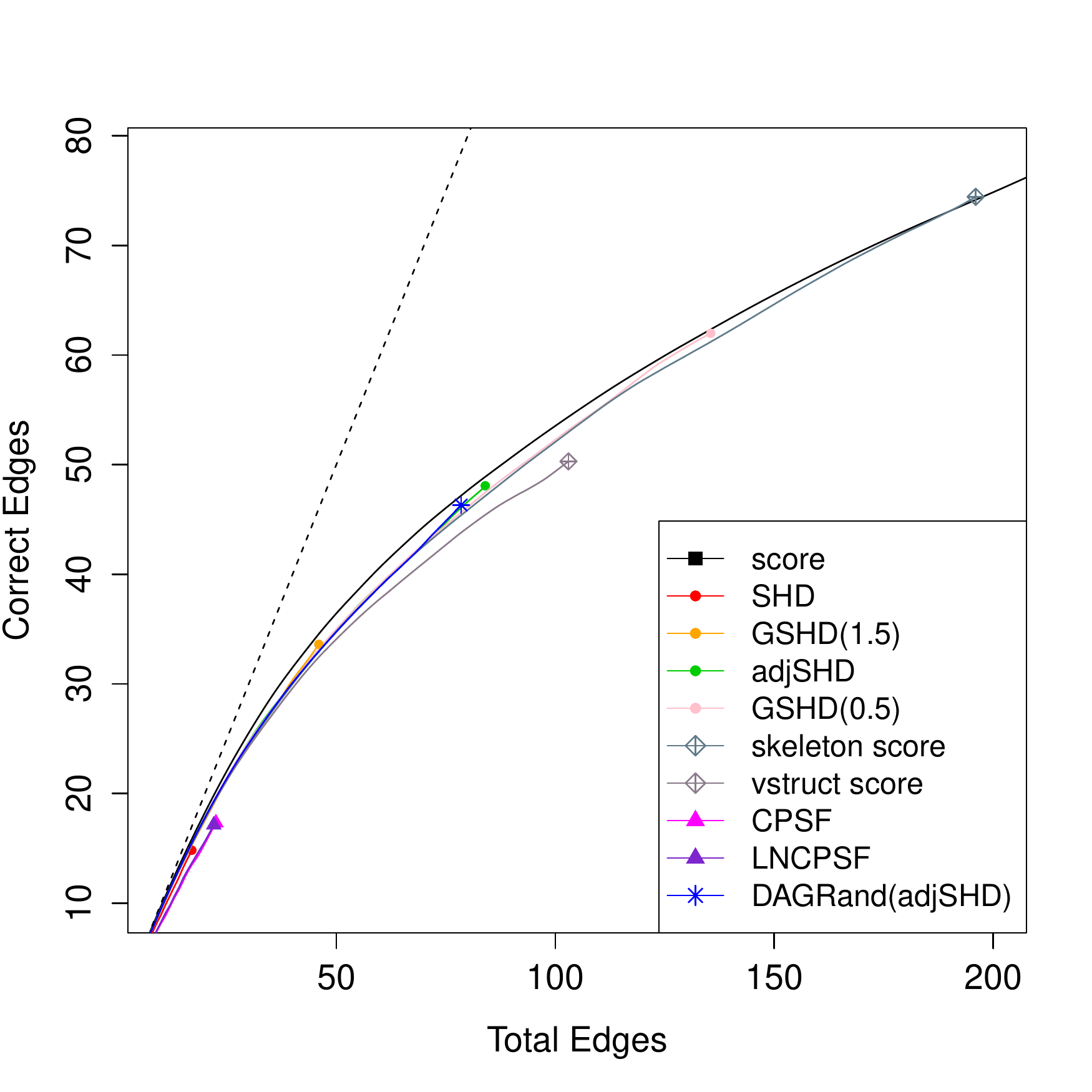}
\caption{Learning Curves  for Dense Graph with $p=104, |\mathbb{E}|=527, n=100$, SNR $\in[0.2, 0.5]$, skeleton edge curves (left), v-structure curves (right), \texttt{score} stops at $(343.9,95.63)$  on skeleton edge curve.\label{fig:dense_0205_lc}}
\end{center}
\end{figure}
\clearpage
\newpage
\renewcommand{\thesection}{C-\arabic{section}}
\renewcommand{\theequation}{C-\arabic{equation}}
\renewcommand{\thetable}{C-\arabic{table}}
\renewcommand{\thefigure}{C-\arabic{figure}}
\setcounter{figure}{0}
\setcounter{table}{0}
\setcounter{section}{0}
\section{Tables}
\label{sec:appC}
\begin{remark}
Results are averaged over $100$ independent replicates.  Numbers in parenthesis are standard deviations.
\end{remark}
\begin{remark}

\begin{itemize}
\item ``Total E" -- number of skeleton edges in the learned graph; ``Correct E" -- number of correctly identified skeleton edges in the learned graph;
\item ``Total V" -- number of v-structures in the learned graph; ``Correct V" -- number of correctly identified v-structures in the learned graph;

\item ``Total M" -- number of edges in the corresponding moral graph of the learned graph; ``Correct M" -- number of correctly identified moral edges by the learned graph;
    \item $p$: number of nodes; $|\mathbb{E}|$: number of edges; $n$: sample size; SNR: signal-to-noise-ratio;
    \item ``$100$ bootstrap resamples": aggregation based on bootstrap resamples;
    %\texttt{DAGRand}: aggregation by \texttt{DAGRand} procedure based on bootstrap resamples;
    ``Independent data": aggregation based on independent data sets (replicates).
\end{itemize}
\end{remark}

\begin{table}[H]%%10 replicates
\begin{center}
\renewcommand{\arraystretch}{0.8}
\caption{Empty Graph with $p=1000, n=250$. \label{tab:empty_network_n250}} %%For score use number of step=2000
\begin{tabular}{rrrrrrr}
\hline\hline
 &Stop/Tuning& Correct E& Total E& Correct V& Total V\\
  \hline
score&N.A.&0(0)&1996.1(1.2)&0(0)&3105.3(102)\\\hline
SHD&N.A.&0(0)&0.1(0.32)&0(0)&0(0)\\
adjSHD&N.A.&0(0)&6(2.31)&0(0)&0(0)\\
%CPSF&N.A.&0(0)&0(0)&0(0)&0(0)\\
%LN.CPSF&N.A.&0(0)&0(0)&0(0)&0(0)\\
\hline
MMHC&5e-04&0(0)&214.1(10.52)&0(0)&21(3.06)\\
&0.005&0(0)&1333.1(18.5)&0(0)&814.8(15.87)\\
&0.05$^{\ast}$&0(0)&4199.6(45.7)&0(0)&9821.1(248.11)\\\hline
PC-Alg&5e-04&0(0)&203.6(8.6)&0(0)&38.2(5.39)\\
 &0.005$^{\ast\ast}$&0(0)&946.5(12.47)&0(0)&560.8(22.44)\\
 &0.01$^{\ast\ast}$&0(0)&1433.1(16.43)&0(0)&1231.9(29.4)\\\hline\hline
 \end{tabular}
 \\ \scriptsize{$^\ast$ default value set in \texttt{bnlearn}; $^{\ast\ast}$ values suggested in \cite{kalisch2007estimating}}
\end{center}
\end{table}

\begin{table}[H]%%100 replicates
\renewcommand{\arraystretch}{0.8}
\begin{center}
\caption{Tree Graph with $p=100, |\mathbb{E}|=99$, $ n=50$,  SNR $\in [0.5, 1.5]$. \label{tab:tree_n50_indep_boot}}
\begin{tabular}{rrrrr}
\hline\hline
& Correct E & Total E & Correct V & Total V \\
\hline
score&89.94(2.73)&473.72(8.07)&0&3103.41(543.52)\\
MST$^\ast$&90.44(2.33)&99(0)&0&0\\
\hline
 \multicolumn{5}{c}{100 bootstrap resamples}\\
\hline
SHD(2)&66.08(3.79)&67.58(3.96)&0&2.22(1.44)\\
GSHD(1.5)&75.03(3.41)&77.27(3.6)&0&5.06(2.25)\\
adjSHD(1)&78.68(3.26)&82.26(3.47)&0&7.43(2.69)\\
GSHD(0.5)&81.68(3.02)&87.17(3.21)&0&10.14(3)\\
%skeleton score&83.85(2.81)&91.82(3.68)&NA&NA\\
%CPSF&65.8(3.9)&68.28(4.19)&0&3.05(1.79)\\
%LN.CPSF&71.44(3.82)&73.83(4.16)&0&3.07(1.88)\\
\hline
\multicolumn{5}{c}{Independent data, 100 replicates}\\
\hline
SHD(2)&68&68&0&0\\
GSHD(1.5)&98&98&0&3\\
adjSHD(1)&99&99&0&4\\
GSHD(0.5)&99&99&0&4\\
%skeleton score&99&99&NA&NA\\
%CPSF&68&68&0&0\\
%LN.CPSF&92&92&0&0\\
\hline
\hline
\end{tabular}
\\$^\ast$ minimum spanning tree
\end{center}
\end{table}

\begin{comment}
\begin{table}[H]%%100 replicates
\renewcommand{\arraystretch}{0.8}
\begin{center}
\caption{Tree Graph with $p=100, |\mathbb{E}|=99, n=100,$ SNR $\in [0.5, 1.5]$. \label{tab:tree_n100_indep_boot}}
\begin{tabular}{rrrrr}
\hline\hline
& Correct E & Total E & Correct V & Total V \\
\hline
score&96.8(1.29)&281.26(12.35)&0&398.35(52.67)\\
MST$^{\ast}$&97.41(1.16)&99(0)&0&0\\
\hline
 \multicolumn{5}{c}{100 bootstrap resamples}\\
\hline
SHD(2)&86.79(2.79)&89.45(3.27)&0&7.71(2.85)\\
GSHD(1.5)&92.4(2.06)&96.44(2.98)&0&12.36(3.47)\\
adjSHD(1)&94.01(1.82)&100.62(3.28)&0&15.91(4.14)\\
GSHD(0.5)&95.08(1.7)&105.27(3.4)&0&20.27(4.94)\\
%skeleton score&95.79(1.64)&110.1(4.06)&NA&NA\\
%CPSF&84.26(2.64)&88.85(3.32)&0&7.41(3.34)\\
%LN.CPSF&90.13(2.31)&94.23(2.86)&0&7.83(3.12)\\
\hline
\multicolumn{5}{c}{Independent data, 100 replicates}\\
\hline
SHD(2)&93&93&0&2\\
GSHD(1.5)&98&98&0&3\\
adjSHD(1)&99&99&0&5\\
GSHD(0.5)&99&99&0&4\\
%skeleton score&99&99&NA&NA\\
%CPSF&92&92&0&1\\
%LN.CPSF&99&99&0&3\\
\hline
\hline
\end{tabular}
\\$^\ast$ minimum spanning tree
\end{center}
\end{table}

\end{comment}

\begin{table}[H]%%%100 replicates
\renewcommand{\arraystretch}{0.8}
\small
\begin{center}
\caption{Sparse Graph with $p=102, |\mathbb{E}|=109$, v-structure$=77$, Moral Edge$=184$, $n=102$, SNR $\in [0.5, 1.5]$. \label{tab:indep_boot}}
\begin{tabular}{rrrrrrr}
\hline\hline
& Correct E & Total E & Correct V & Total V &correct M & Total M\\
\hline
score&99.77(2.5)&336.79(16.79)&36.07(7.41)&672.46(90.68)&155.4(7.06)&961.38(93.28)\\
\hline
 \multicolumn{7}{c}{DAGBag, bootstrap}\\
\hline
SHD(2)&77.64(4.92)&80.34(5.37)&31.68(6.56)&35.06(7)&110.12(11.11)&115.2(11.62)\\
GSHD(1.5)&86.62(3.88)&91.4(4.43)&35.11(6.95)&41.25(7.67)&122.69(10.16)&132.35(11.09)\\
adjSHD(1)&89.87(3.74)&98.43(4.38)&36.78(7.13)&46.94(8.46)&128.17(10.1)&145.02(11.77)\\
GSHD(0.5)&92.33(3.35)&106.17(4.68)&38.23(7.09)&53.64(8.72)&132.77(9.68)&159.4(12.16)\\
%skeleton score&94.26(3.26)&114.39(4.77)&NA&NA&NA&NA\\
%vstruct score&87.6(3.19)&98.59(4.03)&15.85(5.15)&17.52(5.16)&105.66(7.82)&116.11(3.19)\\
%CPSF&80.98(5.19)&86.74(6.12)&37.42(7.06)&46.16(9.03)&120.14(11.52)&132.54(14.25)\\
%LN.CPSF&82.18(4.81)&87.21(5.4)&37.44(7.03)&45.25(8.58)&120.92(10.98)&132.16(13.18)\\
\hline

\multicolumn{7}{c}{Independent data, 100 replicates}\\
\hline
SHD(2)&99&99&60&60&158&158\\
GSHD(1.5)&106&106&65&65&170&170\\
adjSHD(1)&107&107&66&66&172&172\\
GSHD(0.5)&109&109&72&72&180&180\\
%skeleton score&109&109&NA&NA&NA&NA\\
%vstruct score&99&100&39&39&139&139\\
%CPSF&101&101&66&66&166&166\\
%LN.CPSF&105&105&73&73&177&177\\
\hline
\hline
\end{tabular}
\end{center}
\end{table}

\begin{table}[H]%%100 replicates
\renewcommand{\arraystretch}{0.8}
\small
\begin{center}
\caption{Sparse Graph with $p=102, |\mathbb{E}|=109$, v-structure$=77$, Moral Edge$=184$, $n=102$,  SNR $\in [0.2, 0.5]$. \label{tab:indep_boot_0205}}
\begin{tabular}{rrrrrrr}
\hline\hline
& Correct E & Total E & Correct V & Total V & Correct M & Total M\\
\hline
score&58.08(5.22)&307.6(15.81)&4.83(2.47)&630.08(93.83)&80.37(7.43)&896.58(94.87)\\
\hline
\multicolumn{7}{c}{100 bootstrap resamples}\\
\hline
SHD(2)&10.76(2.95)&11.98(3.25)&0.17(0.4)&0.84(0.91)&11(3.08)&12.82(3.72)\\
GSHD(1.5)&23.29(3.61)&28.92(4.06)&0.45(0.63)&3.18(1.74)&24.11(3.81)&32.1(5.02)\\
adjSHD(1)&30.48(4.17)&44.27(5.03)&0.9(0.82)&8.05(3.05)&32.03(4.47)&52.32(7.26)\\
GSHD(0.5)&36.12(4.28)&62.62(6.67)&1.42(1.11)&17.44(5.33)&38.65(4.77)&80.04(11.07)\\

%skeleton score&40.43(4.21)&81.98(7.21)&NA&NA&NA&NA\\
%vstruct score&38.03(3.89)&70.6(4.8)&0.01(0.1)&0.14(0.4)&38.62(3.88)&70.74(3.89)\\
%CPSF&11.47(2.94)&13.75(3.64)&0.4(0.62)&2.95(1.65)&12.17(3.15)&16.7(4.81)\\
%LN.CPSF&11.76(3.01)&13.93(3.67)&0.39(0.62)&2.8(1.62)&12.46(3.2)&16.73(4.78)\\
\hline

\multicolumn{7}{c}{Independent data, 100 replicates}\\
\hline
SHD(2)&2&2&0&0&2&2\\
GSHD(1.5)&23&23&0&1&23&24\\
adjSHD(1)&34&34&0&1&34&34\\
GSHD(0.5)&49&49&1&3&50&52\\
%skeleton score&58&58&NA&NA&NA&NA\\
%vstruct score&51&51&0&0&51&51\\
%CPSF&2&2&0&0&2&2\\
%LN.CPSF&2&2&0&0&2&2\\
\hline
\hline
\end{tabular}
\end{center}
\end{table}

%%%%%%%%%%%%% score comparison on sparse graph
\begin{table}%[H]
\renewcommand{\arraystretch}{0.8}  %%% notes: Bge 10 replicates, need upadte. need add moral results for Bge;
\small
\begin{center}
\caption{Score comparison on Sparse Graph with $p=102, |\mathbb{E}|=109, n=102, SNR \in [0.5, 1.5]$. \label{tab:decomposable_scores}} %,p=102,E=109,Vstructure=77
\begin{tabular}{rrrrrrr}\hline\hline
 & Correct E & Total E & Correct V & Total V & Correct M & Total M\\ \hline
\multicolumn{7}{c}{\texttt{BIC}}\\\hline
score&99.77(2.5)&336.79(16.79)&36.07(7.41)&672.46(90.68)&155.4(7.06)&961.38(93.28)\\
SHD&77.64(4.92)&80.34(5.37)&31.68(6.56)&35.06(7)&110.12(11.11)&115.2(11.62)\\
adjSHD&89.87(3.74)&98.43(4.38)&36.78(7.13)&46.94(8.46)&128.17(10.1)&145.02(11.77)\\
\hline
\multicolumn{7}{c}{\texttt{like}}\\\hline
score&101.13(2.41)&498.59(1.24)&30.98(6.86)&1561.28(117.33)&162.37(6.2)&1802.02(88.53)\\
SHD&76.46(5.06)&79.46(5.42)&29.08(6.88)&32.52(7.55)&106.6(11.41)&111.84(12.13)\\
adjSHD&89.76(3.7)&98.84(4.43)&34.13(7.21)&44.39(8.27)&125.9(10.2)&142.92(11.59)\\
\hline
\multicolumn{7}{c}{\texttt{eBIC}}\\\hline
  \hline
score&83.68(4.87)&91.13(5.63)&30.74(7.61)&36.96(7.9)&119.22(12)&127.74(12.25)\\
SHD&68.33(5.54)&68.97(5.62)&25.65(6.31)&26.45(6.38)&94.37(11.44)&95.31(11.38)\\
adjSHD&78.87(4.84)&80.07(5.02)&29.63(6.68)&31.17(6.84)&109.03(11.12)&111.09(11.27)\\
\hline
\multicolumn{7}{c}{\texttt{GIC}}\\\hline
  \hline
score&97.2(3.04)&174.39(8.46)&40.59(7.44)&159.61(23.07)&147.56(8.74)&331.05(28.83)\\
SHD&77.92(4.93)&80.39(5.24)&32.24(7.11)&35.72(7.86)&110.96(11.74)&115.93(12.44)\\
adjSHD&89.33(3.61)&97.02(4.58)&36.96(6.86)&46.13(8.66)&127.81(9.85)&142.75(12.18)\\
\hline
\multicolumn{7}{c}{\texttt{BGe} with $iss=3$}\\\hline\hline
score&101.3(3.2)&218.7(14.27)&54.6(8.45)&286.7(30.68)&157.7(9.18)&497.4(41.59)\\
SHD&89.2(3.94)&101.2(4.66)&47.6(6.31)&66.5(9.98)&136(9.4)&166.5(13.57)\\
adjSHD&94.9(3.28)&117.7(4.55)&52.1(6.64)&86.8(12.21)&146.5(8.96)&203(15.66)\\\hline

\multicolumn{7}{c}{\texttt{BGe} with $iss=10$}\\\hline\hline
score&104.1&582.9&52.5(7.95)&2045.8(200.59)&169.1(5.74)&2181.2(149.48)\\
SHD&91.5(3.47)&115.3(5.01)&49.3(6.48)&86.1(13.7)&140.1(9.19)&199.8(17.66)\\
adjSHD&98.8(3.29)&166.8(8.38)&55.3(7.23)&163(14.57)&154.4(9.06)&324.7(18.82)\\
\hline
\hline
\end{tabular}
\end{center}
\end{table}

%%%sample size examination on sparse graph
\begin{table}[htbp]
\renewcommand{\arraystretch}{0.8}
\small
\begin{center}
\caption{Effect of sample size on Sparse Graph with $p=102, |\mathbb{E}|=109, \text{SNR} \in [0.5, 1.5]$. \label{tab:sparse_agg_diff_n}}
\begin{tabular}{rrrrr}
\hline\hline
& Correct E & Total E & Correct V & Total V \\
\hline
\multicolumn{5}{c}{n=50}\\
\hline
score&85.15(4.36)&482.1(5.71)&21.92(6.12)&3332.9(583.53)\\
SHD&37.63(4.69)&38.83(4.88)&5.66(2.46)&6.34(2.68)\\
adjSHD&57.42(5.04)&61.6(5.2)&9.86(3.3)&12.61(3.81)\\
%CPSF&40.58(5.33)&42.89(5.65)&9.15(3.42)&11.51(4.15)\\
%LN.CPSF&41.95(5.63)&44.22(5.92)&9.34(3.53)&11.6(4.14)\\
\hline
\multicolumn{5}{c}{n=102}\\
\hline
score&99.77(2.5)&336.79(16.79)&36.07(7.41)&672.46(90.68)\\
SHD&77.64(4.92)&80.34(5.37)&31.68(6.56)&35.06(7)\\
adjSHD&89.87(3.74)&98.43(4.38)&36.78(7.13)&46.94(8.46)\\
%CPSF&80.98(5.19)&86.74(6.12)&37.42(7.06)&46.16(9.03)\\
%LN.CPSF&82.18(4.81)&87.21(5.4)&37.44(7.03)&45.25(8.58)\\
  \hline
\multicolumn{5}{c}{n=200}\\
\hline
score&105.78(1.88)&257.3(11.95)&51.49(6.26)&358.46(40.25)\\
SHD&96.58(3.48)&101.24(4.44)&50.42(6.55)&58.07(7.92)\\
adjSHD&103.05(2.48)&116.13(4.24)&52.58(6.14)&70.67(9.08)\\
%CPSF&98.39(3.28)&107.82(5.1)&54.57(6.13)&69.96(9.45)\\
%LN.CPSF&99.36(2.83)&106.96(4.24)&54.64(6.14)&67.56(8.74)\\
\hline
\multicolumn{5}{c}{n=500}\\
\hline
score&108.35(0.87)&204.8(9.63)&58.34(5.06)&212.78(23.47)\\
SHD&104.64(2.12)&111.55(3.56)&61.93(5.4)&74.87(8.13)\\
adjSHD&107.64(1.13)&124.46(4.69)&61.54(5.03)&84.81(9.24)\\
%CPSF&105.22(2.13)&118.04(4.53)&64.08(4.65)&85.63(9.21)\\
%LN.CPSF&106.25(1.63)&117.51(3.81)&64.12(4.98)&83.07(8.81)\\
\hline
\multicolumn{5}{c}{n=1000}\\
\hline
score&108.71(0.48)&180.85(8.67)&61.74(6.05)&161.24(16.53)\\
SHD&106.62(1.52)&113.27(3.48)&64.79(4.43)&76.03(6.03)\\
adjSHD&108.72(0.55)&124.19(4.37)&63.95(4.22)&83.82(7.66)\\
%CPSF&106.84(1.56)&119.29(4.38)&66.61(4.62)&84.78(7.45)\\
%LN.CPSF&107.84(1.21)&118.56(4.13)&66.97(4.49)&83.41(6.92)\\
\hline
\multicolumn{5}{c}{n=5000}\\
\hline
score&108.99(0.1)&147.69(6.79)&65.04(3.89)&106.76(10.47)\\
SHD&107.68(1.24)&114.09(3.16)&65.18(3.53)&72.82(5.75)\\
adjSHD&108.97(0.17)&123.19(3.89)&64.02(3.57)&75.26(6.61)\\
%CPSF&107.79(1.15)&119.66(3.9)&66.04(3.31)&76.99(5.92)\\
%LN.CPSF&108.49(0.67)&119.68(3.98)&65.64(3.8)&75.61(5.99)\\
\hline\hline
 \end{tabular}
\end{center}
\end{table}

\begin{table}[H]
\renewcommand{\arraystretch}{0.8}
\small
\begin{center}
\caption{Robustness on Sparse Graph with $p=102, |\mathbb{E}|=109$, $n=102$, SNR $\in [0.5, 1.5]$. \label{tab:bic_robust}}
\begin{tabular}{rrrrrrr}
\hline
& Correct E & Total E & Correct V & Total V &Correct M & Total M\\
\hline
\multicolumn{7}{c}{Normal distribution}\\
\hline
score&99.77(2.5)&336.79(16.79)&36.07(7.41)&672.46(90.68)&155.4(7.07)&961.38(93.28)\\
SHD(2)&77.64(4.92)&80.34(5.37)&31.68(6.56)&35.06(7)&110.12(11.11)&115.2(11.62)\\
GSHD(1.5)&86.62(3.88)&91.4(4.43)&35.11(6.95)&41.25(7.67)&122.69(10.16)&132.35(11.09)\\
adjSHD(1)&89.87(3.74)&98.43(4.38)&36.78(7.13)&46.94(8.46)&128.17(10.1)&145.02(11.77)\\
GSHD(0.5)&92.33(3.35)&106.17(4.68)&38.23(7.09)&53.64(8.72)&132.77(9.68)&159.4(12.16)\\
%skeleton score&94.26(3.26)&114.39(4.77)&NA&NA&NA&NA\\
%CPSF&80.98(5.19)&86.74(6.12)&37.42(7.06)&46.16(9.03)&120.14(11.52)&132.54(14.25)\\
%LN.CPSF&82.18(4.81)&87.21(5.4)&37.44(7.03)&45.25(8.58)&120.92(10.98)&132.16(13.18)\\
\hline
\multicolumn{7}{c}{T-distribution, df=3}\\
\hline
score&99.81(2.92)&334.95(17.53)&40.38(6.96)&672.93(94.79)&157.72(7.31)&958.15(98.77)\\
SHD(2)&79.24(5.71)&82.77(6.09)&35.63(6.76)&40.84(7.59)&115.96(11.8)&123.3(12.91)\\
GSHD(1.5)&88.13(4.7)&94.46(5.01)&39.23(6.73)&48.05(7.91)&128.54(10.67)&142.02(11.99)\\
adjSHD(1)&91.29(3.98)&101.98(4.54)&40.81(6.57)&53.83(8.28)&133.72(9.89)&155.21(11.78)\\
GSHD(0.5)&93.44(3.67)&110.63(5.13)&41.95(6.71)&62.17(8.84)&137.63(9.68)&172.09(12.58)\\
%skeleton score&95.02(3.69)&119.56(5.35)&NA&NA&NA&NA\\
%CPSF&83.2(5.91)&91.45(6.38)&42.52(7.32)&56.27(9.31)&127.93(12.27)&147.05(14.72)\\
LN.CPSF&85.15(5.39)&92.57(5.79)&43.03(7.34)&55.39(8.74)&129.7(11.66)&147.31(13.57)\\
\hline
\multicolumn{7}{c}{T-distribution, df=5}\\
\hline
score&100.43(2.83)&333.51(16.03)&37.85(7.72)&663.69(87.57)&156.74(7.54)&948.72(89.64)\\
SHD(2)&78.26(4.59)&81.24(4.58)&31.92(6.38)&36.3(6.72)&111.66(10.06)&117.37(10.55)\\
GSHD(1.5)&87.21(4.09)&92.73(4.65)&35.38(6.47)&42.93(7.45)&124.27(9.31)&135.37(11.03)\\
adjSHD(1)&90.06(3.84)&99.24(4.74)&36.64(6.41)&47.95(7.91)&128.85(9.13)&146.86(11.51)\\
GSHD(0.5)&92.79(3.53)&107.81(4.9)&37.77(6.25)&55.4(8.47)&133.36(8.63)&162.76(11.91)\\
%skeleton score&94.49(3.42)&116.84(5.36)&NA&NA&NA&NA\\
%CPSF&81.71(4.59)&88.25(5.47)&37.91(6.03)&48.46(7.92)&122.03(9.68)&136.26(12.58)\\
%LN.CPSF&82.33(4.42)&87.88(5.15)&37.55(5.99)&46.88(7.56)&121.88(9.69)&134.35(11.89)\\
  \hline
\multicolumn{7}{c}{Gamma distribution, shape=1, scale=2}\\
\hline
score&99.13(3.04)&330.64(13.65)&39.31(6.5)&655.31(75.97)&155(7.44)&938.3(75.26)\\
SHD(2)&77.63(4.88)&80.7(5.13)&32.44(5.32)&36.57(6.46)&110.77(9.41)&117.03(10.7)\\
GSHD(1.5)&86.22(4.38)&91.93(4.75)&35.9(5.29)&43.06(6.66)&123.19(9.17)&134.64(10.52)\\
adjSHD(1)&89.18(4.12)&99.11(4.52)&37.66(5.5)&49.45(7.51)&128.33(8.9)&148.1(10.8)\\
GSHD(0.5)&91.71(3.71)&107.84(4.56)&39.05(5.42)&57.37(8.12)&133.01(8.34)&164.63(11.28)\\
%skeleton score&93.28(3.84)&116.28(5.26)&NA&NA&NA&NA\\
%CPSF&81.24(5.1)&88.28(5.77)&38.74(6.01)&49.91(8.1)&121.92(10.62)&137.63(12.85)\\
%LN.CPSF&82.54(4.85)&88.92(5.42)&38.91(6.14)&48.86(8.02)&123.04(10.25)&137.25(12.49)\\
\hline
 \end{tabular}
\end{center}
 \end{table}

\begin{table}[H]
\renewcommand{\arraystretch}{0.8}
\small
\begin{center}
\caption{Dense Graph with $p=104, |\mathbb{E}|=527, \text{v-structure}=1675$, Moral $|\mathbb{E}|=1670$, $n=100$, SNR $\in [0.5, 1.5]$.  \label{tab:indep_boot_verydense100}}
\begin{tabular}{rrrrrrr}
\hline\hline
& Correct E & Total E & Correct V & Total V & Correct M & Total M \\
\hline
score&254.7&524.69&206.58&1541.32&822.18&1786.11\\
&(12.41)&(22.15)&(37.37)&(165.14)&(52.76)&(136.96)\\
\hline
 \multicolumn{7}{c}{DAGBag}\\
\hline
SHD(2)&93.71(9.63)&98.06(9.6)&50.95(14.4)&66.85(16.87)&153.32(23.71)&164.12(25.17)\\
GSHD(1.5)&134.92(9.95)&145.23(10.08)&78.82(18.11)&122.23(22.94)&233.2(27.1)&264.9(30.98)\\
adjSHD(1)&164.31(10.02)&185.23(9.91)&104.54(20.03)&189.34(25.81)&303.28(27.63)&368.64(32.71)\\
GSHD(0.5)&190.93(10.69)&231.02(9.59)&131.03(23.4)&291.17(31.03)&381.01(29.46)&508.7(35.89)\\

%skeleton score&213.06(11.19)&278.92(10.63)&NA&NA&NA&NA\\
%vstruct score & 118.6(5.93)&131.42(6.03)&15.07(5.74)&24.23(6.51)&142.32(11.75)&155.54(5.93)\\
%CPSF&119.2(12.13)&130.05(12.51)&93.82(22.32)&145.04(29.02)&236.07(33.19)&271.39(38.28)\\
%LN.CPSF&109.67(11.39)&118.21(11.78)&81.69(20.75)&122.13(26.61)&209.47(31.17)&237.54(35.47)\\
\hline

\hline
\multicolumn{7}{c}{Independent data, 100 replicates}\\
\hline
SHD(2)&74&74&47&48&122&122\\
GSHD(1.5)&119&119&104&106&224&226\\
adjSHD(1)&153&153&162&170&314&317\\
GSHD(0.5)&192&192&215&238&410&419\\
%skeleton score&224&224&NA&NA&NA&NA\\
%vstruct score&114&114&13&14&127&128\\
%CPSF&123&123&161&164&281&281\\
%LN.CPSF&117&117&145&148&261&261\\
\hline
\hline
\end{tabular}
\end{center}
\end{table}

\begin{table}[H]
\renewcommand{\arraystretch}{0.8}
\small
\begin{center}
\caption{Dense Graph with $p=104, |\mathbb{E}|=527, \text{v-structure}=1675$, Moral edge$=1670$, $n=100$, SNR $\in [0.2, 0.5]$. \label{tab:indep_boot_verydense100_0205}}
\begin{tabular}{rrrrrrr}
\hline\hline
& Correct E & Total E & Correct V & Total V & Correct M & Total M \\
\hline
score&95.63&343.9&13.16&772.23&384.48&1054.78\\
&(7.02)&(19.57)&(5.39)&(121.11)&(42.27)&(118.02)\\
\hline
 \multicolumn{7}{c}{DAGBag}\\
\hline
SHD(2)&15.11(4.02)&17.42(4.46)&0.33(0.57)&2.01(1.82)&16.83(4.75)&19.43(5.67)\\
GSHD(1.5)&33.01(5.09)&46.14(6.31)&1.13(1.2)&10.92(4.5)&41.15(7.16)&57.04(9.93)\\
adjSHD(1)&47.57(5.58)&84.28(6.68)&2.45(1.9)&36.71(9.14)&70.57(9)&120.83(14.32)\\
GSHD(0.5)&61.75(6.11)&135.95(8.36)&4.56(2.66)&99.04(16.15)&113.95(11.59)&233.93(22.55)\\
%skeleton score&74.86(6.35)&196.27(9.79)&NA&NA&NA&NA\\
%vstruct score&50.15(5.17)&103.23(2.92)&0.08(0.27)&1.12(0.99)&62.53(5.25)&104.35(5.17)\\
%CPSF&17.42(4.86)&22.66(6.09)&0.87(1.07)&9.75(6.18)&23.06(7.13)&32.41(11.3)\\
%LN.CPSF&17.42(4.48)&22.36(5.77)&0.83(1.03)&9.14(6.07)&22.84(6.84)&31.49(10.84)\\
\hline

\multicolumn{7}{c}{Independent data, 100 replicates}\\
\hline
SHD(2)&2&2&0&0&2&2\\
GSHD(1.5)&17&17&0&0&17&17\\
adjSHD(1)&32&32&0&3&32&35\\
GSHD(0.5)&41&41&0&4&42&45\\
%skeleton score&52&52&NA&NA&NA&NA\\
%vstruct score&46&46&0&0&46&46\\
%CPSF&2&2&0&0&2&2\\
%LN.CPSF&2&2&0&0&2&2\\
\hline
\hline
\end{tabular}
\end{center}
\end{table}

\begin{table}[H]
\renewcommand{\arraystretch}{0.8}
\small
\begin{center}
\caption{Large Graph with $p=504, |\mathbb{E}|=515, \text{v-structure}=307$, Moral Edge$=808$, $n=100$, SNR $\in [0.5, 1.5]$. \label{tab:indep_boot_large100}}
\begin{tabular}{rrrrrrr}
\hline\hline
& Correct E & Total E & Correct V & Total V &Correct M &Total M\\
\hline
score&430.3&990.07&123.25&1151.77&585.22&2131.16\\
&(8.72)&(3.21)&(12.04)&(53.56)&(19.47)&(53.83)\\
\hline
 \multicolumn{7}{c}{100 bootstrap resamples}\\
\hline
SHD(2)&289.25(11.17)&292.92(10.94)&63.47(7.97)&66.3(7.74)&354.66(17.33)&422.07(15.65)\\
GSHD(1.5)&327.29(9.4)&332.12(9.39)&68.89(8.11)&73.35(7.92)&398.6(16.35)&405.08(16.07)\\
adjSHD(1)&338.5(8.96)&344.7(8.95)&71.91(8.28)&77.81(8)&413.29(16.07)&422.07(15.65)\\
GSHD(0.5)&347.12(9.09)&354.98(9.11)&74.57(8.63)&81.78(8.2)&424.82(16.42)&436.28(15.93)\\
%skeleton score&353.36(8.92)&363.02(8.8)&NA&NA&NA&NA\\
%vstruct score&340.89(8.08)&348.74(7.83)&40.02(6.54)&41.5(6.49)&383.46(13.11)&389.97(12.84)\\
%CPSF&298.96(11.74)&304.72(11.75)&78.03(9.21)&83.06(9.16)&380.74(19.39)&387.22(19.28)\\
%LN.CPSF&306.85(11.39)&312.69(11.13)&77.92(9.25)&82.9(9.17)&388.41(19.29)&395.04(18.95)\\
\hline

\multicolumn{7}{c}{Independent data, 100 replicates}\\
\hline
SHD(2)&421&421&178&178&591&591\\
GSHD(1.5)&463&463&194&195&650&650\\
adjSHD(1)&473&473&203&207&670&672\\
GSHD(0.5)&477&478&209&213&681&683\\
%skeleton score&483&483&NA&NA&NA&NA\\
%vstruct score&444&445&121&122&564&564\\
%CPSF&442&442&216&216&649&649\\
%LN.CPSF&460&460&225&225&676&676\\
\hline
\hline
\end{tabular}
\end{center}
\end{table}

\begin{table}[H]
\renewcommand{\arraystretch}{0.8}
\small
\begin{center}
\caption{Large Graph with $p=504, |\mathbb{E}|=515, \text{v-structure}=307$, Moral edge$=808$, $n=250$, SNR $\in [0.5, 1.5]$ . \label{tab:indep_boot_large250}}
\begin{tabular}{rrrrrrr}
\hline\hline
& Correct E & Total E & Correct V & Total V &Correct M &Total M\\
\hline
score&494.08&971.41&212.42&992.5&735&1945.78\\
&(4.94)&(6.2)&(12.07)&(35.9)&(14.73)&(37.82)\\
\hline
 \multicolumn{7}{c}{100 bootstrap resamples}\\
\hline
SHD(2)&440.28(8.57)&447.04(8.64)&182.7(13.07)&191.83(13.13)&624.32(19.95)&633.13(19.89)\\
GSHD(1.5)&458.99(7.4)&467.37(7.5)&186.54(12.54)&197.97(12.5)&647.71(18.32)&659.35(18.29)\\
adjSHD(1)&463.29(7.08)&473.57(7.38)&188.72(12.41)&201.37(12.31)&655.34(17.97)&668.87(17.84)\\
GSHD(0.5)&466.07(7.05)&478.04(7.44)&190.14(12.48)&204.15(12.26)&660.58(17.8)&676(17.68)\\
%skeleton score&468.08(7.18)&481.72(7.62)&NA&NA&NA&NA\\
%vstruct score&455.11(7.55)&465.89(7.67)&136.23(13.34)&142.19(13.52)&594.8(19.29)&603.08(7.55)\\
%CPSF&444.13(8.51)&453.61(8.68)&193.39(13.04)&202.9(13.07)&641.92(19.98)&650.32(19.83)\\
%LN.CPSF&452.44(8.04)&461.77(8.62)&192.47(12.59)&201.91(12.76)&648.57(19.38)&657.64(19.33)\\
\hline

\multicolumn{7}{c}{Independent data, 100 replicates}\\
\hline
SHD(2)&497&497&274&275&758&758\\
GSHD(1.5)&511&513&275&279&775&776\\
adjSHD(1)&513&515&275&280&777&779\\
GSHD(0.5)&507&509&259&263&758&759\\
%skeleton score&514&517&NA&NA&NA&NA\\
%vstruct score&495&497&209&215&697&700\\
%CPSF&499&499&280&281&766&766\\
%LN.CPSF&506&506&280&281&773&773\\
\hline
\hline
\end{tabular}
\end{center}
\end{table}

\begin{table}[H]
\renewcommand{\arraystretch}{0.8}
\small
\begin{center}
\caption{Large Graph with $p=504, |\mathbb{E}|=515, \text{v-structure}=307$, Moral Edge$=808$, $n=100$, SNR $\in [0.2, 0.5]$. \label{tab:indep_boot_large100_0205}}
\begin{tabular}{rrrrrrr}
\hline\hline
& Correct E & Total E & Correct V & Total V &Correct M &Total M\\
\hline
score&371.45&991.51&94.71&1232.6&494.96&2214.02\\
&(12.04)&(2.79)&(13.42)&(64.51)&(22.88)&(64.05)\\
\hline
 \multicolumn{7}{c}{100 bootstrap resamples}\\
\hline
SHD(2)&231.09(11.63)&233.52(11.97)&49(8.48)&51.29(8.57)&281.95(18.6)&284.56(18.86)\\
GSHD(1.5)&265.86(10.72)&269.23(10.87)&53.94(8.97)&57.1(8.88)&321.77(18.2)&325.99(18.23)\\
adjSHD(1)&276.54(10.9)&281.1(10.87)&56.55(9.08)&60.6(9.09)&335.34(18.54)&341.31(18.52)\\
GSHD(0.5)&285.37(10.79)&291.75(10.98)&58.82(9.16)&63.93(9.35)&346.72(18.47)&355.21(18.74)\\

%skeleton score&291.21(10.98)&299.72(11.29)&NA&NA&NA&NA\\
%vstruct score&282.89(10.38)&290.33(10.46)&26.75(6.4)&27.95(6.65)&312(14.98)&318.1(10.38)\\
%CPSF&239.25(12.34)&243.31(12.49)&61.24(9.12)&65.3(9.19)&303.88(20.09)&308.14(20.25)\\
%LN.CPSF&244.1(12.99)&248.19(13.25)&60.56(9.22)&64.6(9.33)&307.89(20.7)&312.32(21.1)\\
\hline

\multicolumn{7}{c}{Independent data, 100 replicates}\\
\hline
SHD(2)&271&271&74&74&344&344\\
GSHD(1.5)&391&391&122&122&511&511\\
adjSHD(1)&427&427&140&142&565&566\\
GSHD(0.5)&448&448&160&163&606&607\\
%skeleton score&457&457&NA&NA&NA&NA\\
%vstruct score&407&407&39&39&445&445\\
%CPSF&325&325&150&150&470&470\\
%LN.CPSF&342&342&152&152&489&489\\
\hline
\hline
\end{tabular}
\end{center}
\end{table}

\begin{table}[H]
\renewcommand{\arraystretch}{0.8}
\small
\begin{center}
\caption{Large Graph with $p=504, |\mathbb{E}|=515, \text{v-structure}=307$, Moral edge$=808$, $n=250$, SNR $\in [0.2, 0.5]$. \label{tab:indep_boot_large250_0205}}
\begin{tabular}{rrrrrrr}
\hline\hline
& Correct E & Total E & Correct V & Total V &Correct M &Total M\\
\hline
score&451.33&979.49&165.88&1038.99&643.48&2003.73\\
&(8.74)&(5.03)&(14.15)&(36.97)&(20.52)&(37.33)\\
\hline
 \multicolumn{7}{c}{100 bootstrap resamples}\\
\hline
SHD(2)&365.33(11.02)&370.45(11.17)&137.59(13.9)&144.51(14.48)&504.69(22.42)&511.07(22.91)\\
GSHD(1.5)&388.42(10.47)&395.17(10.48)&140.9(14.03)&149.57(14.83)&532.02(21.66)&540.71(22.5)\\
adjSHD(1)&394.57(10.44)&402.67(10.4)&142.64(14.17)&152.46(14.92)&540.38(21.66)&551(22.53)\\
GSHD(0.5)&399.39(10.5)&409.31(10.45)&144.25(14.45)&155.67(15.59)&547.51(22.36)&560.79(23.35)\\

%skeleton score&403.19(10.05)&414.92(10.03)&NA&NA&NA&NA\\
%vstruct score&393.37(9.55)&403.22(9.66)&95.77(14.69)&99.82(15.13)&492.59(21.88)&499.39(9.55)\\
%CPSF&369.88(11.01)&377.17(11.15)&147.35(14.25)&155.21(14.64)&520.96(22.55)&528.16(23.04)\\
%LN.CPSF&376.28(11.41)&383.33(11.4)&146.67(14.16)&154.35(14.78)&526.14(22.63)&533.44(23.19)\\
\hline

\multicolumn{7}{c}{Independent data, 100 replicates}\\
\hline
SHD(2)&463&463&256&258&707&707\\
GSHD(1.5)&504&504&268&270&760&760\\
adjSHD(1)&512&512&268&273&768&770\\
GSHD(0.5)&512&512&268&273&768&770\\
%skeleton score&514&515&NA&NA&NA&NA\\
%vstruct score&486&487&191&191&669&669\\
%CPSF&478&478&276&278&741&741\\
%LN.CPSF&493&493&280&282&760&760\\
\hline
\hline
\end{tabular}
\end{center}
\end{table}

\begin{table}[H]%%100 replicates
\renewcommand{\arraystretch}{0.8}
\small
\begin{center}
\caption{Extra-large Graph, $p=1000,|\mathbb{E}|=1068, \text{v-structure}=785$, Moral $|\mathbb{E}|=1823$, $n=100$, SNR $\in [0.5, 1.5]$. \label{tab:extra_large_100}}
\begin{tabular}{rrrrrrr}
\hline\hline
&Correct E & Total E & Correct V & Total V &Correct M & Total M\\
\hline
score&881.2&4944.07&317.64&74803.84&1339.19&74325.35\\
&(13.03)&(9.47)&(22.52)&(4928.25)&(30.62)&(4284.2)\\
\hline
SHD(2)&552.26(18.07)&557.37(18.02)&145.94(12.91)&149.7(13.13)&700.5(29.77)&706.21(29.96)\\
GSHD(1.5)&625.57(17.04)&632.67(16.96)&162.05(13.4)&167.8(13.72)&790.34(28.99)&799.38(29.41)\\
adjSHD(1)&654.31(16.56)&663.81(16.55)&171.51(14.08)&179.13(14.32)&828.86(29.2)&841.74(29.43)\\
GSHD(0.5)&676.94(15.99)&690.17(15.91)&180.23(14.16)&190.86(14.17)&860.94(28.5)&879.78(28.51)\\
%skeleton score&692.21(15.07)&709.84(15.21)&NA &NA &NA &NA\\
%vstruct score&667.38(14.44)&681.72(14.33)&80.46(11.64)&82.69(11.97)&751.64(23.91)&763.79(24.18)\\
%CPSF&578.68(18.7)&587.64(18.66)&185.45(14.98)&193.77(15.36)&769.3(32.13)&780.03(32.39)\\
%LN.CPSF&588.92(18.68)&597.37(18.52)&184.44(14.67)&192.36(14.88)&778.09(31.53)&788.39(31.75)\\
\hline\hline
\end{tabular}
\end{center}
\end{table}

\begin{table}[H]%%100 replicates
\renewcommand{\arraystretch}{0.8}
\small
\begin{center}
\caption{Extra-large Graph, $p=1000,|\mathbb{E}|=1068, \text{v-structure}=785$, Moral $|\mathbb{E}|=1823$, $n=250$,SNR $\in [0.5, 1.5]$. \label{tab:extra_large_250}}
\begin{tabular}{rrrrrrr}
\hline\hline
& Correct E & Total E & Correct V & Total V& Correct M & Total M\\
\hline
score&1028.11&4935.95&539.58&17197.54&1648.67&21761.42\\
&(6.56)&(8.44)&(21.12)&(418.56)&(23.95)&(406.46)\\
\hline
SHD(2)&921.9(12.99)&934.66(13.48)&478.58(24.76)&494.38(25.19)&1398.23(35.21)&1417.66(35.99)\\
GSHD(1.5)&955.04(12.01)&972.65(13.05)&490.92(24.59)&511.88(25.18)&1445.92(34.14)&1472.74(35.26)\\
adjSHD(1)&964.23(11.62)&987.12(12.89)&496.74(24.68)&521.67(25.38)&1463.01(34.21)&1496.75(35.55)\\
GSHD(0.5)&971.4(10.68)&1001.74(12.41)&502.39(24.14)&532.38(25.15)&1478.01(32.67)&1521.87(34.79)\\
%skeleton&975.6(10.23)&1014.38(12.26)&NA&NA&NA&NA\\
%vstruct score&951.67(11.45)&980.9(12.92)&365.6(24.04)&375.44(23.92)&1322.25(32.89)&1346.41(33.51)\\
%CPSF&934.19(13.07)&954.75(13.68)&508.7(24.92)&531.34(25.53)&1445.84(35.27)&1473.66(36.5)\\
%LN.CPSF&941.41(13.11)&960.65(13.79)&507.2(25.66)&527.91(26.07)&1450.54(35.86)&1476.58(37.28)\\
\hline\hline
\end{tabular}
\end{center}
\end{table}

\begin{table}[H]%%100 replicates
\renewcommand{\arraystretch}{0.8}
\small
\begin{center}
\caption{Ultra-large Graph, $p=2639,|\mathbb{E}|=2603, \text{v-structure}=1899$, Moral $|\mathbb{E}|=4481$, $n=250$, SNR $\in [0.5, 1.5]$. \label{tab:super_large_250}}
\begin{tabular}{rrrrrrr}
\hline\hline
& Correct E & Total E & Correct V & Total V& Correct M & Total M\\
\hline
score&2500(11)&9861(11)&1290(40)&26476(3534)&3965(43)&36211(3513)\\
SHD(2)&2162(24)&2175(24)&1095(42)&1109(43)&3265(63)&3279(64)\\
GSHD(1.5)&2268(21)&2286(21)&1127(40)&1148(40)&3404(58)&3428(59)\\
adjSHD(1)& 2296(20)&2318(21)&1144(39)&1169(39)&3452(56)&3481(57)\\
GSHD(0.5)& 2317(18)&2344(19)&1158(38)&1188(38)&3489(54)&3526(54)\\
%skeleton score&2330(18)&2364(19)& NA&NA &NA&NA\\
%CPSF&2202(24)&2222(25)&1190(40)&1209(41)&3406(62)&3425(63)\\
%LN.CPSF& 2226(24)&2245(25)&1186(40)&1204(40)&3424(61)&3443(62)\\
\hline\hline
\end{tabular}
\end{center}
\end{table}

\begin{table}%[H]
\renewcommand{\arraystretch}{0.8}
\begin{center}
\caption{Bagged estimate of log-likelihood score \cite{elidan2011bagged} on Sparse Graph $p=102, |\mathbb{E}|=109, n=102, SNR \in [0.5, 1.5]$. \label{tab:bagged_est}} %p=102,E=109,Vstructure=77, one replicate
\begin{tabular}{rrrr}\hline\hline
Number of bootstrap resamples &Correct E & Total E\\ \hline
 B=10& 101 &629 \\
 B=50& 102 & 511 \\
 B=100& 102 & 509 \\
 B=500& 102 & 561\\
\hline\hline
\end{tabular}
\end{center}
\end{table}

%%%%%%%%%%%%%%%%%%

\begin{thebibliography}{10}

\bibitem{Bach08}
Francis~R Bach.
\newblock Bolasso: model consistent lasso estimation through the bootstrap.
\newblock ICML, 2008.

\bibitem{bishop2006pattern}
C.M. Bishop et~al.
\newblock {\em Pattern recognition and machine learning}, volume~4.
\newblock springer New York, 2006.

\bibitem{Breiman96}
L.~Breiman.
\newblock Bagging predictors.
\newblock {\em Machine learning}, 24(2):123--140, 1996.

\bibitem{broom2012model}
B.M. Broom, K.A. Do, and D.~Subramanian.
\newblock Model averaging strategies for structure learning in bayesian
  networks with limited data.
\newblock {\em BMC Bioinformatics}, 13(Suppl 13):S10, 2012.

\bibitem{chen2008extended}
Jiahua Chen and Zehua Chen.
\newblock Extended bayesian information criteria for model selection with large
  model spaces.
\newblock {\em Biometrika}, 95(3):759--771, 2008.

\bibitem{chickering2002optimal}
David~Maxwell Chickering.
\newblock Optimal structure identification with greedy search.
\newblock {\em The Journal of Machine Learning Research}, 3:507--554, 2002.

\bibitem{chickering2002learning}
D.M. Chickering.
\newblock Learning equivalence classes of bayesian-network structures.
\newblock {\em The Journal of Machine Learning Research}, 2:445--498, 2002.

\bibitem{de2000approximating}
L.M. de~Campos and J.F. Huete.
\newblock Approximating causal orderings for bayesian networks using genetic
  algorithms and simulated annealing.
\newblock In {\em Proceedings of the Eight Conference on Information Processing
  and Management of Uncertainty in Knowledge-Based Systems}, pages 333--340,
  2000.

\bibitem{elidan2011bagged}
G.~Elidan.
\newblock Bagged structure learning of bayesian networks.
\newblock 2011.

\bibitem{elidan2002data}
G.~Elidan, M.~Ninio, N.~Friedman, and D.~Shuurmans.
\newblock Data perturbation for escaping local maxima in learning.
\newblock In {\em PROCEEDINGS OF THE NATIONAL CONFERENCE ON ARTIFICIAL
  INTELLIGENCE}, pages 132--139. Menlo Park, CA; Cambridge, MA; London; AAAI
  Press; MIT Press; 1999, 2002.

\bibitem{friedman1999data}
N.~Friedman, M.~Goldszmidt, and A.~Wyner.
\newblock Data analysis with bayesian networks: A bootstrap approach.
\newblock In {\em Proceedings of the Fifteenth conference on Uncertainty in
  artificial intelligence}, pages 196--205. Morgan Kaufmann Publishers Inc.,
  1999.

\bibitem{friedman2000using}
N.~Friedman, M.~Linial, I.~Nachman, and D.~Pe'er.
\newblock Using bayesian networks to analyze expression data.
\newblock {\em Journal of computational biology}, 7(3-4):601--620, 2000.

\bibitem{geiger1994learning}
D.~Geiger and D.~Heckerman.
\newblock Learning gaussian networks.
\newblock In {\em Proceedings of the Tenth international conference on
  Uncertainty in artificial intelligence}, pages 235--243. Morgan Kaufmann
  Publishers Inc., 1994.

\bibitem{heckerman2008tutorial}
D.~Heckerman.
\newblock A tutorial on learning with bayesian networks.
\newblock {\em Innovations in Bayesian Networks}, pages 33--82, 2008.

\bibitem{heckerman1995learning}
D.~Heckerman, D.~Geiger, and D.M. Chickering.
\newblock Learning bayesian networks: The combination of knowledge and
  statistical data.
\newblock {\em Machine learning}, 20(3):197--243, 1995.

\bibitem{imoto2002bootstrap}
S.~Imoto, S.Y. Kim, H.~Shimodaira, S.~Aburatani, K.~Tashiro, S.~Kuhara, and
  S.~Miyano.
\newblock Bootstrap analysis of gene networks based on bayesian netowrks and
  nonparamatric regression.
\newblock {\em Genome Informatics Series}, pages 369--370, 2002.

\bibitem{kalisch2007estimating}
M.~Kalisch and P.~B{\"u}hlmann.
\newblock Estimating high-dimensional directed acyclic graphs with the
  pc-algorithm.
\newblock {\em The Journal of Machine Learning Research}, 8:613--636, 2007.

\bibitem{kim2012consistent}
Yongdai Kim, Sunghoon Kwon, and Hosik Choi.
\newblock Consistent model selection criteria on high dimensions.
\newblock {\em The Journal of Machine Learning Research}, 98888:1037--1057,
  2012.

\bibitem{koller2009probabilistic}
D.~Koller and N.~Friedman.
\newblock {\em Probabilistic graphical models: principles and techniques}.
\newblock MIT press, 2009.

\bibitem{kruskal1956shortest}
Joseph~B Kruskal.
\newblock On the shortest spanning subtree of a graph and the traveling
  salesman problem.
\newblock {\em Proceedings of the American Mathematical society}, 7(1):48--50,
  1956.

\bibitem{lauritzen1996graphical}
S.L. Lauritzen.
\newblock {\em Graphical models}, volume~17.
\newblock Oxford University Press, USA, 1996.

\bibitem{li2011bootstrap}
S.~Li, L.~Hsu, J.~Peng, and P.~Wang.
\newblock Bootstrap inference for network construction.
\newblock {\em The Annals of Applied Statistics}, 7(1):391--417, 2013.

\bibitem{margaritis2003learning}
Dimitris Margaritis.
\newblock {\em Learning Bayesian network model structure from data}.
\newblock PhD thesis, University of Pittsburgh, 2003.

\bibitem{meinshausen2010stability}
N.~Meinshausen and P.~B{\"u}hlmann.
\newblock Stability selection.
\newblock {\em Journal of the Royal Statistical Society: Series B (Statistical
  Methodology)}, 72(4):417--473, 2010.

\bibitem{neto2010causal}
Elias~Chaibub Neto, Mark~P Keller, Alan~D Attie, and Brian~S Yandell.
\newblock Causal graphical models in systems genetics: a unified framework for
  joint inference of causal network and genetic architecture for correlated
  phenotypes.
\newblock {\em The annals of applied statistics}, 4(1):320, 2010.

\bibitem{pearl2000causality}
Judea Pearl.
\newblock {\em Causality: models, reasoning and inference}, volume~29.
\newblock Cambridge Univ Press, 2000.

\bibitem{pe2001inferring}
D.~PeÕer, A.~Regev, G.~Elidan, and N.~Friedman.
\newblock Inferring subnetworks from perturbed expression profiles.
\newblock {\em Bioinformatics}, 17(suppl 1):S215--S224, 2001.

\bibitem{perrier2008finding}
E.~Perrier, S.~Imoto, and S.~Miyano.
\newblock Finding optimal bayesian network given a super-structure.
\newblock {\em Journal of Machine Learning Research}, 9(2):2251--2286, 2008.

\bibitem{prim1957shortest}
Robert~Clay Prim.
\newblock Shortest connection networks and some generalizations.
\newblock {\em Bell system technical journal}, 36(6):1389--1401, 1957.

\bibitem{russell2010artificial}
S.J. Russell, P.~Norvig, E.~Davis, S.J. Russell, and S.J. Russell.
\newblock {\em Artificial intelligence: a modern approach}.
\newblock Prentice hall Upper Saddle River, NJ, 2010.

\bibitem{sachs2005causal}
K.~Sachs, O.~Perez, D.~Pe'er, D.A. Lauffenburger, and G.P. Nolan.
\newblock Causal protein-signaling networks derived from multiparameter
  single-cell data.
\newblock {\em Science Signalling}, 308(5721):523, 2005.

\bibitem{schmidt2007learning}
Mark Schmidt, Alexandru Niculescu-Mizil, and Kevin Murphy.
\newblock Learning graphical model structure using l1-regularization paths.
\newblock In {\em Proceedings of the National Conference on Artificial
  Intelligence}, volume~22, page 1278. Menlo Park, CA; Cambridge, MA; London;
  AAAI Press; MIT Press; 1999, 2007.

\bibitem{scutari2009learning}
M.~Scutari.
\newblock Learning bayesian networks with the bnlearn r package.
\newblock {\em Journal of Statistical Software}, 35(3), 2010.

\bibitem{spirtes2001causation}
P.~Spirtes, C.~Glymour, and R.~Scheines.
\newblock {\em Causation, prediction, and search}, volume~81.
\newblock MIT press, 2001.

\bibitem{teyssier2012ordering}
M.~Teyssier and D.~Koller.
\newblock Ordering-based search: A simple and effective algorithm for learning
  bayesian networks.
\newblock {\em arXiv preprint arXiv:1207.1429}, 2012.

\bibitem{tsamardinos2006max}
I.~Tsamardinos, L.E. Brown, and C.F. Aliferis.
\newblock The max-min hill-climbing bayesian network structure learning
  algorithm.
\newblock {\em Machine learning}, 65(1):31--78, 2006.

\bibitem{Verma1991equivalence}
T.~Verma and J.~Pearl.
\newblock Equivalence and synthesis of causal models.
\newblock {\em In Henrion, M., Shachter, R. Kanal, L., and Lemmer, J., editors,
  Proceeding of the Sixth Conference on Uncertainty in Artificial
  Intelligence}, pages 220--227, 1991.

\bibitem{wang2011random}
S.~Wang, B.~Nan, S.~Rosset, and J.~Zhu.
\newblock Random lasso.
\newblock {\em The annals of applied statistics}, 5(1):468, 2011.

\bibitem{yaramakala2005speculative}
S.~Yaramakala and D.~Margaritis.
\newblock Speculative markov blanket discovery for optimal feature selection.
\newblock In {\em Data Mining, Fifth IEEE International Conference on}, pages
  4--pp. IEEE, 2005.

\end{thebibliography}
\end{document}